\documentclass{article}

\usepackage{graphicx}

\usepackage[preprint]{neurips_2025}

\usepackage[utf8]{inputenc} 
\usepackage[T1]{fontenc}    
\usepackage{hyperref}       
\usepackage{url}            
\usepackage{booktabs}       
\usepackage{amsfonts}       
\usepackage{nicefrac}       
\usepackage{microtype}      
\usepackage{xcolor}

\title{Measuring Time-Series Dataset Similarity using Wasserstein Distance}

\author{
  Hongjie Chen \\
  Dolby Labs. \\
  Atlanta, GA \\
  \texttt{hongjie.chen@dolby.com} \\
  \And
  Akshay Mehra \\
  Dolby Labs. \\
  Atlanta, GA \\
  \And
  Josh Kimball \\
  Dolby Labs. \\
  Atlanta, GA \\
  \And
  Ryan A. Rossi \\
  Adobe Research \\
  San Jose, CA \\
}
\newcommand{\coloneqq}{\mathrel{\mathop:}=}
\usepackage{lipsum}

\usepackage{graphicx}
\usepackage{tikz}
\usepackage{amsmath, amssymb}

\definecolor{Teal}{HTML}{008080}
\definecolor{Coral}{HTML}{FF7F50}
\definecolor{Goldenrod}{HTML}{DAA520}
\definecolor{DarkOrchid}{HTML}{9932CC}
\definecolor{SlateBlue}{HTML}{6A5ACD}

\DeclareMathAlphabet\mathbfcal{OMS}{cmsy}{b}{n}

\providecommand{\confPaper}[1]{}
\usepackage{comment}
\usepackage{bm}

\usepackage{algorithm}
\usepackage{algorithmic}
\usepackage{xcolor}

\usepackage{multicol}
\usepackage{multirow}
\usepackage{booktabs}
\usepackage{tabularx}
\usepackage{subfigure}
\definecolor{thedarkblue}{RGB}{0,0,120}
\definecolor{mygreen}{RGB}{0,100,0}
\definecolor{mydarkblue}{rgb}{0,0.08,0.45}
\definecolor{darkblue}{rgb}{0,0.08,180}
\colorlet{TufteRed}{red!80!black}

\definecolor{theblue}{RGB}{0,0,180}
\colorlet{thered}{TufteRed}
\definecolor{googleblue}{HTML}{4285F4}
\definecolor{googlegreen}{HTML}{0F9D58}
\definecolor{googlered}{HTML}{DB4437}
\usepackage{todonotes}

\usepackage{stmaryrd}

\providecommand{\eat}[1]{\ignorespaces}
\providecommand{\addressed}[1]{\ignorespaces}
\providecommand{\todoeat}[1]{\ignorespaces}
\providecommand{\supp}[1]{\ignorespaces}

\providecommand{\pare}[1]{\ensuremath{\left(#1\right)}}

\providecommand{\brac}[1]{\ensuremath{\left[#1\right]}}

\providecommand{\abs}[1]{\left|#1\right|}

\newcolumntype{H}{>{\setbox0=\hbox\bgroup}c<{\egroup}@{}}

\providecommand{\mat}[1]{\boldsymbol{\mathrm{#1}}}

\DeclareMathOperator*{\DTW}{DTW}

\newcommand{\norm}[1]{\left\lVert#1\right\rVert}

\DeclareMathOperator{\hugeE}{\mbox{\huge\raise-0.3ex\hbox{E}}}
\DeclareMathOperator{\p}{\mathbb{P}}
\DeclareMathOperator{\hugep}{\mbox{\huge\raise-0.3ex\hbox{$\p$}}}

\DeclareMathOperator{\trace}{tr}

\DeclareMathOperator{\bigO}{\mathcal{O}}

\providecommand{\Ws}{\mathrm{Ws}}
\providecommand{\RR}{\mathbb{R}}

\providecommand{\MVN}{\mathcal{MVN}}

\providecommand{\vmu}{\boldsymbol{\mu}}
\providecommand{\mSigma}{\boldsymbol{\Sigma}}

\def\transpose{^\mathsf{T}}

\providecommand{\mX}{\ensuremath{\mat{X}}}
\providecommand{\bmX}{\ensuremath{\boldsymbol{X}}}
\providecommand{\mY}{\ensuremath{\mat{Y}}}
\providecommand{\bmY}{\ensuremath{\boldsymbol{Y}}}

\renewcommand\vec{\mathbf}

\providecommand{\vx}{\ensuremath{\vec{x}}}
\providecommand{\vy}{\ensuremath{\vec{y}}}

\providecommand{\calD}{\ensuremath{\mathcal{D}}}

\providecommand{\domA}[1]{\ensuremath{\RR^{#1} }}
\providecommand{\dom}[2]{\ensuremath{\RR^{#1 \times #2} }}

\usepackage{amsthm}

\begin{document}

\maketitle

\begin{abstract}
The emergence of time-series foundation model research elevates the growing need to measure the (dis)similarity of time-series datasets.
A time-series dataset similarity measure aids research in multiple ways, including model selection, finetuning, and visualization.
In this paper, we propose a distribution-based method to measure time-series dataset similarity by leveraging the Wasserstein distance.
We consider a time-series dataset an empirical instantiation of an underlying multivariate normal distribution (MVN).
The similarity between two time-series datasets is thus computed as the Wasserstein distance between their corresponding MVNs. 
Comprehensive experiments and visualization show the effectiveness of our approach.
Specifically, we show how the Wasserstein distance helps identify similar time-series datasets and facilitates inference performance estimation of foundation models in both out-of-distribution and transfer learning evaluation, with high correlations between our proposed measure and the inference loss ($>0.60$).
\end{abstract}

\section{Introduction}
Recent advances in deep learning have enabled time-series applications in a wide range of disciplines, including agriculture, business, meteorology, finance, healthcare, and physiology, among others~\cite{morid2023time,wang2016research,sezer2020financial,chen2025probabilistic,iizumi2024hybrid,adebayo2021economic,lim2021time,chen2024graph,benidis2022deep}.
In circumstances where the time-series datasets of interest (target datasets) only contain limited data, methods such as transfer learning and cross-domain learning~\cite{liu2024unitime} harness the abundant data from other time-series datasets (source datasets).
This idea is amplified in foundation models, which are trained on a large collection of source datasets.
Once foundation models are trained, users only need to plug in target datasets for inference, without the need of further training~\cite{ansari2024chronos,Yuqietal-2023-PatchTST,rasul2024lag,das2023decoder}.
Since foundation models are not necessarily trained on the target datasets, a natural research question arises: 
\begin{itemize}
    \item \textit{How to estimate the performance of foundation models on unseen target datasets?}
\end{itemize}

\begin{figure}[t]
\centering
\includegraphics[width=0.75\linewidth]{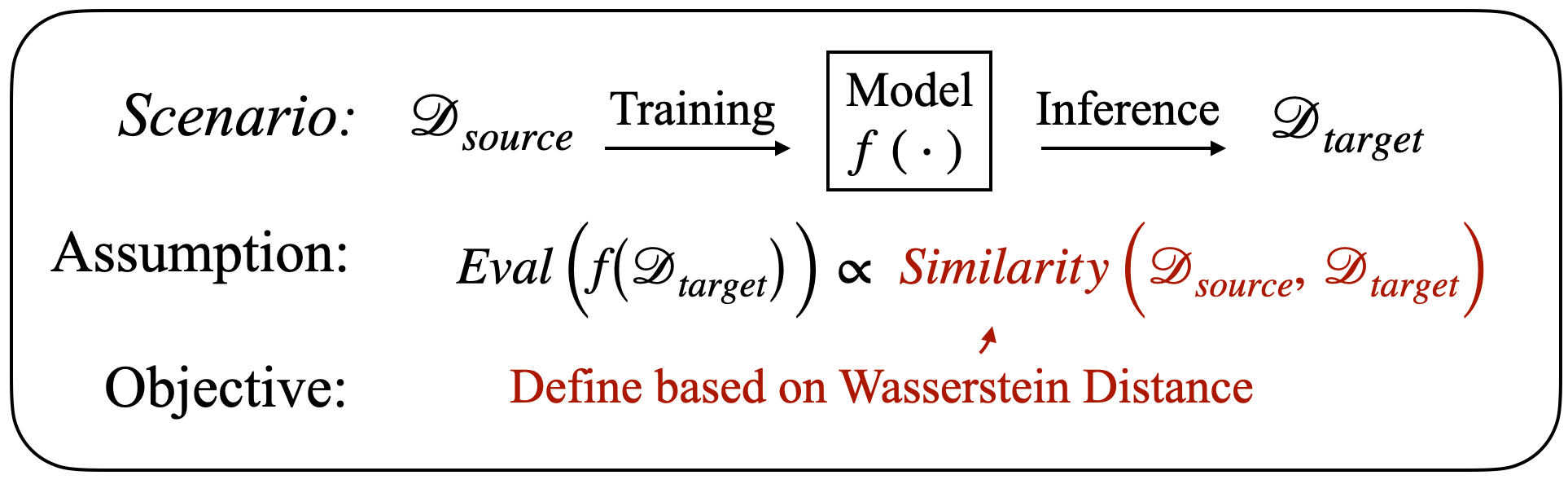}
\caption{
Foundation models are trained on source datasets and applied to target datasets.
Presumably, the inference performance on the target dataset depends on the similarity between the source and target datasets.
Therefore, this paper aims to define a time-series dataset similarity measure using the Wasserstein Distance, which helps estimate the performance of foundation models on target datasets.
}
\label{fig:motivation}
\end{figure}

An efficient estimation can significantly help model users save time by avoiding the random application of foundation models to their data.
Intuitively, the performance of foundation models depends on at least two factors: 
(1) the generalization capacity of the foundation models on the source datasets; and 
(2) the similarity between the source and the target datasets~\cite{mehraunderstanding}.
In this paper, we focus on the latter factor.
We assume
foundation models and other related deep learning models (e.g., transfer learning models) generally perform better on target datasets that are similar to source datasets than those that are less similar.
A key step to tackle this factor is to define \textit{similarity} or \textit{distance} between time-series datasets.
Hence, this paper aims to propose a distance function that measures the (dis)similarity between time-series datasets.
Our motivation is depicted in Fig.~\ref{fig:motivation}.

In addition to determining the performance of foundation models on target datasets, a similarity measure provides multifold benefits, including effectiveness, fairness, as well as visualization.
In terms of \textit{effectiveness}, measuring dataset distances helps researchers determine strategies for finetuning models.
On one hand, in-distribution datasets or closely similar datasets do not provide models with much extra information, and thus may only need limited training on them.
On the other hand, training on out-of-distribution datasets requires precautions to prevent forgetting earlier training datasets~\cite{gururangan2020don,sanyal2024accuracy}.
Our proposed similarity measure also benefits the measurement of \textit{fairness} for foundation models.
One aspect of fairness relies on whether models deliver similar results from similar input datasets~\cite{dwork2012fairness}, which can be assessed by our proposed similarity measure.
Moreover, the similarity measure facilitates an insightful understanding of the relationships between time-series datasets through \textit{visualization}.
For example, pairwise similarities between datasets can be computed to build a graph (Sec.~\ref{sec:visualization}), which can be further visualized using graph layout algorithms~\cite{zhu2020drgraph,belkin2003laplacian}.

In this paper, we propose a similarity measure to assess the (dis)similarity between any two time-series datasets.
Our similarity measure is based on a specific form of the Wasserstein distance, which treats time-series datasets as multivariate normal distributions.
Our key contributions are listed below:
\begin{enumerate}
    \item We propose modeling time-series datasets as multivariate normal distributions and leverage the Wasserstein distance to measure their similarity / distance.
    \item We validate the effectiveness of our time-series dataset similarity approach through visualizations and comparison with other similarity measures.
    \item Experiments on out-of-distribution and transfer learning tasks demonstrate that our proposed similarity serves as an informative indicator of model performance, with high correlations between our proposed measure and the inference loss ($>0.60$).
\end{enumerate}

\section{Related Work}
In this section, we discuss closely related work on dataset distances and distribution-based distances.

\subsection{Dataset Distances}
Research on measuring dataset distances has mainly focused on datasets for classification, clustering, and searching tasks~\cite{behme2024fainder,shnitzer2022log,lin2023integrated,chen2023graph}.
For example, \citet{alvarez2020geometric} leverage distances based on the optimal transport theory to compute distances between image datasets.
More specifically, they aim to determine the optimal transport from the label distribution of one dataset to that of the other.
Their work is the most similar to ours, as both approaches are model-agnostic and require no training.
However, unlike their focus on classification datasets, our similarity measure is specifically proposed for time-series datasets,
which functions in continuous space instead of the discrete class space.
Metashift~\cite{liang2022metashift} is another work that examines image dataset distances across distributions.
The model draws label distributions of images from their image metadata.
It further represents images as nodes and visualizes them based on graph geometry~\cite{belkin2003laplacian}.

In terms of other types of datasets, ~\citet{yang2022fast} leverage Earth mover's distance to measure the similarity between two spatial datasets, which enables fast searching of similar datasets.
Similarity measures between stochastic differential systems have also been explored by~\cite{wang2023similarity}, which is based on the conjugate theory of stochastic systems.
Moreover, similarity measures for time series have been extensively studied, including those based on Fréchet distance~\cite{driemel2016clustering}, Dynamic Time Warping~\cite{cuturi2017soft}, among others~\cite{abanda2019review,nowozin2016f,ding2008querying}.
However, instead of our focus on time-series datasets, these measures all operate on individual time series, making them unsuitable for our objective.

\subsection{Distribution-based Distances}
Measuring similarities between distributions is a crucial step for many deep learning models~\cite{sierra2025divshift,mad2023}.
One line of research addresses distribution matches for the purposes of dataset condensation and dataset distillation~\cite{ijcai2024p186,zhao2023improved}, which allow smaller datasets to represent larger ones without losing training effectiveness.

Distance measures for comparing distributions have also been proposed and studied~\cite{mallasto2017learning,masarotto2019procrustes}.
The Wasserstein distance, in particular, is widely used to measure the distance between multivariate normal distributions~\cite{wu2018wasserstein}.
Moreover, ~\citet{asencio2021similarity} proposes a similarity measure between Gaussian processes.
~\citet{delon2020wasserstein} introduce a Wasserstein distance on Gaussian mixture models.
In addition to distribution-based distance, ~\citet{blondel2021differentiable} introduce a differentiable variant of Dynamic Time Warping (named soft-DTW) as a divergence function.
Although soft-DTW measures the similarity between time series, it is a divergence function rather than a real similarity measure, as divergence functions do not satisfy the triangle inequality.

Unlike all existing work, this paper models time-series datasets as multivariate normal distributions and let Wasserstein similarity measure the distribution distances to represent dataset (dis)similarities.

\section{Similarity modeling between Time-series Datasets}
Our approach models the (dis)similarity between two time-series datasets by applying the Wasserstein distance to two distributions, each generalizing one of the datasets.
After problem formulation (Sec,~\ref{sec:problem-formulation}), we model time-series datasets as multivariate normal distributions (Sec.~\ref{sec:distribution-tsd}).
We propose using the Wasserstein distance to compute the dataset distance between distributions (Sec.~\ref{sec:distance-wass}).

\subsection{Problem Formulation}
\label{sec:problem-formulation}
Time-series datasets often contain time series of vastly different number of time steps or lengths.
However, in time-series forecasting tasks, a sampling window is typically leveraged to construct data samples for training and testing, resulting in time-series samples of the same length.
Hence, after the sampling process, we formulate our problem as measuring the (dis)similarity between the resulting time-series datasets, each containing time series of equal length.

Let $\mX$ and $\mY$ denote two time-series datasets. 
Both datasets contain $N$ time series of $L$ time steps: $\mX = \brac{\vx_1 \vx_2 \cdots \vx_N}\transpose \in \dom{N}{L}$ and $\mY = \brac{\vy_1 \vy_2 \cdots \vy_N}\transpose \in \dom{N}{L}$.
Here $\vx_1, \vx_2, \ldots, \vx_N, \vy_1, \vy_2, \ldots, \vy_N$ each denotes a time series.
Our objective is to propose a function $d$ that measures the (dis)similarity between the two time-series datasets, denoted as $d\pare{\mX, \mY}$.

\subsection{Distributions of Time-series Datasets}
\label{sec:distribution-tsd}

Assume $N$ random vectors $\bmX_1, \bmX_2, \ldots, \bmX_N \in \domA{L}$ that are  independent and identically distributed (i.i.d.), and are sampled from a multivariate normal distribution $\MVN\pare{\vmu, \mSigma}$.
Here $\vmu\in\domA{L}$ denotes the mean vector and $\mSigma=\pare{\sigma_{ij}} \in \dom{L}{L}$ denotes the covariance matrix of the distribution.
The unknown parameters $\pare{\vmu, \mSigma}$ are estimated as $\pare{\hat{\vmu}, \hat{\mSigma}}$, by applying maximum likelihood estimation to the observed samples $\bmX_1, \bmX_2, \ldots, \bmX_N$,
\begin{align}
    \hat{\vmu}_{\bmX}    &= \frac{1}{N} \sum_{i=1}^N \bmX_i &
    \hat{\mSigma}_{\bmX} &= \frac{1}{N} \pare{\bmX-\hat{\vmu}_{\bmX}}\transpose \pare{\bmX-\hat{\vmu}_{\bmX}}
\end{align}
In this paper, we assume that time series $\vx_1, \vx_2, \ldots, \vx_N$, $\vy_1, \vy_2, \ldots, \vy_N$ in the datasets $\mX$ and $\mY$ adhere to the i.i.d. requirement.
We also assume signals do not cancel each other out in real-world datasets when averaged over.
By respectively letting $\bmX_i=\vx_i$ and $\bmX_i=\vy_i$, we model each dataset as a multivariate normal distribution, denoted by $\calD_X$ and $\calD_Y$,
\begin{align}
    \calD_X &= \MVN\pare{\hat{\vmu}_{\bmX}, \hat{\mSigma}_{\bmX}} &
    \calD_Y &= \MVN\pare{\hat{\vmu}_{\bmY}, \hat{\mSigma}_{\bmY}} 
\end{align}
$\MVN$ modeling gives multifold benefits:
(1) it standardizes time series of potentially variable lengths to a common length;
(2) it enables the use of the Wasserstein distance;
(3) it aligns with the configurations of many tasks and applications, such as time-series forecasting.

\subsection{Wasserstein Distance on Time-series Datasets}
\label{sec:distance-wass}
We define $d\pare{\cdot}$ as the Wasserstein distance function between the distributions of two datasets,
\begin{align}
    d\pare{\mX, \mY} = d_\Ws\pare{\calD_X, \calD_Y}
\end{align}
Wasserstein distance is a metric that measures the distance between two probability distributions~\cite{panaretos2020invitation,dowson1982frechet}.
It is most used in \textit{optimal transport} problems to determine the minimum distance between two distributions based on the least costive couplings.
The $p$-Wasserstein distance is defined by
\begin{align}
W_p(\mu, \nu)=\min_{\gamma \in \Gamma_{\mu, \nu}}\pare{\int \|x-y\|^p\, \gamma(dx, dy)}^{1/p}
\end{align}
where $\mu, \nu \in \domA{d}$ are two probability measures, and $\Gamma$ denotes the set of couplings of $\mu$ and $\nu$.
~\cite{{chewi2024statistical}} provides a comprehensive summary of various Wasserstein distances.

In this paper, we primarily focus on the Wasserstein distance $d_\Ws$ between two multivariate normal distributions, $\calD_X$ and $\calD_Y$, which is defined by
\begin{align}
    d_\Ws^2\pare{\calD_X, \calD_Y} &= \norm{\hat{\vmu}_{\bmX} - \hat{\vmu}_{\bmY}}^2 
    + \trace\pare{\hat{\mSigma}_{\bmX} + \hat{\mSigma}_{\bmY} - 2\sqrt{\hat{\mSigma}_{\bmX} \hat{\mSigma}_{\bmY}}} \label{eq:Wd}
\end{align}
where the first term measures the distance between the mean vectors, while the second term captures the difference between the covariance matrices~\cite{okano2024distribution}.
Intuitively, Eq.~\ref{eq:Wd} suggests that for two datasets to be close to each other, they must not only have similar mean vectors but also similar covariate matrices.
Note that Eq.~\ref{eq:Wd} is a metric~\cite{dowson1982frechet}.

\begin{table}[t]
\caption{A summary of dataset statistics.}
\label{tab:dataset-details}
\small
\centering
\resizebox{\linewidth}{!}{
\begin{tabular}{@{}l@{\hspace{1pt}}rrrrr|l@{\hspace{1pt}}rrrrr@{}}
\toprule
Gluonts Dataset & Frequency & Mean $\pm$ Std. & Min & Median & Max & Gluonts Dataset & Frequency & Mean $\pm$ Std. & Min & Median & Max\\
\midrule
\texttt{kdd\_cup\_2018\_wm}    & H      & $39.8 \pm 56.1$   & $-14.0$ & $20.0$  & $3.0k$   & \texttt{traffic}             & H      & $0.1 \pm 0.1$     & $0.0$ & $0.0$   & $0.7$    \\
\texttt{rideshare\_wm}         & H      & $6.1 \pm 11.0$    & $0.0$   & $1.0$   & $97.5$   & \texttt{traffic\_nips}       & H      & $0.1 \pm 0.0$     & $0.0$ & $0.0$   & $1.0$    \\
\texttt{taxi\_30min}           & 30 min & $8.8 \pm 9.5$     & $0.0$   & $6.0$   & $235.0$  & \texttt{solar-energy}        & H      & $40.6 \pm 62.0$   & $0.0$ & $0.1$   & $501.9$  \\
\texttt{weather}               & D      & $14.7 \pm 11.2$   & $-38.4$ & $14.9$  & $369.6$  & \texttt{solar\_nips}         & H      & $40.9 \pm 62.3$   & $0.0$ & $0.1$   & $506.6$  \\
\texttt{temperature\_rain\_wm} & D      & $7.1 \pm 15.3$    & $-7.7$  & $0.0$   & $750.8$  & \texttt{solar\_10\_minutes}  & 10 min & $8.2 \pm 10.9$    & $0.0$ & $2.8$   & $84.4$   \\
\texttt{exchange\_rate}        & 2 W    & $0.7 \pm 0.5$     & $0.0$   & $0.7$   & $2.1$    & \texttt{kaggle\_web\_wm}     & D      & $1.0k \pm 13.6k$  & $0.0$ & $149.0$ & $2.0m$   \\
\texttt{exchange\_rate\_nips}  & 2 W    & $0.7 \pm 0.5$     & $0.0$   & $0.7$   & $2.1$    & \texttt{kaggle\_web\_weekly} & W      & $7.3k \pm 213.8k$ & $0.0$ & $984.0$ & $120.1m$ \\
\texttt{ett\_small\_15min}     & 15 min & $10.8 \pm 15.0$   & $-31.5$ & $5.2$   & $107.9$  & \texttt{uber\_tlc\_hourly}   & H      & $12.9 \pm 28.6$   & $0.0$ & $2.0$   & $609.0$  \\
\texttt{ett\_small\_1h}        & H      & $10.7 \pm 15.0$   & $-31.5$ & $5.0$   & $107.9$  & \texttt{wiki2000\_nips}      & D      & $3.8k \pm 9.5k$   & $0.0$ & $2.5k$  & $3.5m$   \\
\texttt{electricity\_weekly}   & W      & $434.1k \pm 2.3m$ & $0.0$   & $94.8k$ & $63.0m$  & \texttt{wiki-rolling\_nips}  & D      & $3.8k \pm 9.0k$   & $0.0$ & $2.5k$  & $2.5m$   \\
\texttt{electricity}           & H      & $2.6k \pm 14.9k$  & $0.0$   & $552.0$ & $749.1k$ & \texttt{nn5\_daily\_wm}      & D      & $18.5 \pm 9.7$    & $0.0$ & $16.6$  & $96.0$   \\
\texttt{electricity\_nips}     & H      & $622.4 \pm 3.6k$  & $0.0$   & $120.5$ & $168.1k$ & \texttt{tourism\_monthly}    & M      & $10.1k \pm 34.3k$ & $0.0$ & $1.3k$  & $1.2m$   \\
\texttt{electricity\_hourly}   & H      & $2.6k \pm 15.6k$  & $0.0$   & $547.0$ & $672.4k$ & \texttt{saugeenday}          & D      & $30.3 \pm 39.3$   & $2.3$ & $17.4$  & $640.0$  \\
\texttt{ercot}                 & H      & $4.8k \pm 4.7k$   & $498.4$ & $2.4k$  & $25.7k$  & \texttt{london\_meters\_wm}  & 30 min & $0.2 \pm 0.3$     & $0.0$ & $0.1$   & $8.0$    \\
\texttt{m4\_hourly}            & H      & $7.3k \pm 41.5k$  & $10.0$  & $55.0$  & $703.0k$ & \texttt{covid\_deaths}       & D      & $1.2k \pm 7.1k$   & $0.0$ & $10.0$  & $141.7k$ \\
\bottomrule
\end{tabular}
}
\end{table}

\begin{figure*}[t]
\centering
\includegraphics[width=0.48\linewidth]{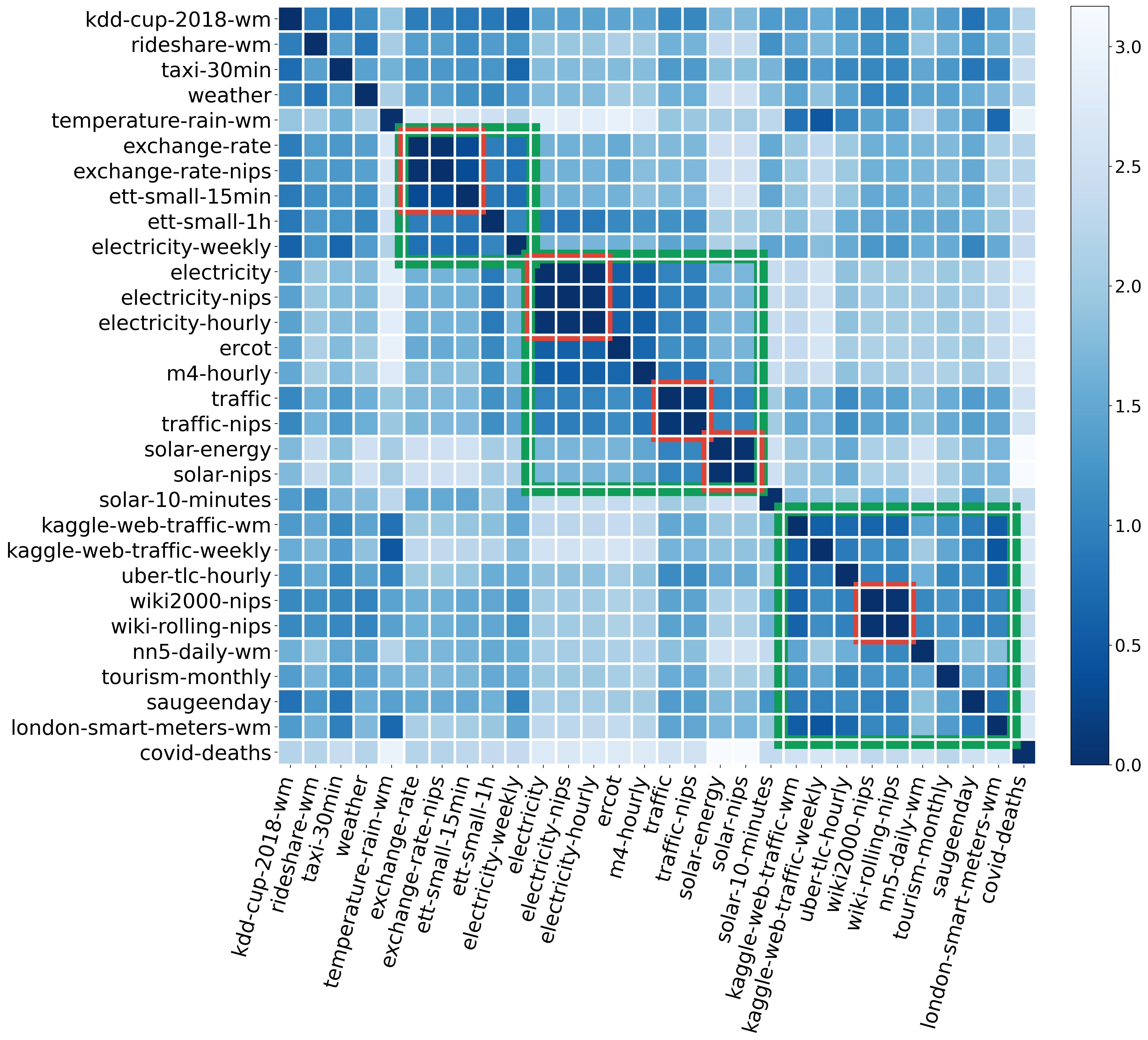}
\includegraphics[width=0.51\linewidth]{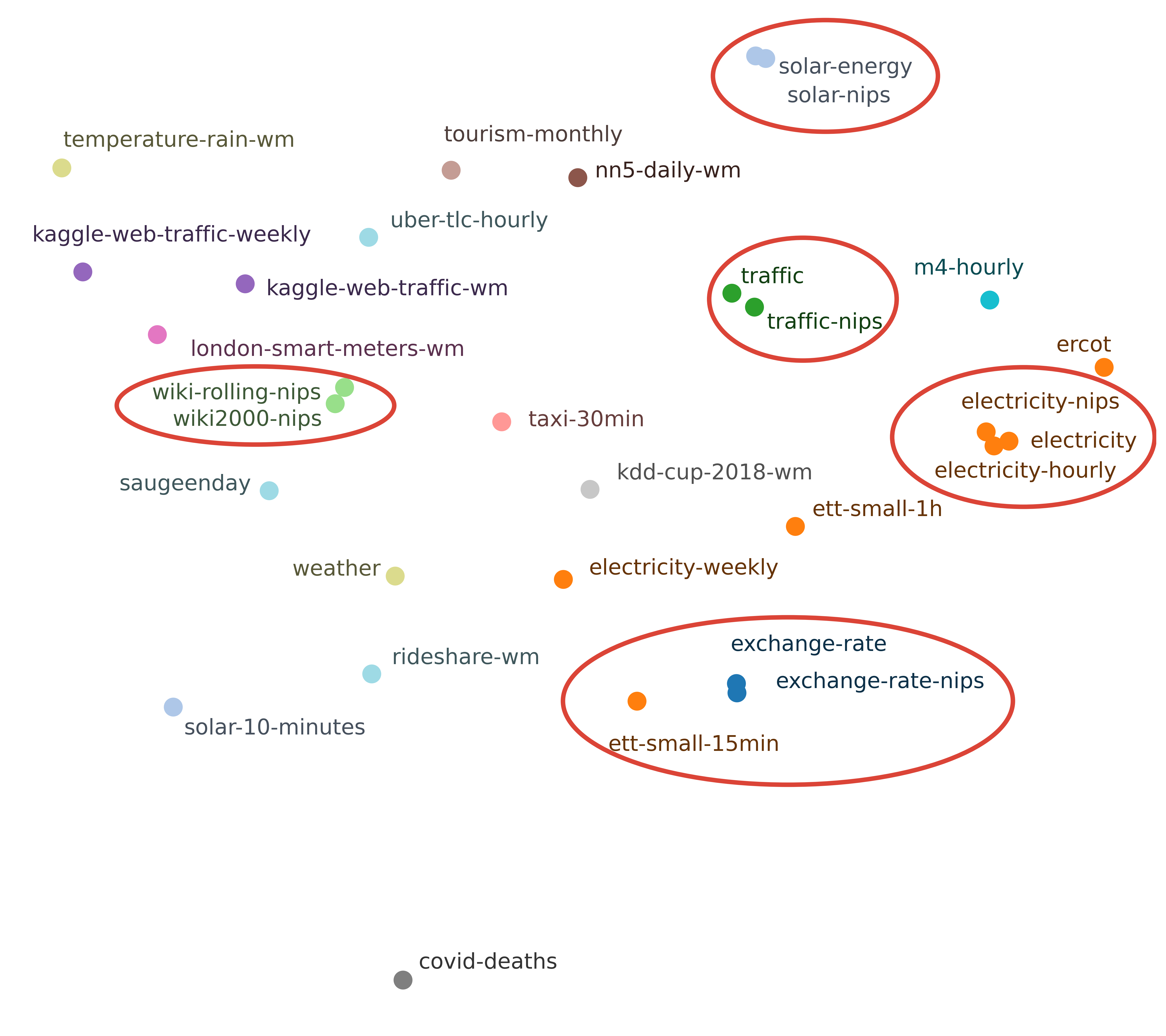}
\caption{
    (Left) A heatmap of pairwise \textit{Wasserstein distances} of thirty time-series datasets.
    The colorbar indicates the distance, where darker colors indicate higher similarities and closer distances.
    Wasserstein distance helps identify hierarchical clusters (highlighted in \textcolor{googlegreen}{green}) and subclusters (highlighted in \textcolor{googlered}{red}) of datasets.
    (Right) A visualization using the computed Wasserstein distances is rendered based on a force-directed graph layout algorithm.
    Datasets belonging to the same domain are pre-annotated in the same text colors.
    Identified subclusters by our proposed Wasserstein distance are highlighted within ellipses.
}
\label{fig:30datasets-Wasserstein}
\end{figure*}

\section{Experiments}
\label{sec:exp}
To validate the effectiveness of our proposed similarity modeling, 
we design experiments with visualization (Sec.~\ref{sec:visualization}), which show that the Wasserstein distance helps identify and group similar time-series datasets.
Comparisons of the Wasserstein distance with other distances (Sec.~\ref{sec:baseline-comp}) further strengthen our confidence in the approach.
Experiments with foundation models (Sec.~\ref{sec:transfer-lr}) show that the Wasserstein distance serves as a reliable indicator of model performance across various time-series datasets.
We also discuss time complexity and outlier datasets (Sec.~\ref{sec:bad-dataset}).

\noindent\textbf{\textit{Datasets.}}
We utilize $M = 30$ time-series datasets collected by Gluonts~\cite{alexandrov2019gluonts}.
A summary of data statistics is provided in Table~\ref{tab:dataset-details}.
Notably, the utilized datasets cover a wide range of granularities, including hours, ten minutes, fifteen minutes, thirty minutes, days, weeks, biweeks, and months.
Moreover, values in these datasets span a wide range, from close to zero up to $120$ million.
We apply min-max normalization to scale the data to the range of $\brac{0, 1}$.
When constructing data samples, we draw a sufficient number of samples to achieve an accurate distribution estimate, while ensuring the sample size remains manageable to keep the runtime efficient.
In the extremely rare case where a sampled time series is constant, we resample the time series to ensure variability.
Hence, we randomly sample $N=20,000$ data points from each dataset, where each data point consists of $L=48$ time steps.
As a result, each dataset yields a $N\times L$ matrix.

\subsection{Visualization}
\label{sec:visualization}
This section uses visualization to demonstrate the effectiveness of our similarity modeling in mining similar datasets.
After estimating the multivariate normal distribution parameters for each of the $M$ dataset, we compute the pairwise Wasserstein distance between all pairs of datasets.
This results in a $M\times M$ Wasserstein distance matrix, as depicted in Fig.~\ref{fig:30datasets-Wasserstein} (left).
The color in the matrix represents the distance between the corresponding datasets in each row and column, with darker colors indicating higher similarities (i.e., smaller distances).
From Fig.~\ref{fig:30datasets-Wasserstein} (left), we observe that Wasserstein distance helps identify clusters (highlighted in green) and subclusters (highlighted in red) of the studied time-series datasets, as highlighted along the diagonal of the matrix.
In other words, time-series datasets within each cluster or subcluster are close to one another,  as suggested by the Wasserstein distance.
For example, the bottom right $2 \times 2$ red square contains the pairwise distance of two datasets, \texttt{wiki2000-nips} and \texttt{wiki-rolling-nips}.

To provide a more intuitive understanding of the (dis)similarities between datasets, a graph is constructed from the computed pairwise distances.
We employ the Kamada-Kawai force-directed algorithm from Python's NetworkX library to position datasets in the graph, where each node represents a time-series dataset, and the edge lengths between nodes are proportional to their Wasserstein distances~\cite{kamada1989algorithm}.
The resulting graph is depicted in Fig.~\ref{fig:30datasets-Wasserstein} (right).
We leverage expert knowledge to color code time-series datasets in the same domains with the same colors.
Notably, Fig.~\ref{fig:30datasets-Wasserstein} (right) shows that our proposed similarity measure groups similar datasets together.

As highlighted within the red ellipses, the time-series datasets for solar, traffic, electricity, exchange rate, and Wikipedia visits are each tightly grouped together.
Additionally, isolated time-series datasets such as \texttt{covid-deaths} and \texttt{m4-hourly} are also identified by our method.

\subsection{Comparison with Other Distances}
\label{sec:baseline-comp}
To demonstrate the advantages of our approach, we compare the Wasserstein distance results with those of other similarity measures.
These baselines are categorized into two groups: the first group is based on the averaged time series, while the second group is based on clustering.

\subsubsection{Distances based on Averaged Time Series}
For baselines in the first group, we represent each time-series dataset with its average time-series $\hat{\vmu}$, and compute the inter-dataset similarities accordingly.
The first group contains two similarity measures, one with the Euclidean distance, while the other one with the Dynamic Time Warping (DTW).
The Euclidean similarity measure is the same as the square root of the first term in Eq.~\ref{eq:Wd},
\begin{align}
    d_{Euclidean}\pare{\mX, \mY} &= \norm{\hat{\vmu}_{\bmX} - \hat{\vmu}_{\bmY}}
\end{align}

\begin{figure*}[t]
\centering
\includegraphics[width=\linewidth]{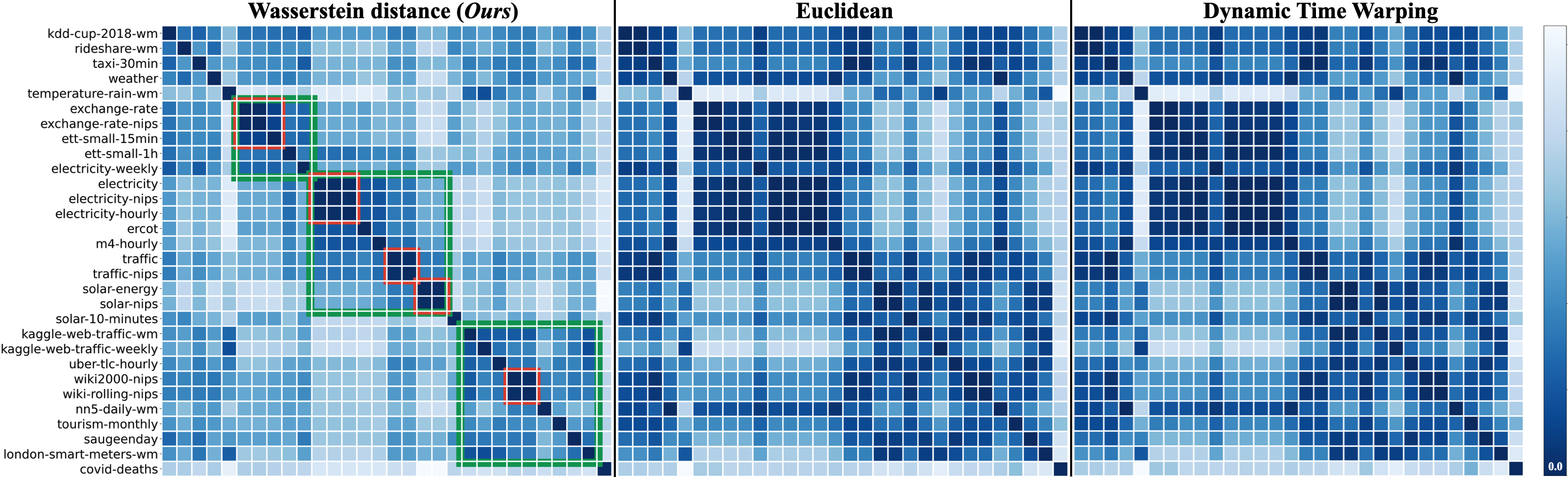}
\caption{
    A comparison of distance heatmaps of various approaches: 
    (Left) Our Wasserstein distance approach;
    (Middle) Euclidean distance;
    (Right) DTW distance.
    Each heatmap demonstrates the pairwise distances of $30$ datasets.
    Darker colors indicate higher similarities.
    The heatmap of Wasserstein distance is notably more informative with more refined blocks / clusters.
}
\label{fig:30datasets-Eucli-dtw}
\end{figure*}

DTW measures the similarity between two aligned time series~\cite{berndt1994using}.
We let $d_{DTW}$ denote the DTW similarity returned by the algorithm,
\begin{align}
    d_{DTW}\pare{\mX, \mY} &\coloneqq \DTW\pare{\hat{\vmu}_{\bmX}, \hat{\vmu}_{\bmY}}
\end{align}
Its computation involves dynamic programming,
\begin{align}
    \DTW(\hat{\vmu}_{\bmX}, \hat{\vmu}_{\bmY}) &= D_{L,L}, &
    D_{i, j} &= \abs{\hat{\vmu}_{{\bmX}_i}-\hat{\vmu}_{{\bmY}_j}} + \min\pare{D_{i-1, j-1}, D_{i, j-1}, D_{i-1, j}}
\end{align}
We construct the $M\times M$ distance matrices using both measures, as shown in Fig.~\ref{fig:30datasets-Eucli-dtw}.
Compared to the matrix derived from the Wasserstein distance, the matrices generated using the Euclidean distance and the DTW similarity are less informative.
Not only are they similar to each other, but they also exhibit coarser clustering, as shown by the large blocks along the diagonal.

\begin{figure*}[t]
\centering
\includegraphics[width=\linewidth]{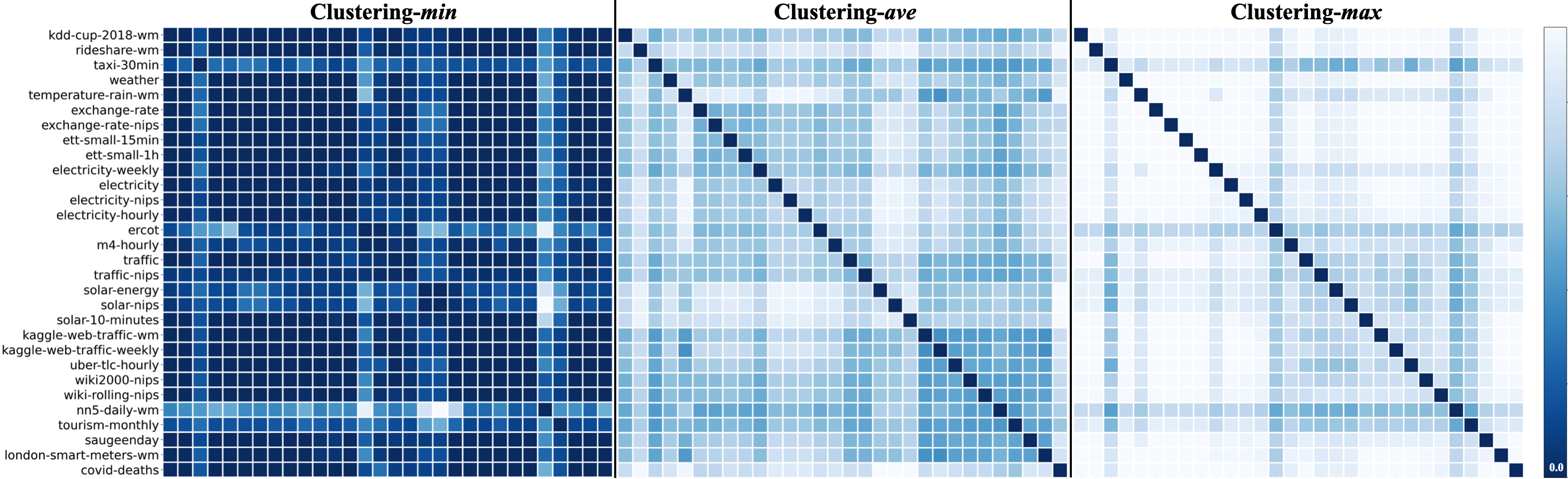}
\caption{
    Additional heatmaps of clustering-based distances for comparison:
    (Left) minimum distance;
    (Middle) average distance;
    (Right) maximum distance.
    Darker colors show closer distances.
    Note that Wasserstein distance in Fig.~\ref{fig:30datasets-Wasserstein} is more informative than any of these.
}
\label{fig:30datasets-mam}
\end{figure*}

\begin{figure*}[t]
\centering
\includegraphics[width=0.49\linewidth]{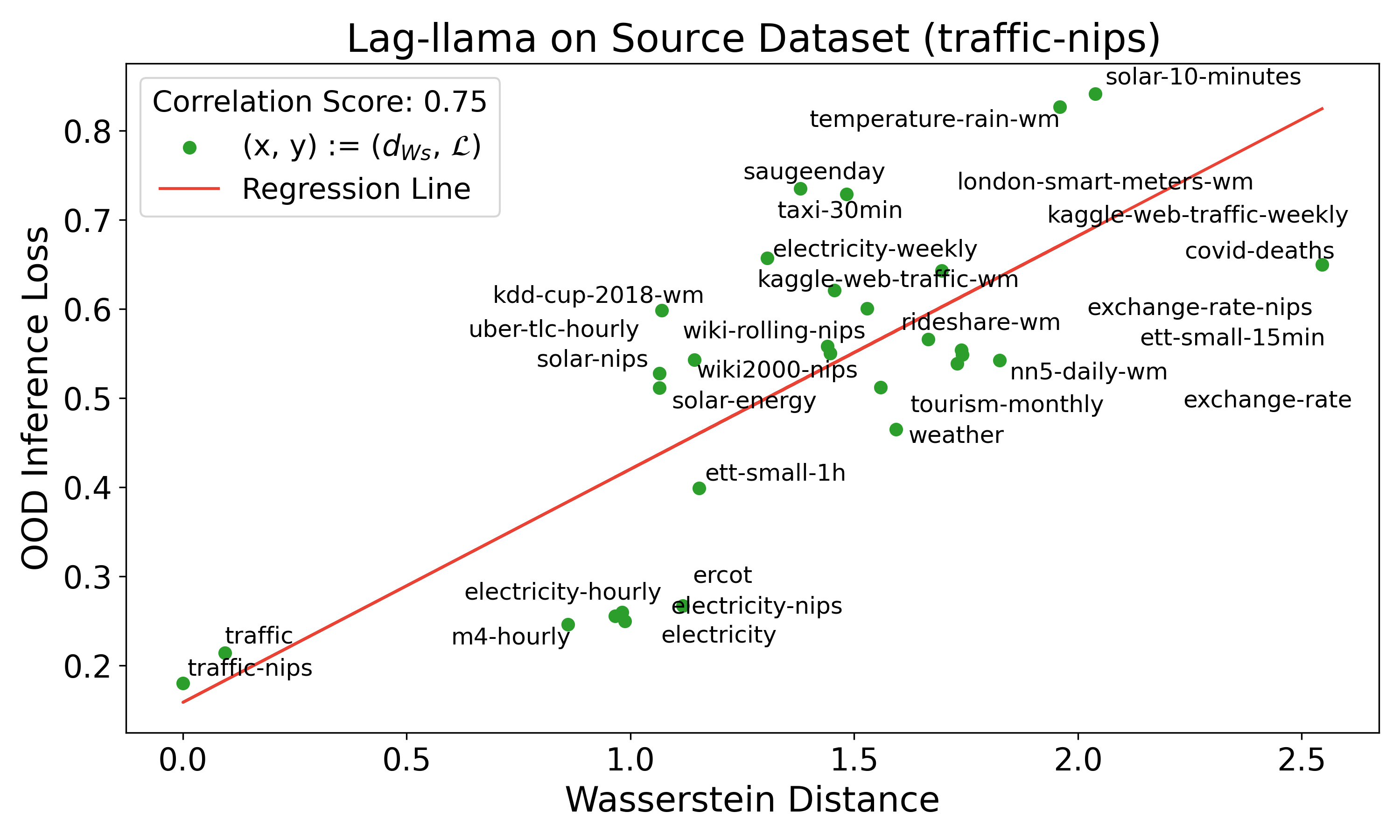}
\includegraphics[width=0.49\linewidth]{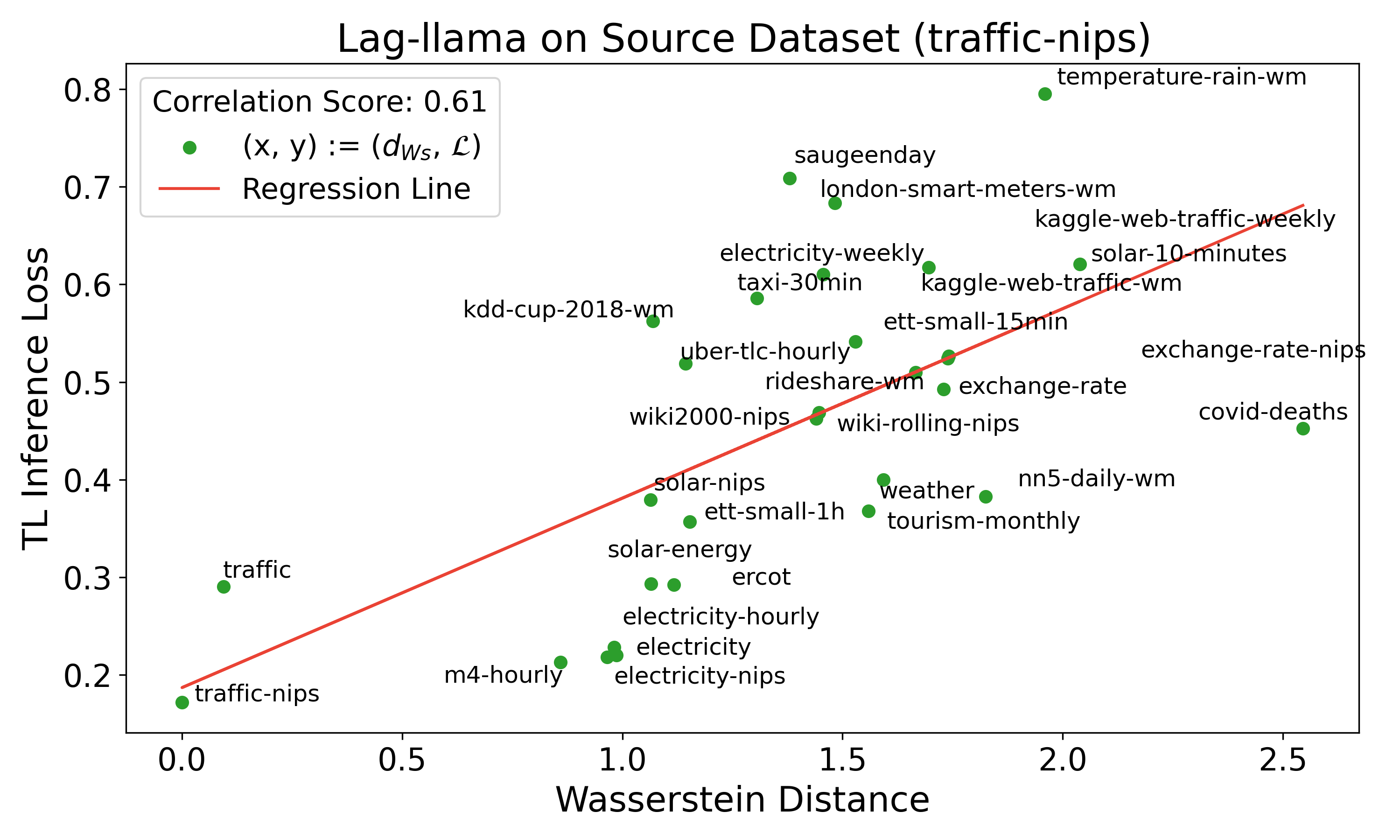}
 \rule{\linewidth}{0.5pt} 
\includegraphics[width=0.49\linewidth]{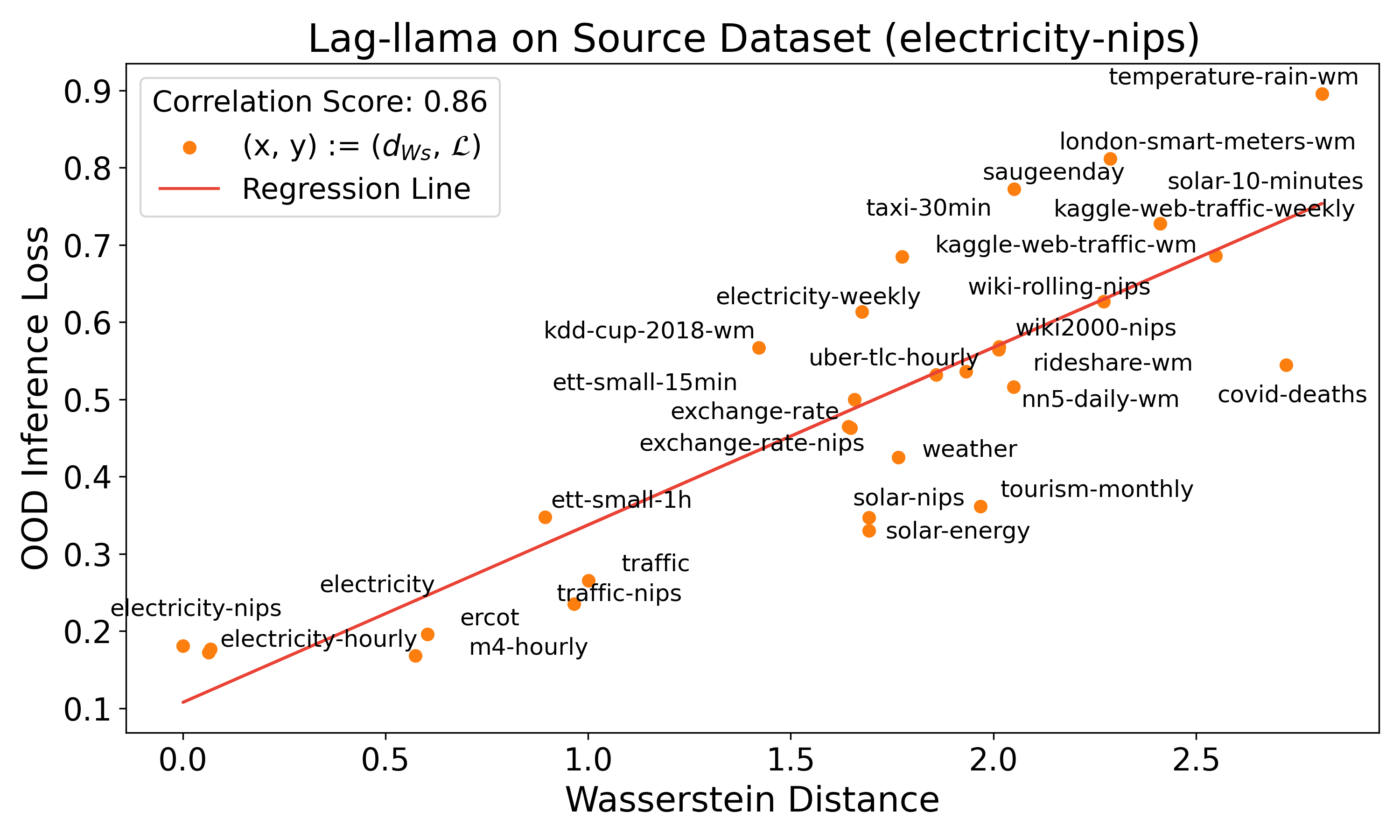}
\includegraphics[width=0.49\linewidth]{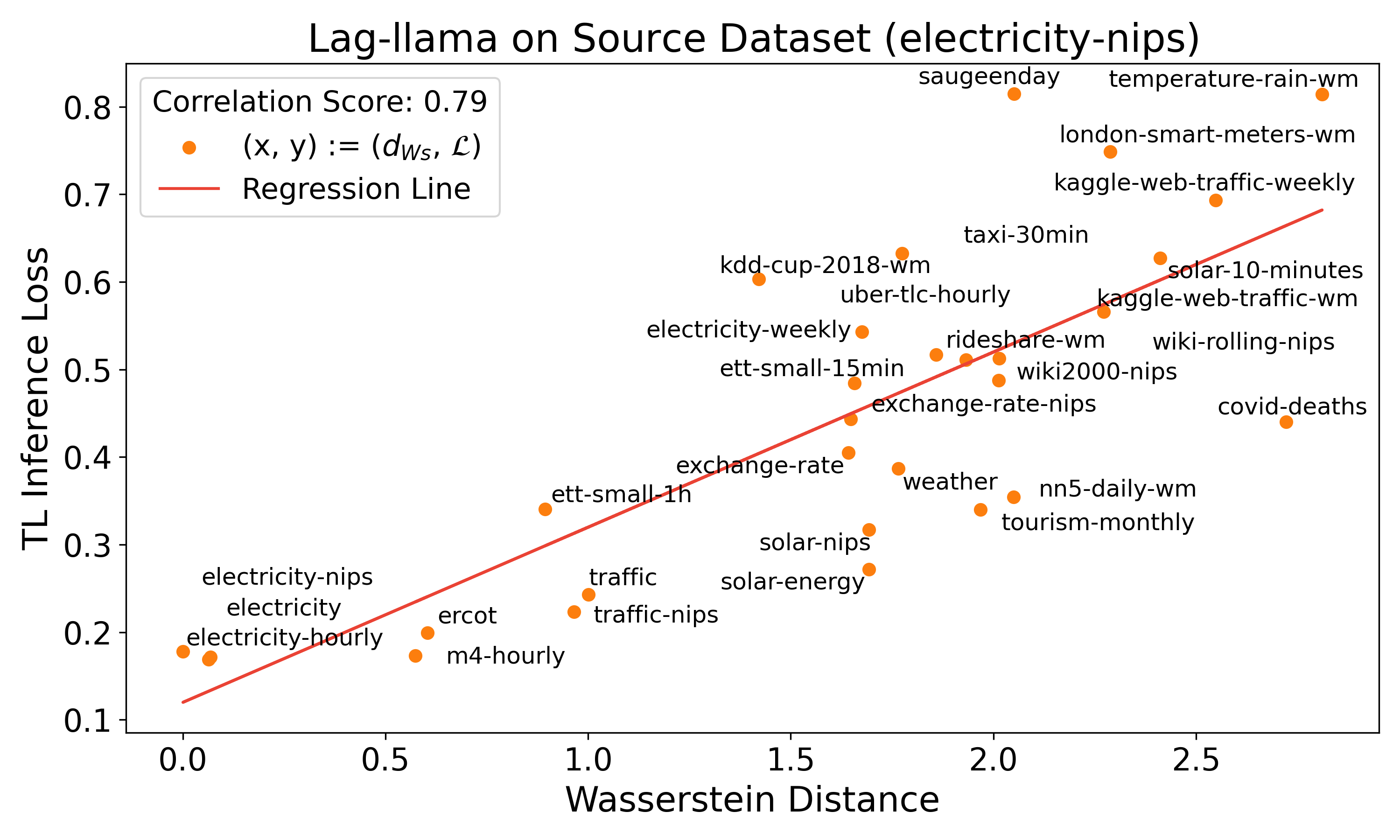}
\caption{
Out-Of-Distribution (OOD) and Transfer Learning (TL) inference losses of \textit{Lag-llama} trained on \texttt{traffic-nips} (top) and \texttt{electricity-nips} (bottom).
Each dot represents a target dataset, where the x-axis represents the Wasserstein distance between the source dataset and the target dataset, while the y-axis represents the OOD or TL inference loss on the target dataset.
A strong correlation is observed between the inference loss and the Wasserstein distance, suggesting that the Wasserstein distance is an informative indicator of inference loss.
}
\label{fig:4datasets-Wasserstein-compared-to-losses}
\end{figure*}

\subsubsection{Dataset Distance based on Clustering}
Cluster-based distances treat each time-series dataset as a cluster containing time series nodes.
In the context of clustering, the minimum, average, and maximum distances correspond to the single-linkage clustering, unweighted average linkage clustering, and complete-linkage clustering, respectively.
We compute the minimum, average, and maximum pairwise distances between any two datasets:
\begin{align}
    d_{min / max}\pare{\mX, \mY} &= {\min / \max}_{\vx\in\bmX,\vy\in\bmY}\norm{\vx-\vy}, &
    d_{avg}\pare{\mX, \mY} &= \frac{1}{|\mX|}\frac{1}{|\mY|}\sum_{\vx\in\bmX,\vy\in\bmY}\norm{\vx-\vy} \nonumber
\end{align}
The resulting three $M\times M$ distance matrices are depicted in Fig.~\ref{fig:30datasets-mam}.
Most distances from the minimum distance are close to zero, as shown by the dominant dark colors in Fig.~\ref{fig:30datasets-mam} (left).
In contrast to the minimum distance, most distances from the maximum distance are large, as shown by the dominant white colors in Fig.~\ref{fig:30datasets-mam} (right).
In either cases, there is little information available in terms of what datasets are similar.
The distance matrix based on average distance gives slightly better results where one large cluster of datasets is identified, as highlighted in the bottom right part of Fig.~\ref{fig:30datasets-mam} (middle).

Overall, our Wasserstein distance approach is more advantageous than these other distances as Wasserstein distance provides more information of dataset similarity.

\begin{figure*}[t]
\centering
\includegraphics[width=0.49\linewidth]{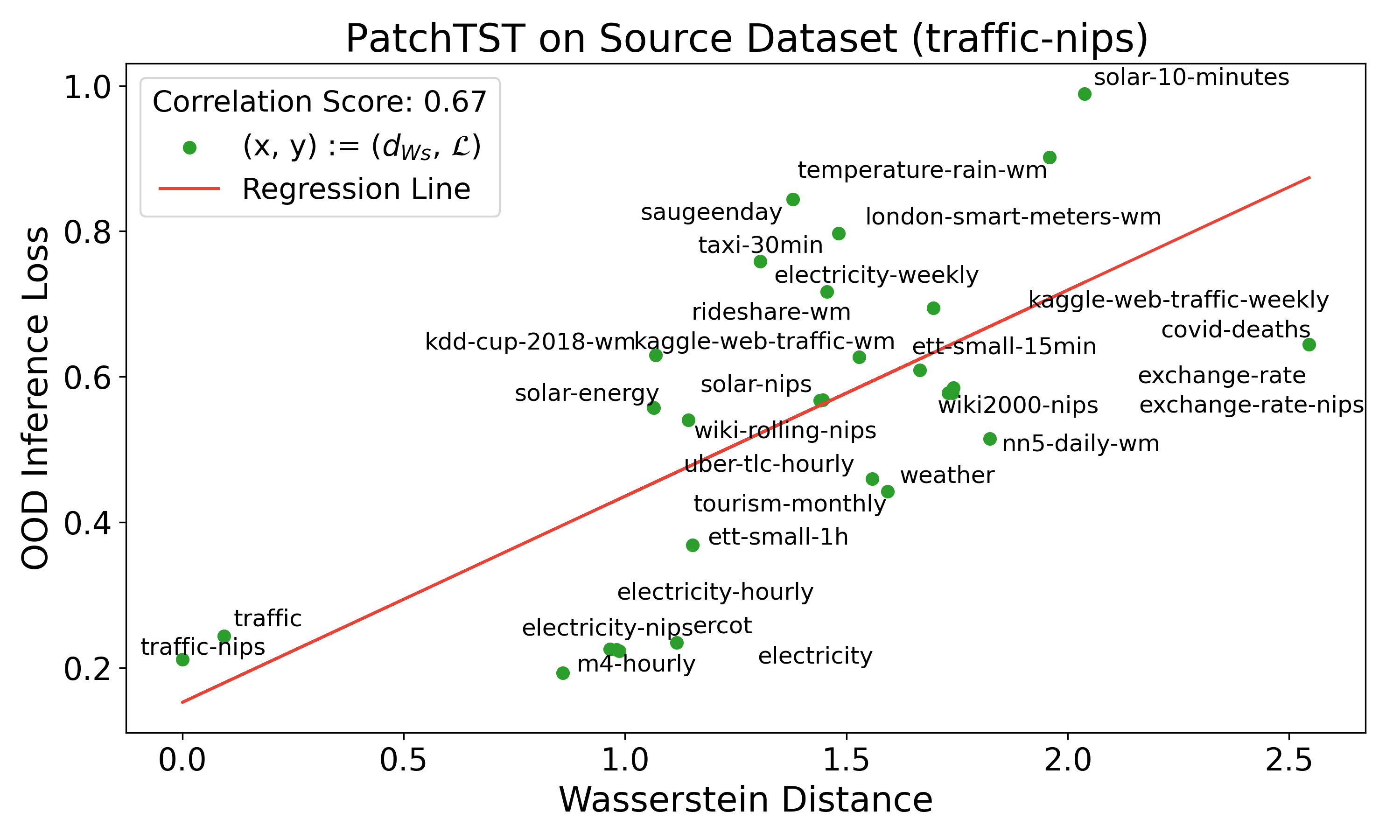}
\includegraphics[width=0.49\linewidth]{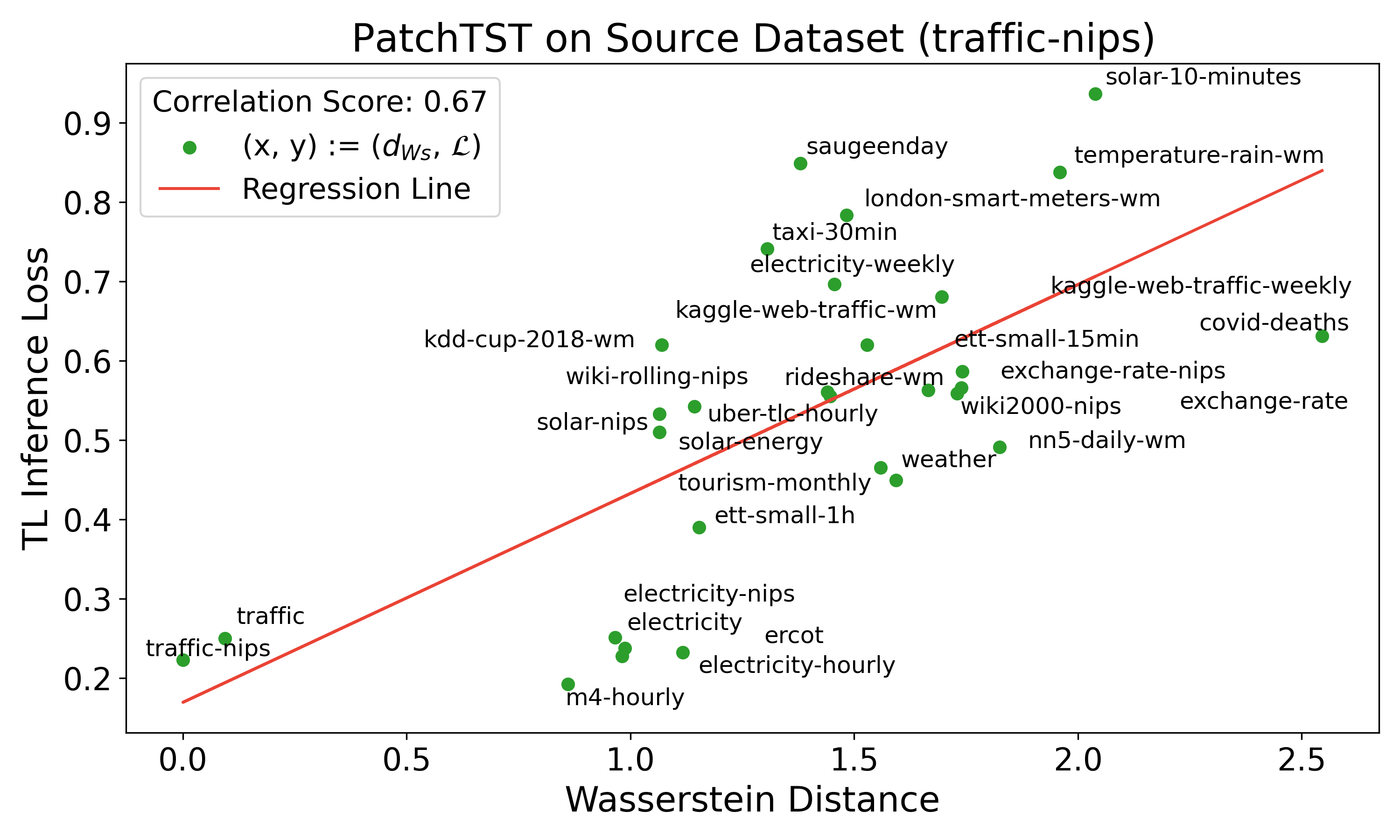}
 \rule{\linewidth}{0.5pt} 
\includegraphics[width=0.49\linewidth]{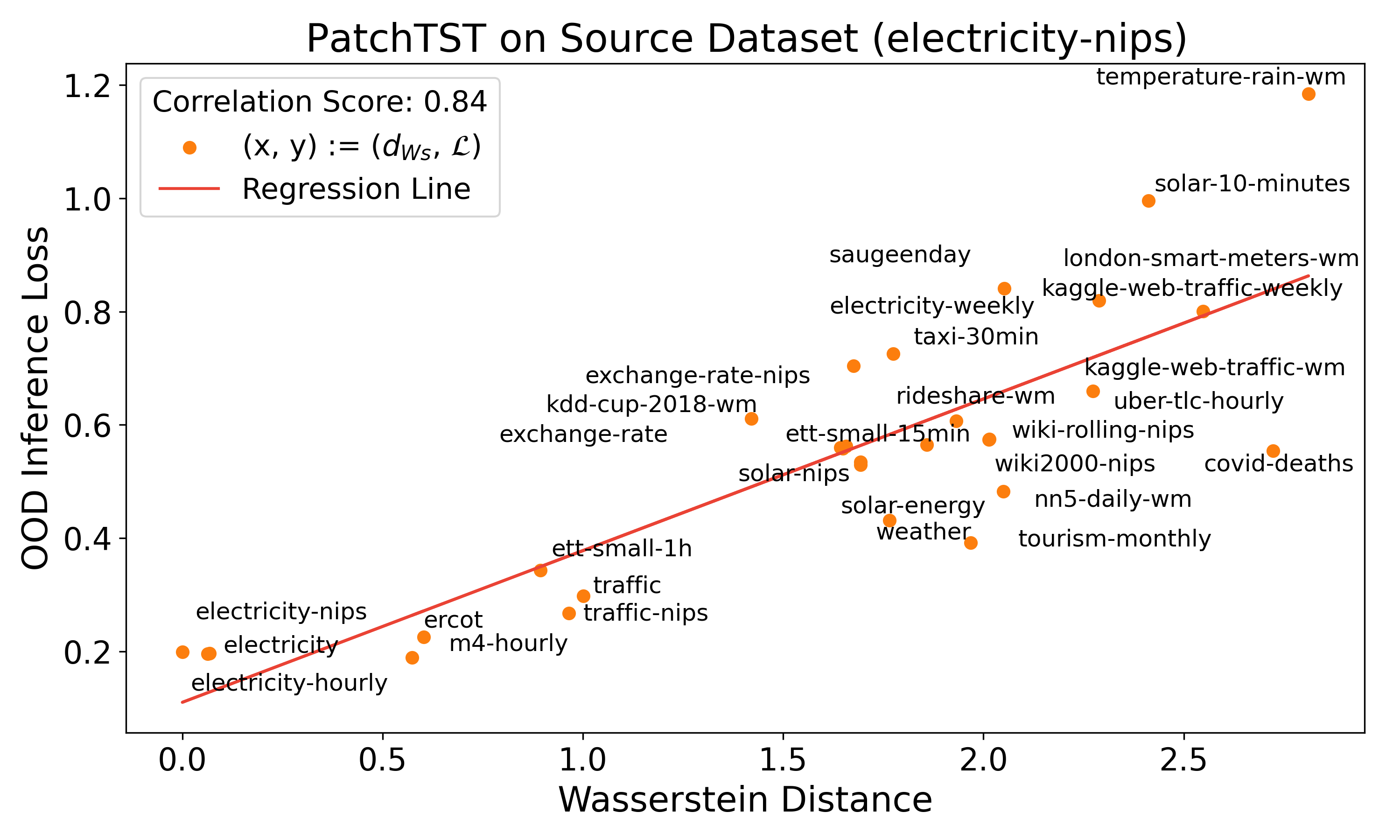}
\includegraphics[width=0.49\linewidth]{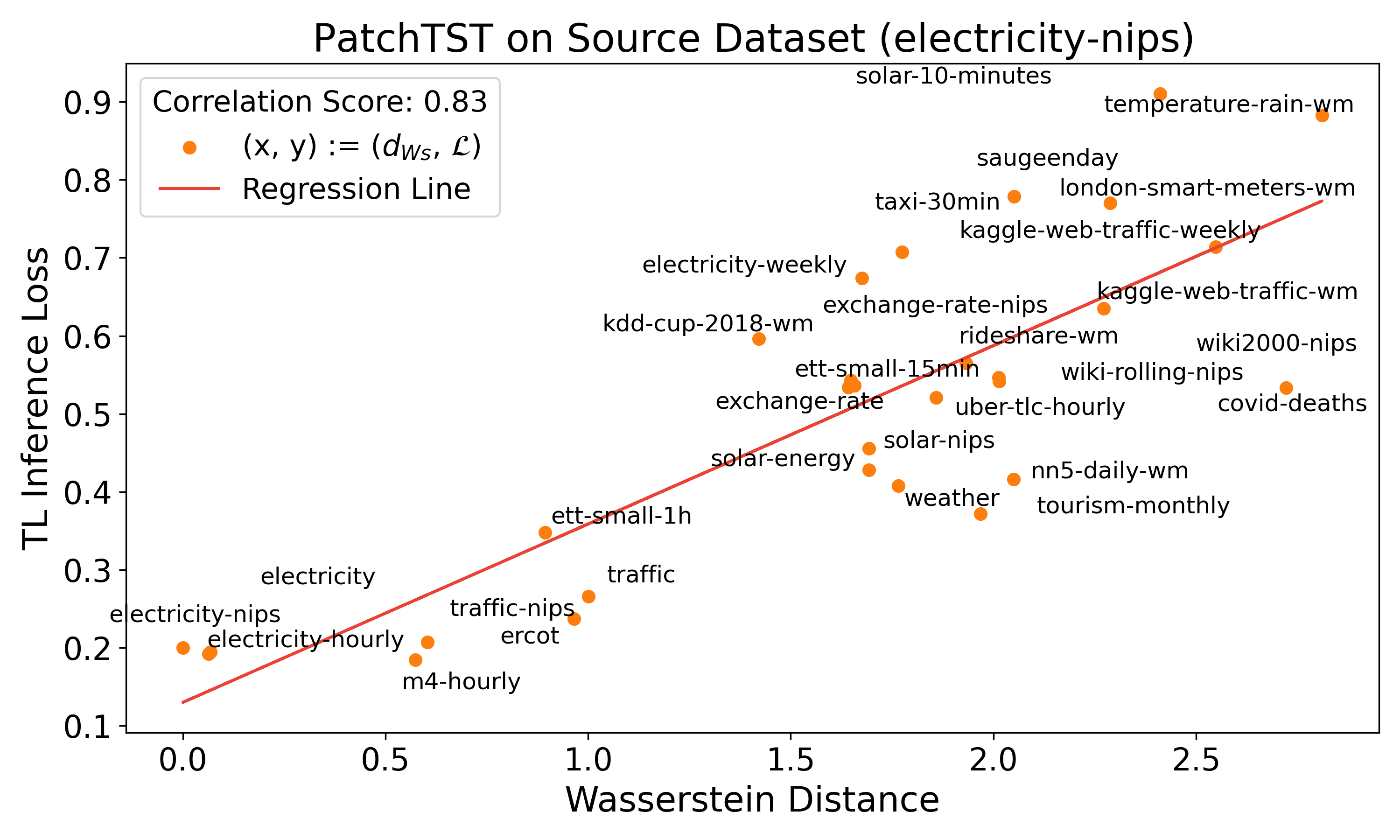}
\caption{
Out-Of-Distribution (OOD) and Transfer Learning (TL) inference losses of \textit{PatchTST} trained on \texttt{traffic-nips} (top) and \texttt{electricity-nips} (bottom).
Each dot represents a target dataset, where the x-axis represents the Wasserstein distance between the source dataset and the target dataset, while the y-axis represents the OOD or TL inference loss on the target dataset.
A strong correlation is observed between the inference loss and the Wasserstein distance, suggesting that the Wasserstein distance is an informative indicator of inference loss.
}
\label{fig:TST4datasets-Wasserstein-compared-to-losses}
\end{figure*}

\subsection{Out-of-distribution Evaluation and Transfer Learning}
\label{sec:transfer-lr}
Recall that we are originally motivated by the question: \textit{How can we estimate the performance of foundation models on target datasets if the models have never trained on these datasets?}
To validate that the Wasserstein distance serves as a performance indicator, this section designs experiments with time-series forecasting foundation models.

We utilize two time-series foundation models, namely \textit{Lag-llama}~\cite{rasul2024lag} and \textit{PatchTST}~\cite{Yuqietal-2023-PatchTST}.
We design experiments in two different setups, Out-Of-Distribution (OOD) and Transfer Learning (TL).
In the OOD setup, we train from scratch a model on each individual source dataset, and test on each target dataset.
The TL setup is the same as OOD except we finetune on each target dataset before we test on it.
Hence, for the $M=30$ datasets in our study, we have $M^2$ experiments covering each source-target dataset pair.
Experimental details are provided in the appendix~\ref{sec:append-exp-details}.

For the sake of clarity, we illustrate results of \textit{Lag-llama} on two representative datasets, \texttt{traffic-nips} and \texttt{electricity-nips}.
Complete results are provided in the appendix~\ref{sec:append-complete-results}.
Each scatter plot in Fig.~\ref{fig:4datasets-Wasserstein-compared-to-losses} contains $M=30$ dots that depict the relationships between inference loss and Wasserstein distance.
The values on the x-axis represent the Wasserstein distances to the source dataset, while the values on the y-axis represent the inference loss on the target datasets after training on the source dataset, for both OOD and TL setups.
Since \textit{Lag-llama} is a probabilistic model, we compute the mean weighted quantile loss.

From the figure we have the following observations:
(1) For both datasets, strong correlations ($> 0.60$) exist between the inference loss and the Wasserstein distance, indicating that the Wasserstein distance provides an informative estimate of model performance on the target datasets.
(2) The TL inference loss is generally lower than OOD for each target dataset, this is aligned with the intuition that finetuning improves prediction performance.
Results of \textit{PatchTST} on two representative datasets, \texttt{traffic-nips} and \texttt{electricity-nips} are given in Fig.~\ref{fig:TST4datasets-Wasserstein-compared-to-losses}.
Similar to the result from Fig.~\ref{fig:4datasets-Wasserstein-compared-to-losses}, We also observe a strong correlation between the inference loss and the Wasserstein distance, suggesting that the Wasserstein distance is an informative indicator of inference loss.

Overall, our approach servers as a performance indicator that helps researchers select the best time-series dataset as the source dataset for model finetuning.
Without our method, researchers would have to execute the actual finetuning to find the best source dataset.
Our approach significantly reduces computational cost and time, since computing the distance  using our method is much more efficient than conducting multiple finetuning runs.

\subsection{Time Complexity and Outlier Datasets}
\label{sec:bad-dataset}
The overall time complexity of our approach is $\bigO\pare{\pare{N+L} L^2}$, where $N$ is the number of samples and $L$ denotes the length of sampled time series.
Our approach is well-suited for scaling to massive datasets as $L$ is chosen such that $L \ll N$, which makes the computation grow linearly with the number of samples $N$.
A detailed analysis is included in the appendix~\ref{sec:append-time-complexity}.

\begin{figure}[t]
\centering
\includegraphics[width=0.6\linewidth]{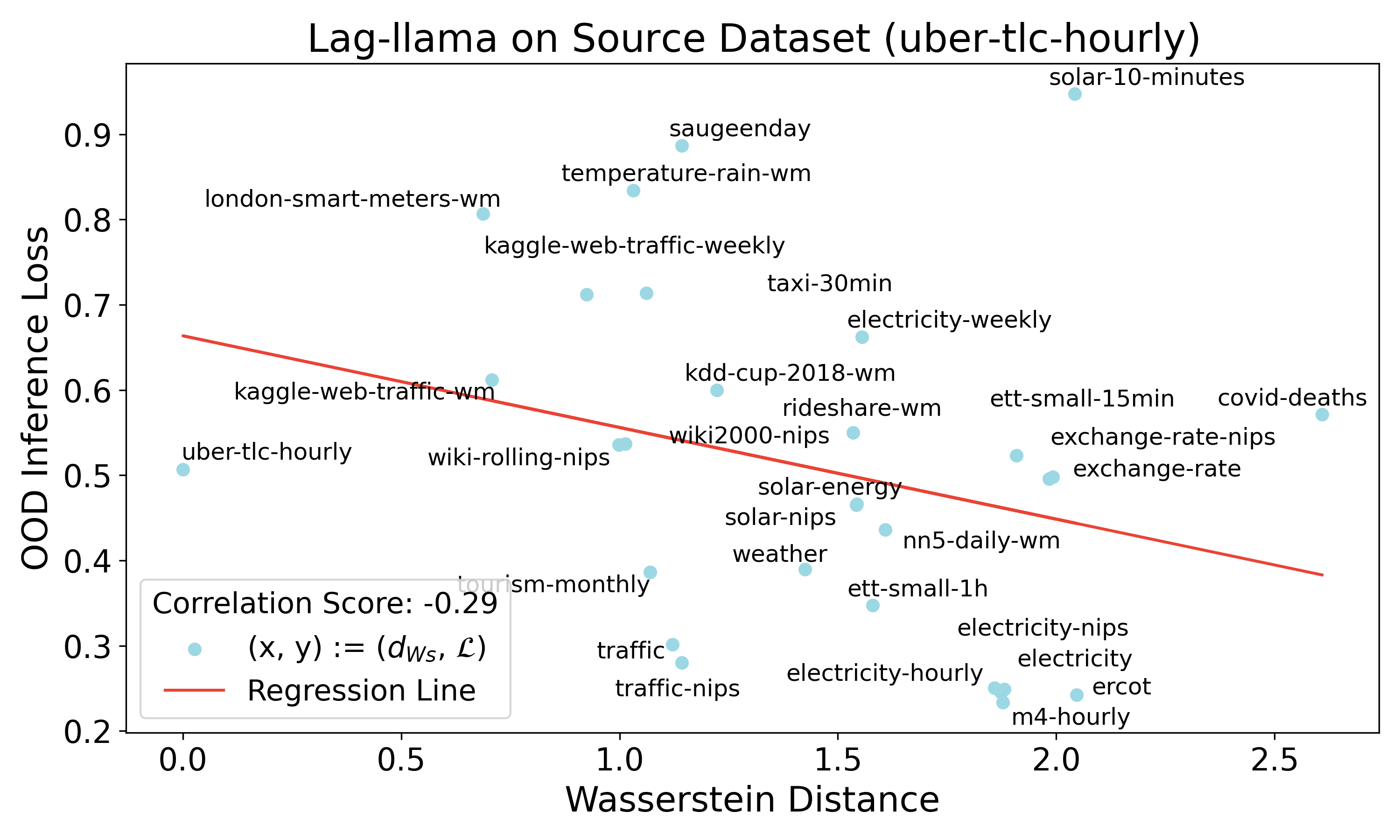}
\caption{
Out-Of-Distribution (OOD) inference loss of \textit{Lag-llama} trained on \texttt{uber-tlc-hourly}.
A negative correlation is observed between the inference loss and the Wasserstein distance.
}
\label{fig:uber-hourly-exception}
\end{figure}

Our experiments find that some datasets have a negative correlation between the inference loss on the target dataset and the Wasserstein distance.
One example is \texttt{uber-tlc-hourly}, as shown in Fig.~\ref{fig:uber-hourly-exception}.
Potential contributing factors include:
(1) deviations of the dataset from the MVN assumption; and
(2) limited relevance of the dataset for improving the performance of the backbone models during finetuning.
We suggest future research efforts be made to investigate the factors that decide the indication performance of Wasserstein distance.

\section{Conclusion}
\label{sec:conclusion}
In this paper, we propose modeling time-series datasets as multivariate normal distributions (MVNs) and propose using the Wasserstein distance to quantify the distance between MVNs, hence measuring the similarity between time-series datasets.
Through visualization and experiments, we show that the Wasserstein distance is an effective measure to help identify similar time-series datasets, which can be further used to estimate the foundation model performance on the target datasets.
We specifically demonstrate the effectiveness of our approach on two time-series forecasting foundation models over $30$ real-world datasets, in both set up of Out-Of-Distribution and Transfer Learning.
The time complexity and outlier datasets are also discussed, while complete experimental results are provided in the appendix.
In the future, we anticipate that our work paves the way for future time-series research in the areas of meta-learning, data distillation, and foundation models.

\subsection{Potential Constraints of Our Work}
Our proposed similarity measure relies on the i.i.d. and $\MVN$ assumption of the sampled data.
Further investigation may be required to assess its applicability to non-i.i.d. data.

\clearpage
\newpage

\bibliographystyle{plainnat}  
\bibliography{main}           

\begin{thebibliography}{50}
\providecommand{\natexlab}[1]{#1}
\providecommand{\url}[1]{\texttt{#1}}
\expandafter\ifx\csname urlstyle\endcsname\relax
  \providecommand{\doi}[1]{doi: #1}\else
  \providecommand{\doi}{doi: \begingroup \urlstyle{rm}\Url}\fi

\bibitem[Abanda et~al.(2019)Abanda, Mori, and Lozano]{abanda2019review}
Amaia Abanda, Usue Mori, and Jose~A Lozano.
\newblock A review on distance based time series classification.
\newblock \emph{Data Mining and Knowledge Discovery}, 33\penalty0 (2):\penalty0 378--412, 2019.

\bibitem[Adebayo et~al.(2021)Adebayo, Akinsola, Kirikkaleli, Bekun, Umarbeyli, and Osemeahon]{adebayo2021economic}
Tomiwa~Sunday Adebayo, Gbenga~Daniel Akinsola, Dervis Kirikkaleli, Festus~Victor Bekun, Sukru Umarbeyli, and Oseyenbhin~Sunday Osemeahon.
\newblock Economic performance of indonesia amidst co2 emissions and agriculture: a time series analysis.
\newblock \emph{Environmental Science and Pollution Research}, 28\penalty0 (35):\penalty0 47942--47956, 2021.

\bibitem[Alexandrov et~al.(2019)Alexandrov, Benidis, Bohlke-Schneider, Flunkert, Gasthaus, Januschowski, Maddix, Rangapuram, Salinas, Schulz, et~al.]{alexandrov2019gluonts}
Alexander Alexandrov, Konstantinos Benidis, Michael Bohlke-Schneider, Valentin Flunkert, Jan Gasthaus, Tim Januschowski, Danielle~C Maddix, Syama Rangapuram, David Salinas, Jasper Schulz, et~al.
\newblock Gluonts: Probabilistic time series models in python.
\newblock \emph{arXiv preprint arXiv:1906.05264}, 2019.

\bibitem[Alvarez-Melis and Fusi(2020)]{alvarez2020geometric}
David Alvarez-Melis and Nicolo Fusi.
\newblock Geometric dataset distances via optimal transport.
\newblock \emph{Advances in Neural Information Processing Systems}, 33:\penalty0 21428--21439, 2020.

\bibitem[Ansari et~al.(2024)Ansari, Stella, Turkmen, Zhang, Mercado, Shen, Shchur, Rangapuram, Arango, Kapoor, et~al.]{ansari2024chronos}
Abdul~Fatir Ansari, Lorenzo Stella, Caner Turkmen, Xiyuan Zhang, Pedro Mercado, Huibin Shen, Oleksandr Shchur, Syama~Sundar Rangapuram, Sebastian~Pineda Arango, Shubham Kapoor, et~al.
\newblock Chronos: Learning the language of time series.
\newblock \emph{arXiv preprint arXiv:2403.07815}, 2024.

\bibitem[Asencio~Mart{\'\i}n and Garrido-Merch{\'a}n(2021)]{asencio2021similarity}
Lucia Asencio~Mart{\'\i}n and Eduardo~C Garrido-Merch{\'a}n.
\newblock A similarity measure of gaussian process predictive distributions.
\newblock In \emph{Advances in Artificial Intelligence: 19th Conference of the Spanish Association for Artificial Intelligence, CAEPIA 2020/2021, M{\'a}laga, Spain, September 22--24, 2021, Proceedings 19}, pages 150--159. Springer, 2021.

\bibitem[Behme et~al.(2024)Behme, Galhotra, Beedkar, and Markl]{behme2024fainder}
Lennart Behme, Sainyam Galhotra, Kaustubh Beedkar, and Volker Markl.
\newblock Fainder: A fast and accurate index for distribution-aware dataset search.
\newblock \emph{Proceedings of the VLDB Endowment}, 17\penalty0 (11):\penalty0 3269--3282, 2024.

\bibitem[Belkin and Niyogi(2003)]{belkin2003laplacian}
Mikhail Belkin and Partha Niyogi.
\newblock Laplacian eigenmaps for dimensionality reduction and data representation.
\newblock \emph{Neural computation}, 15\penalty0 (6):\penalty0 1373--1396, 2003.

\bibitem[Benidis et~al.(2022)Benidis, Rangapuram, Flunkert, Wang, Maddix, Turkmen, Gasthaus, Bohlke-Schneider, Salinas, Stella, et~al.]{benidis2022deep}
Konstantinos Benidis, Syama~Sundar Rangapuram, Valentin Flunkert, Yuyang Wang, Danielle Maddix, Caner Turkmen, Jan Gasthaus, Michael Bohlke-Schneider, David Salinas, Lorenzo Stella, et~al.
\newblock Deep learning for time series forecasting: Tutorial and literature survey.
\newblock \emph{ACM Computing Surveys}, 55\penalty0 (6):\penalty0 1--36, 2022.

\bibitem[Berndt and Clifford(1994)]{berndt1994using}
Donald~J Berndt and James Clifford.
\newblock Using dynamic time warping to find patterns in time series.
\newblock In \emph{Proceedings of the 3rd international conference on knowledge discovery and data mining}, pages 359--370, 1994.

\bibitem[Blondel et~al.(2021)Blondel, Mensch, and Vert]{blondel2021differentiable}
Mathieu Blondel, Arthur Mensch, and Jean-Philippe Vert.
\newblock Differentiable divergences between time series.
\newblock In \emph{International Conference on Artificial Intelligence and Statistics}, pages 3853--3861. PMLR, 2021.

\bibitem[Chen and Eldardiry(2024)]{chen2024graph}
Hongjie Chen and Hoda Eldardiry.
\newblock Graph time-series modeling in deep learning: a survey.
\newblock \emph{ACM Transactions on Knowledge Discovery from Data}, 18\penalty0 (5):\penalty0 1--35, 2024.

\bibitem[Chen et~al.(2023)Chen, Rossi, Mahadik, Kim, and Eldardiry]{chen2023graph}
Hongjie Chen, Ryan~A Rossi, Kanak Mahadik, Sungchul Kim, and Hoda Eldardiry.
\newblock Graph deep factors for probabilistic time-series forecasting.
\newblock \emph{ACM Transactions on Knowledge Discovery from Data}, 17\penalty0 (2):\penalty0 1--30, 2023.

\bibitem[Chen et~al.(2025)Chen, Rossi, Kim, Mahadik, and Eldardiry]{chen2025probabilistic}
Hongjie Chen, Ryan~A Rossi, Sungchul Kim, Kanak Mahadik, and Hoda Eldardiry.
\newblock Probabilistic hypergraph recurrent neural networks for time-series forecasting.
\newblock In \emph{Proceedings of the 31st ACM SIGKDD Conference on Knowledge Discovery and Data Mining V. 1}, pages 82--93, 2025.

\bibitem[Chewi et~al.(2024)Chewi, Niles-Weed, and Rigollet]{chewi2024statistical}
Sinho Chewi, Jonathan Niles-Weed, and Philippe Rigollet.
\newblock Statistical optimal transport.
\newblock \emph{arXiv preprint arXiv:2407.18163}, 2024.

\bibitem[Cuturi and Blondel(2017)]{cuturi2017soft}
Marco Cuturi and Mathieu Blondel.
\newblock Soft-dtw: a differentiable loss function for time-series.
\newblock In \emph{International conference on machine learning}, pages 894--903. PMLR, 2017.

\bibitem[Das et~al.(2023)Das, Kong, Sen, and Zhou]{das2023decoder}
Abhimanyu Das, Weihao Kong, Rajat Sen, and Yichen Zhou.
\newblock A decoder-only foundation model for time-series forecasting.
\newblock \emph{arXiv preprint arXiv:2310.10688}, 2023.

\bibitem[Delon and Desolneux(2020)]{delon2020wasserstein}
Julie Delon and Agnes Desolneux.
\newblock A wasserstein-type distance in the space of gaussian mixture models.
\newblock \emph{SIAM Journal on Imaging Sciences}, 13\penalty0 (2):\penalty0 936--970, 2020.

\bibitem[Ding et~al.(2008)Ding, Trajcevski, Scheuermann, Wang, and Keogh]{ding2008querying}
Hui Ding, Goce Trajcevski, Peter Scheuermann, Xiaoyue Wang, and Eamonn Keogh.
\newblock Querying and mining of time series data: experimental comparison of representations and distance measures.
\newblock \emph{Proceedings of the VLDB Endowment}, 1\penalty0 (2):\penalty0 1542--1552, 2008.

\bibitem[Dowson and Landau(1982)]{dowson1982frechet}
DC~Dowson and BV666017 Landau.
\newblock The fr{\'e}chet distance between multivariate normal distributions.
\newblock \emph{Journal of multivariate analysis}, 12\penalty0 (3):\penalty0 450--455, 1982.

\bibitem[Driemel et~al.(2016)Driemel, Krivo{\v{s}}ija, and Sohler]{driemel2016clustering}
Anne Driemel, Amer Krivo{\v{s}}ija, and Christian Sohler.
\newblock Clustering time series under the fr{\'e}chet distance.
\newblock In \emph{Proceedings of the twenty-seventh annual ACM-SIAM symposium on Discrete algorithms}, pages 766--785. SIAM, 2016.

\bibitem[Dwork et~al.(2012)Dwork, Hardt, Pitassi, Reingold, and Zemel]{dwork2012fairness}
Cynthia Dwork, Moritz Hardt, Toniann Pitassi, Omer Reingold, and Richard Zemel.
\newblock Fairness through awareness.
\newblock In \emph{Proceedings of the 3rd innovations in theoretical computer science conference}, pages 214--226, 2012.

\bibitem[Gururangan et~al.(2020)Gururangan, Marasovi{\'c}, Swayamdipta, Lo, Beltagy, Downey, and Smith]{gururangan2020don}
Suchin Gururangan, Ana Marasovi{\'c}, Swabha Swayamdipta, Kyle Lo, Iz~Beltagy, Doug Downey, and Noah~A Smith.
\newblock Don't stop pretraining: Adapt language models to domains and tasks.
\newblock \emph{arXiv preprint arXiv:2004.10964}, 2020.

\bibitem[Iizumi et~al.(2024)Iizumi, Takimoto, Masaki, Maruyama, Kayaba, Takaya, and Masutomi]{iizumi2024hybrid}
Toshichika Iizumi, Takahiro Takimoto, Yoshimitsu Masaki, Atsushi Maruyama, Nobuyuki Kayaba, Yuhei Takaya, and Yuji Masutomi.
\newblock A hybrid reanalysis-forecast meteorological forcing data for advancing climate adaptation in agriculture.
\newblock \emph{Scientific Data}, 11\penalty0 (1):\penalty0 849, 2024.

\bibitem[Kamada et~al.(1989)Kamada, Kawai, et~al.]{kamada1989algorithm}
Tomihisa Kamada, Satoru Kawai, et~al.
\newblock An algorithm for drawing general undirected graphs.
\newblock \emph{Information processing letters}, 31\penalty0 (1):\penalty0 7--15, 1989.

\bibitem[Liang and Zou(2022)]{liang2022metashift}
Weixin Liang and James Zou.
\newblock Metashift: A dataset of datasets for evaluating contextual distribution shifts and training conflicts.
\newblock \emph{arXiv preprint arXiv:2202.06523}, 2022.

\bibitem[Lim and Zohren(2021)]{lim2021time}
Bryan Lim and Stefan Zohren.
\newblock Time-series forecasting with deep learning: a survey.
\newblock \emph{Philosophical Transactions of the Royal Society A}, 379\penalty0 (2194):\penalty0 20200209, 2021.

\bibitem[Lin and Ruszczy{\'n}ski(2023)]{lin2023integrated}
Zhengqi Lin and Andrzej Ruszczy{\'n}ski.
\newblock An integrated transportation distance between kernels and approximate dynamic risk evaluation in markov systems.
\newblock \emph{SIAM Journal on Control and Optimization}, 61\penalty0 (6):\penalty0 3559--3583, 2023.

\bibitem[Liu et~al.(2024)Liu, Hu, Li, Diao, Liang, Hooi, and Zimmermann]{liu2024unitime}
Xu~Liu, Junfeng Hu, Yuan Li, Shizhe Diao, Yuxuan Liang, Bryan Hooi, and Roger Zimmermann.
\newblock Unitime: A language-empowered unified model for cross-domain time series forecasting.
\newblock In \emph{Proceedings of the ACM on Web Conference 2024}, pages 4095--4106, 2024.

\bibitem[Mallasto and Feragen(2017)]{mallasto2017learning}
Anton Mallasto and Aasa Feragen.
\newblock Learning from uncertain curves: The 2-wasserstein metric for gaussian processes.
\newblock \emph{Advances in Neural Information Processing Systems}, 30, 2017.

\bibitem[Masarotto et~al.(2019)Masarotto, Panaretos, and Zemel]{masarotto2019procrustes}
Valentina Masarotto, Victor~M Panaretos, and Yoav Zemel.
\newblock Procrustes metrics on covariance operators and optimal transportation of gaussian processes.
\newblock \emph{Sankhya A}, 81:\penalty0 172--213, 2019.

\bibitem[Mehra et~al.(2024)Mehra, Zhang, and Hamm]{mehraunderstanding}
Akshay Mehra, Yunbei Zhang, and Jihun Hamm.
\newblock Understanding the transferability of representations via task-relatedness.
\newblock In \emph{The Thirty-eighth Annual Conference on Neural Information Processing Systems}, 2024.

\bibitem[Morid et~al.(2023)Morid, Sheng, and Dunbar]{morid2023time}
Mohammad~Amin Morid, Olivia R~Liu Sheng, and Joseph Dunbar.
\newblock Time series prediction using deep learning methods in healthcare.
\newblock \emph{ACM Transactions on Management Information Systems}, 14\penalty0 (1):\penalty0 1--29, 2023.

\bibitem[Nie et~al.(2023)Nie, H.~Nguyen, Sinthong, and Kalagnanam]{Yuqietal-2023-PatchTST}
Yuqi Nie, Nam H.~Nguyen, Phanwadee Sinthong, and Jayant Kalagnanam.
\newblock A time series is worth 64 words: Long-term forecasting with transformers.
\newblock In \emph{International Conference on Learning Representations}, 2023.

\bibitem[Nowozin et~al.(2016)Nowozin, Cseke, and Tomioka]{nowozin2016f}
Sebastian Nowozin, Botond Cseke, and Ryota Tomioka.
\newblock f-gan: Training generative neural samplers using variational divergence minimization.
\newblock \emph{Advances in neural information processing systems}, 29, 2016.

\bibitem[Okano and Imaizumi(2024)]{okano2024distribution}
Ryo Okano and Masaaki Imaizumi.
\newblock Distribution-on-distribution regression with wasserstein metric: Multivariate gaussian case.
\newblock \emph{Journal of Multivariate Analysis}, 203:\penalty0 105334, 2024.

\bibitem[Painblanc et~al.(2023)Painblanc, Chapel, Courty, Friguet, Pelletier, and Tavenard]{mad2023}
François Painblanc, Laetitia Chapel, Nicolas Courty, Chloé Friguet, Charlotte Pelletier, and Romain Tavenard.
\newblock {Match-And-Deform: Time Series Domain Adaptation through Optimal Transport and Temporal Alignment}.
\newblock In \emph{{European Conference on Machine Learning and Principles and Practice of Knowledge Discovery}}, 2023.

\bibitem[Panaretos and Zemel(2020)]{panaretos2020invitation}
Victor~M Panaretos and Yoav Zemel.
\newblock \emph{An invitation to statistics in Wasserstein space}.
\newblock Springer Nature, 2020.

\bibitem[Rasul et~al.(2024)Rasul, Ashok, Williams, Ghonia, Bhagwatkar, Khorasani, Bayazi, Adamopoulos, Riachi, Hassen, et~al.]{rasul2024lag}
Kashif Rasul, Arjun Ashok, Andrew~Robert Williams, Hena Ghonia, Rishika Bhagwatkar, Arian Khorasani, Mohammad Javad~Darvishi Bayazi, George Adamopoulos, Roland Riachi, Nadhir Hassen, et~al.
\newblock Lag-llama: Towards foundation models for probabilistic time series forecasting.
\newblock \emph{Preprint}, 2024.

\bibitem[Sanyal et~al.(2024)Sanyal, Hu, Yu, Ma, Wang, and Sch{\"o}lkopf]{sanyal2024accuracy}
Amartya Sanyal, Yaxi Hu, Yaodong Yu, Yian Ma, Yixin Wang, and Bernhard Sch{\"o}lkopf.
\newblock Accuracy on the wrong line: On the pitfalls of noisy data for out-of-distribution generalisation.
\newblock \emph{arXiv preprint arXiv:2406.19049}, 2024.

\bibitem[Sezer et~al.(2020)Sezer, Gudelek, and Ozbayoglu]{sezer2020financial}
Omer~Berat Sezer, Mehmet~Ugur Gudelek, and Ahmet~Murat Ozbayoglu.
\newblock Financial time series forecasting with deep learning: A systematic literature review: 2005--2019.
\newblock \emph{Applied soft computing}, 90:\penalty0 106181, 2020.

\bibitem[Shnitzer et~al.(2022)Shnitzer, Yurochkin, Greenewald, and Solomon]{shnitzer2022log}
Tal Shnitzer, Mikhail Yurochkin, Kristjan Greenewald, and Justin~M Solomon.
\newblock Log-euclidean signatures for intrinsic distances between unaligned datasets.
\newblock In \emph{International Conference on Machine Learning}, pages 20106--20124. PMLR, 2022.

\bibitem[Sierra et~al.(2025)Sierra, Gillespie, Soltani, Exposito-Alonso, and Kattenborn]{sierra2025divshift}
Elena Sierra, Lauren~E Gillespie, Salim Soltani, Moises Exposito-Alonso, and Teja Kattenborn.
\newblock Divshift: Exploring domain-specific distribution shifts in large-scale, volunteer-collected biodiversity datasets.
\newblock In \emph{Proceedings of the AAAI Conference on Artificial Intelligence}, volume~39, pages 28386--28396, 2025.

\bibitem[Wang et~al.(2016)Wang, Zhao, Xiong, Fan, Sun, Ma, and Liu]{wang2016research}
Kai Wang, Youjin Zhao, Qingyu Xiong, Min Fan, Guotan Sun, Longkun Ma, and Tong Liu.
\newblock Research on healthy anomaly detection model based on deep learning from multiple time-series physiological signals.
\newblock \emph{Scientific Programming}, 2016\penalty0 (1):\penalty0 5642856, 2016.

\bibitem[Wang et~al.(2023)Wang, Han, and Li]{wang2023similarity}
Xiaoying Wang, Yuecai Han, and Yong Li.
\newblock Similarity between two stochastic differential systems.
\newblock \emph{arXiv preprint arXiv:2310.10901}, 2023.

\bibitem[Wu et~al.(2018)Wu, Huang, Thoma, Acharya, and Van~Gool]{wu2018wasserstein}
Jiqing Wu, Zhiwu Huang, Janine Thoma, Dinesh Acharya, and Luc Van~Gool.
\newblock Wasserstein divergence for gans.
\newblock In \emph{Proceedings of the European conference on computer vision (ECCV)}, pages 653--668, 2018.

\bibitem[Yang et~al.(2022)Yang, Wang, Sun, and Peng]{yang2022fast}
Wenzhe Yang, Sheng Wang, Yuan Sun, and Zhiyong Peng.
\newblock Fast dataset search with earth mover's distance.
\newblock \emph{Proceedings of the VLDB Endowment}, 15\penalty0 (11):\penalty0 2517--2529, 2022.

\bibitem[Zhang et~al.(2024)Zhang, Li, Lin, Wang, Qian, and Ge]{ijcai2024p186}
Hansong Zhang, Shikun Li, Fanzhao Lin, Weiping Wang, Zhenxing Qian, and Shiming Ge.
\newblock Dance: Dual-view distribution alignment for dataset condensation.
\newblock In Kate Larson, editor, \emph{Proceedings of the Thirty-Third International Joint Conference on Artificial Intelligence, {IJCAI-24}}, pages 1679--1687. International Joint Conferences on Artificial Intelligence Organization, 8 2024.
\newblock \doi{10.24963/ijcai.2024/186}.
\newblock URL \url{https://doi.org/10.24963/ijcai.2024/186}.
\newblock Main Track.

\bibitem[Zhao et~al.(2023)Zhao, Li, Qin, and Yu]{zhao2023improved}
Ganlong Zhao, Guanbin Li, Yipeng Qin, and Yizhou Yu.
\newblock Improved distribution matching for dataset condensation.
\newblock In \emph{Proceedings of the IEEE/CVF Conference on Computer Vision and Pattern Recognition}, pages 7856--7865, 2023.

\bibitem[Zhu et~al.(2020)Zhu, Chen, Hu, Hou, Liu, and Zhang]{zhu2020drgraph}
Minfeng Zhu, Wei Chen, Yuanzhe Hu, Yuxuan Hou, Liangjun Liu, and Kaiyuan Zhang.
\newblock Drgraph: An efficient graph layout algorithm for large-scale graphs by dimensionality reduction.
\newblock \emph{IEEE Transactions on Visualization and Computer Graphics}, 27\penalty0 (2):\penalty0 1666--1676, 2020.

\end{thebibliography}
\clearpage
\newpage
\appendix

\begin{figure*}[t]
\centering
\includegraphics[width=0.16\linewidth]{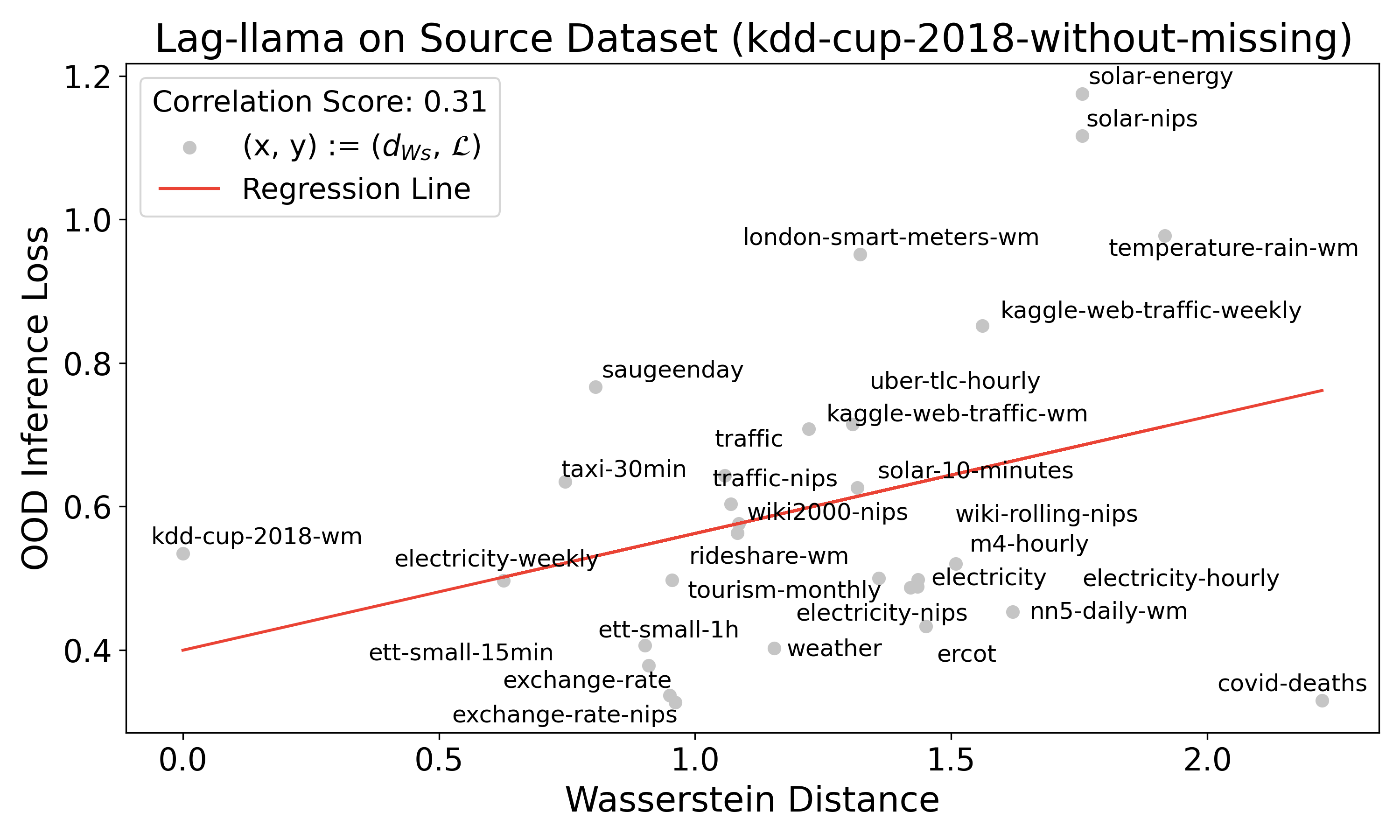}
\includegraphics[width=0.16\linewidth]{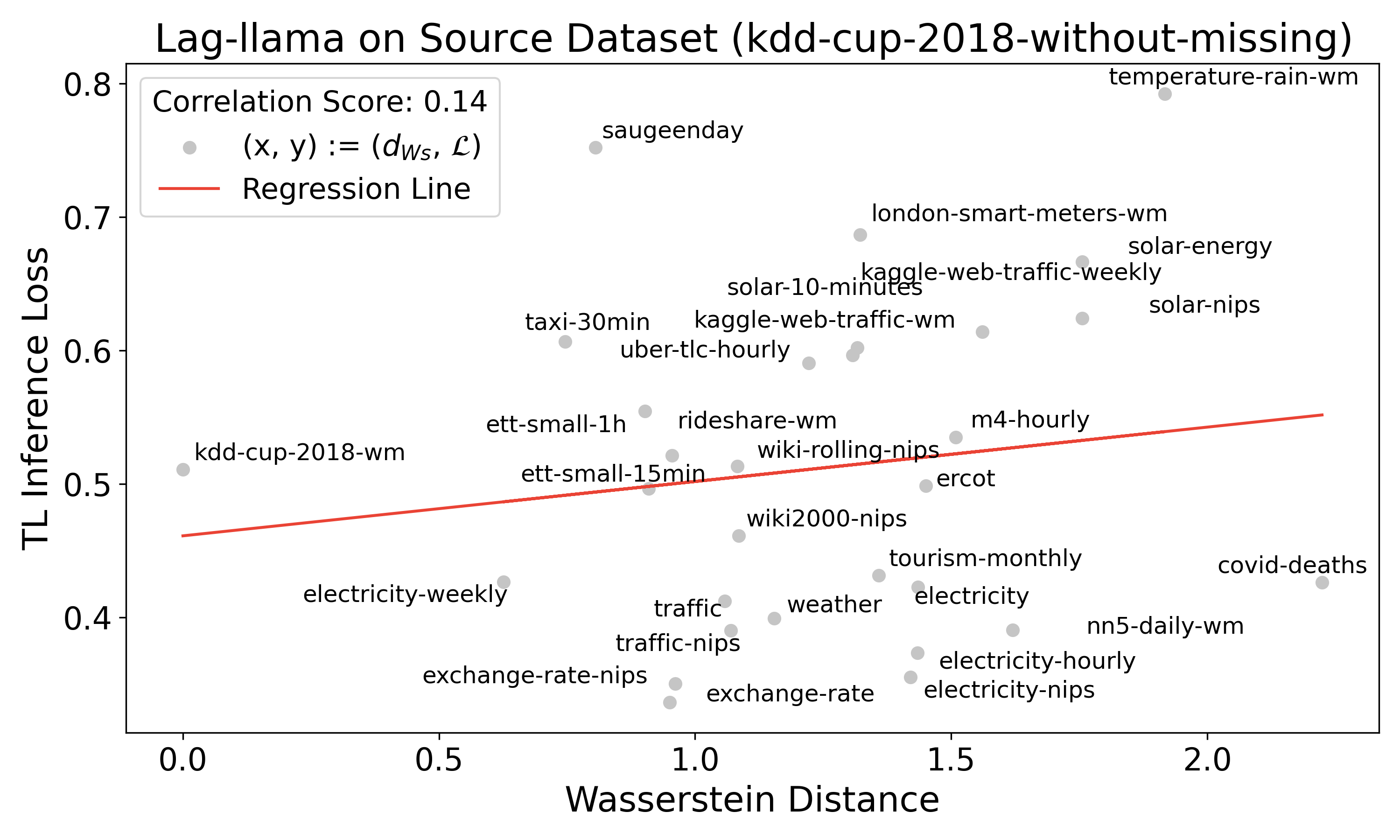}
\includegraphics[width=0.16\linewidth]{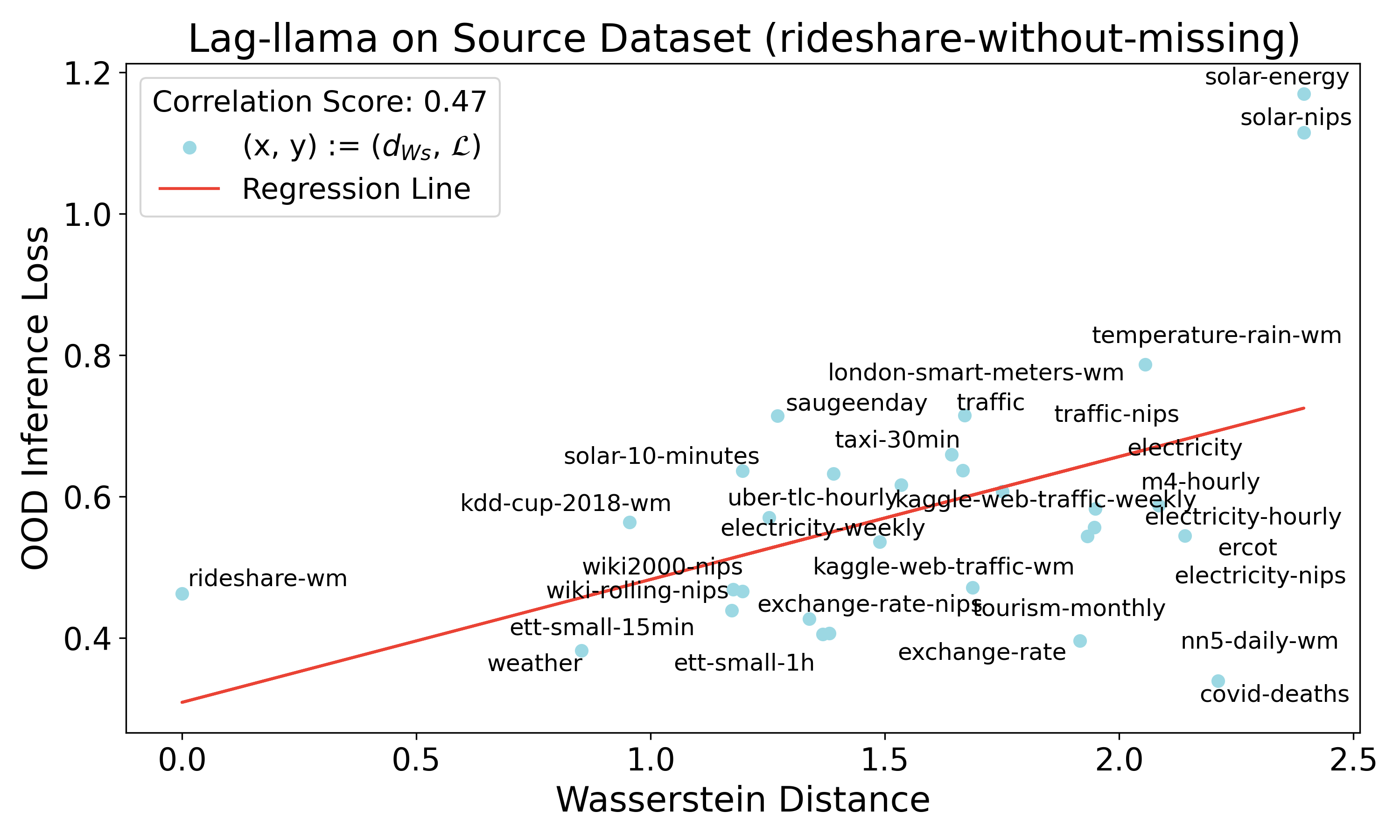}
\includegraphics[width=0.16\linewidth]{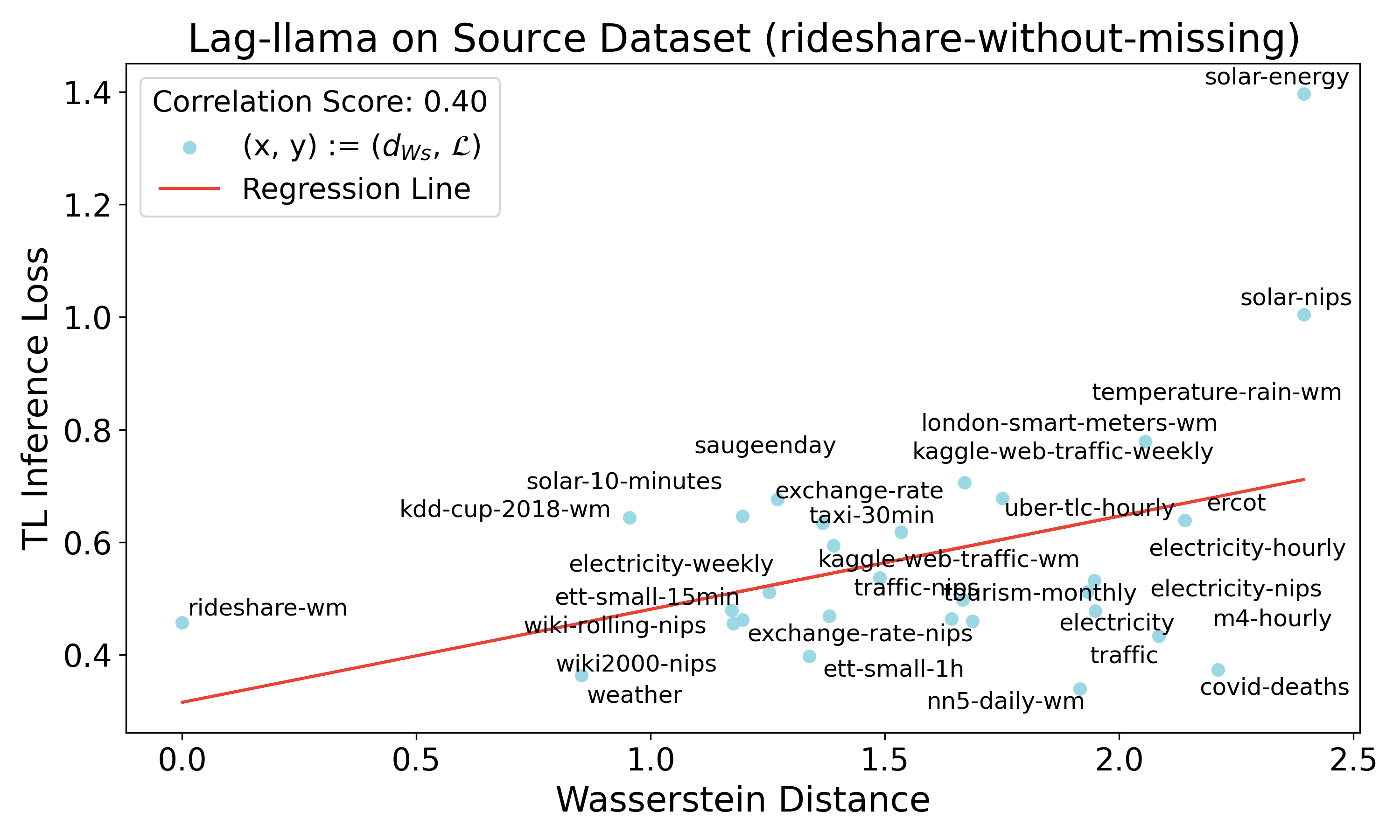}
\includegraphics[width=0.16\linewidth]{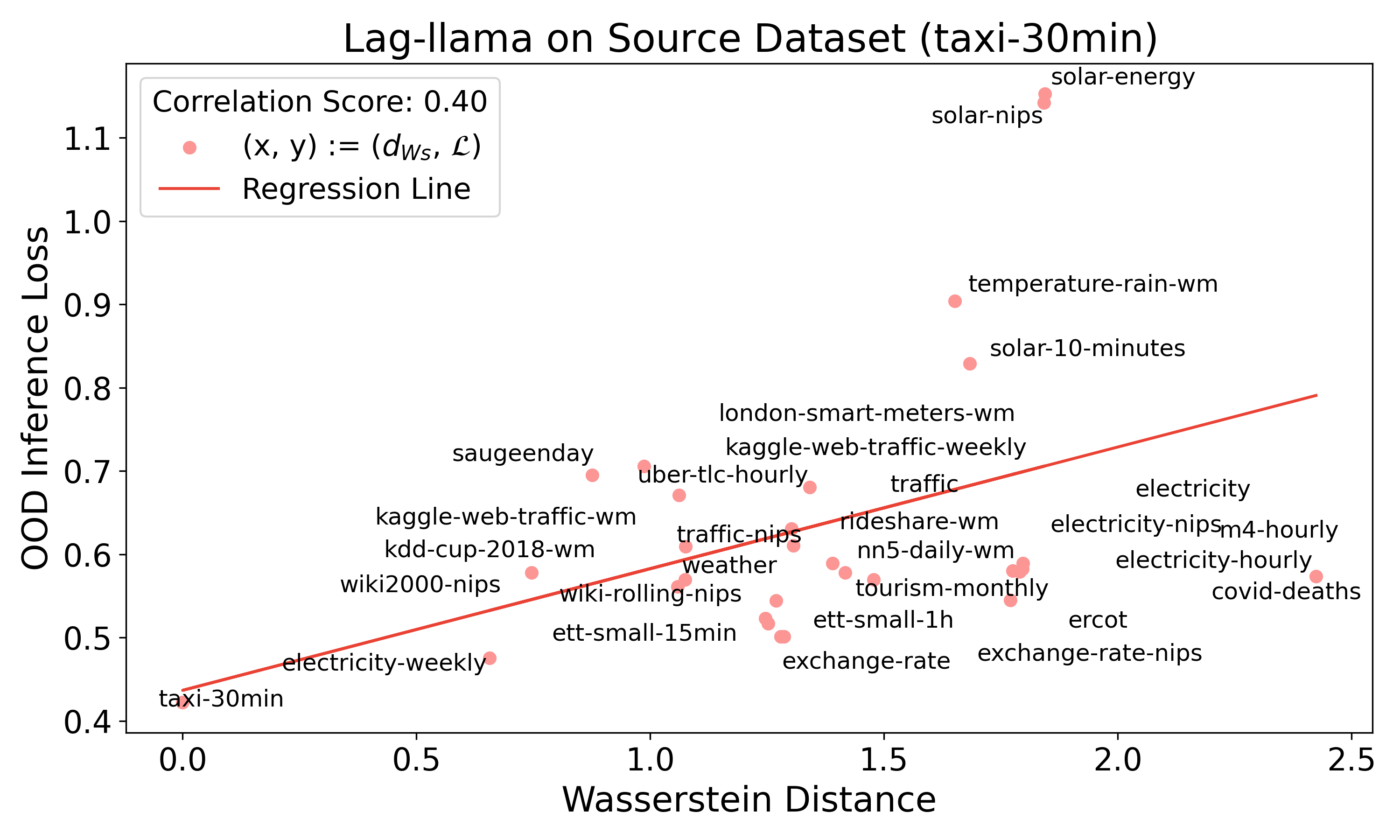}
\includegraphics[width=0.16\linewidth]{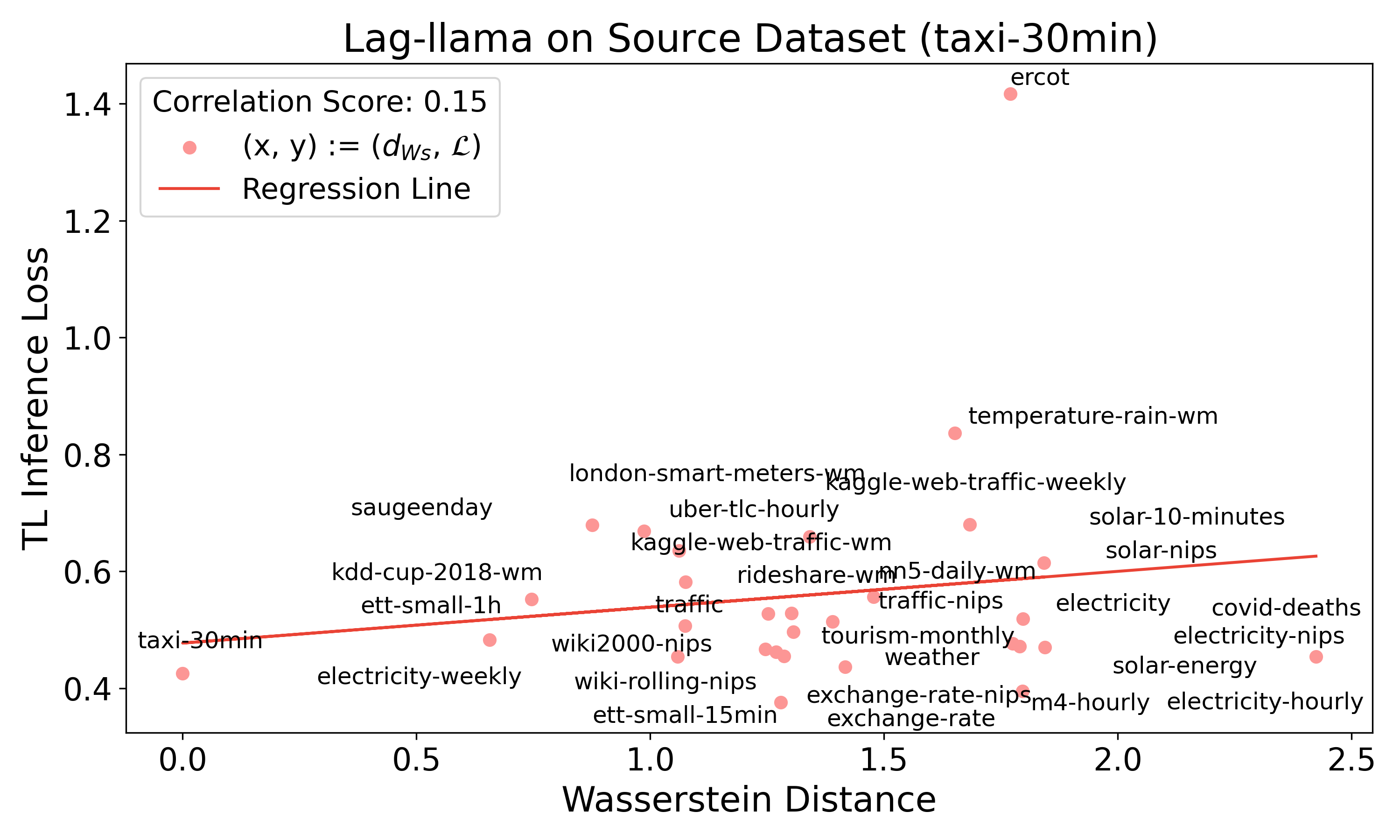}

\includegraphics[width=0.16\linewidth]{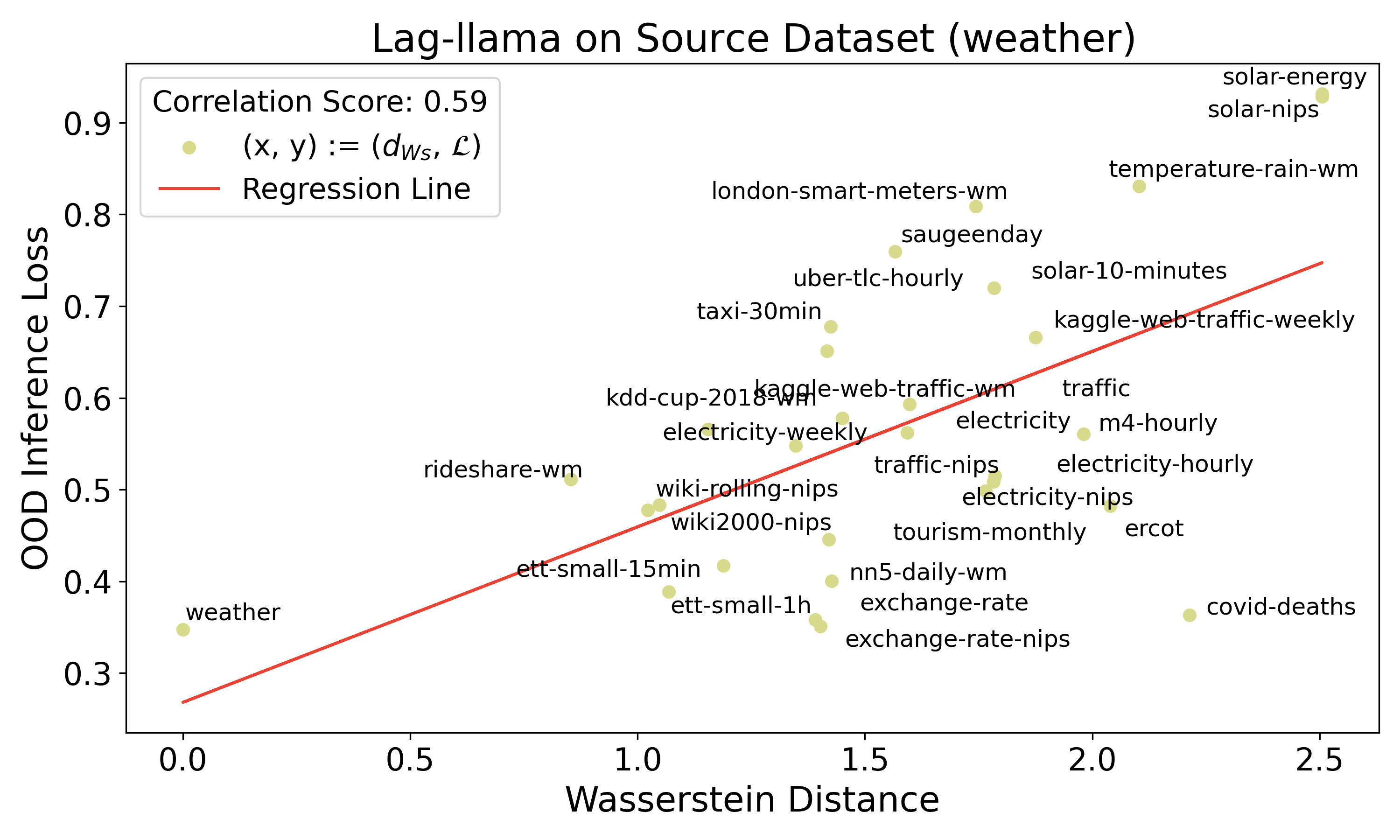}
\includegraphics[width=0.16\linewidth]{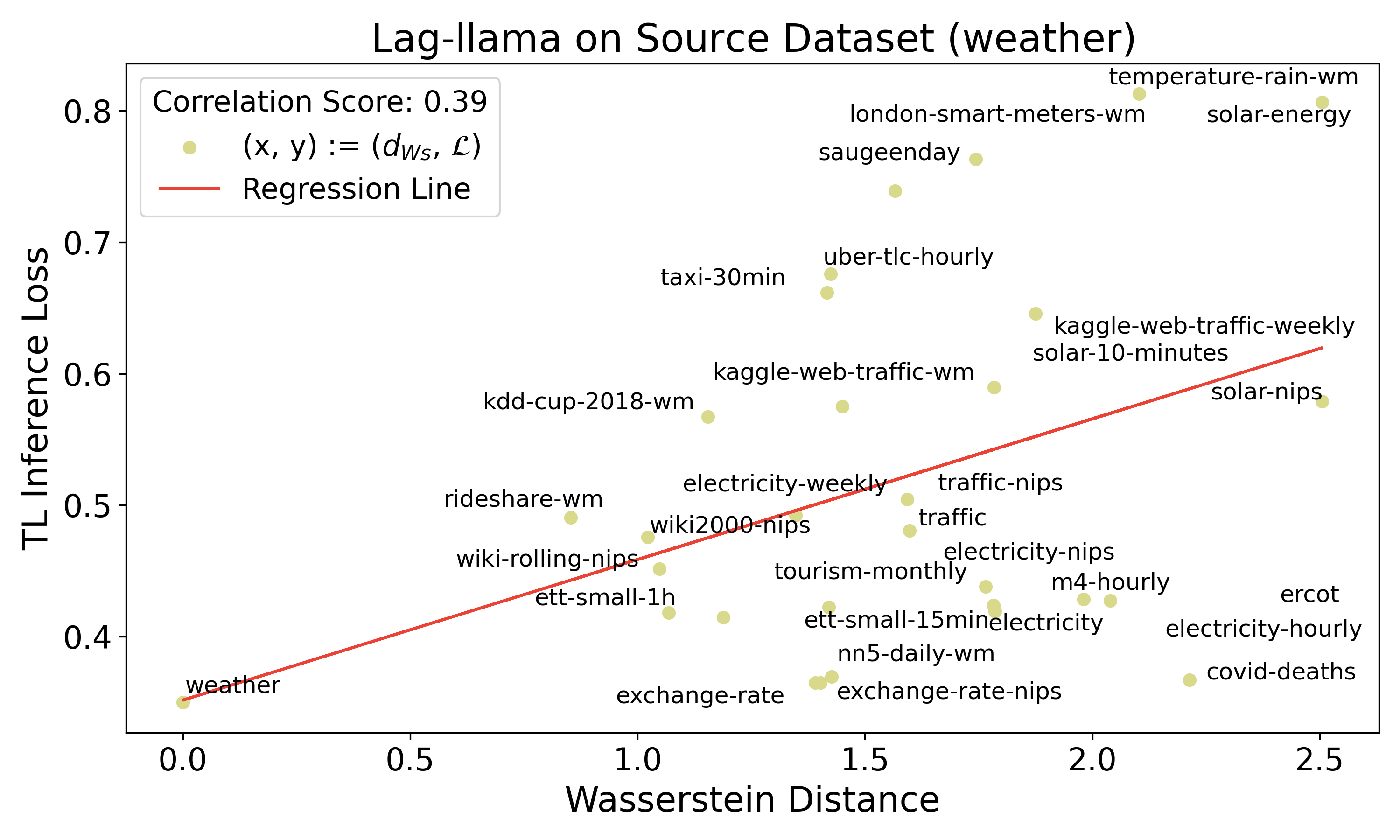}
\includegraphics[width=0.16\linewidth]{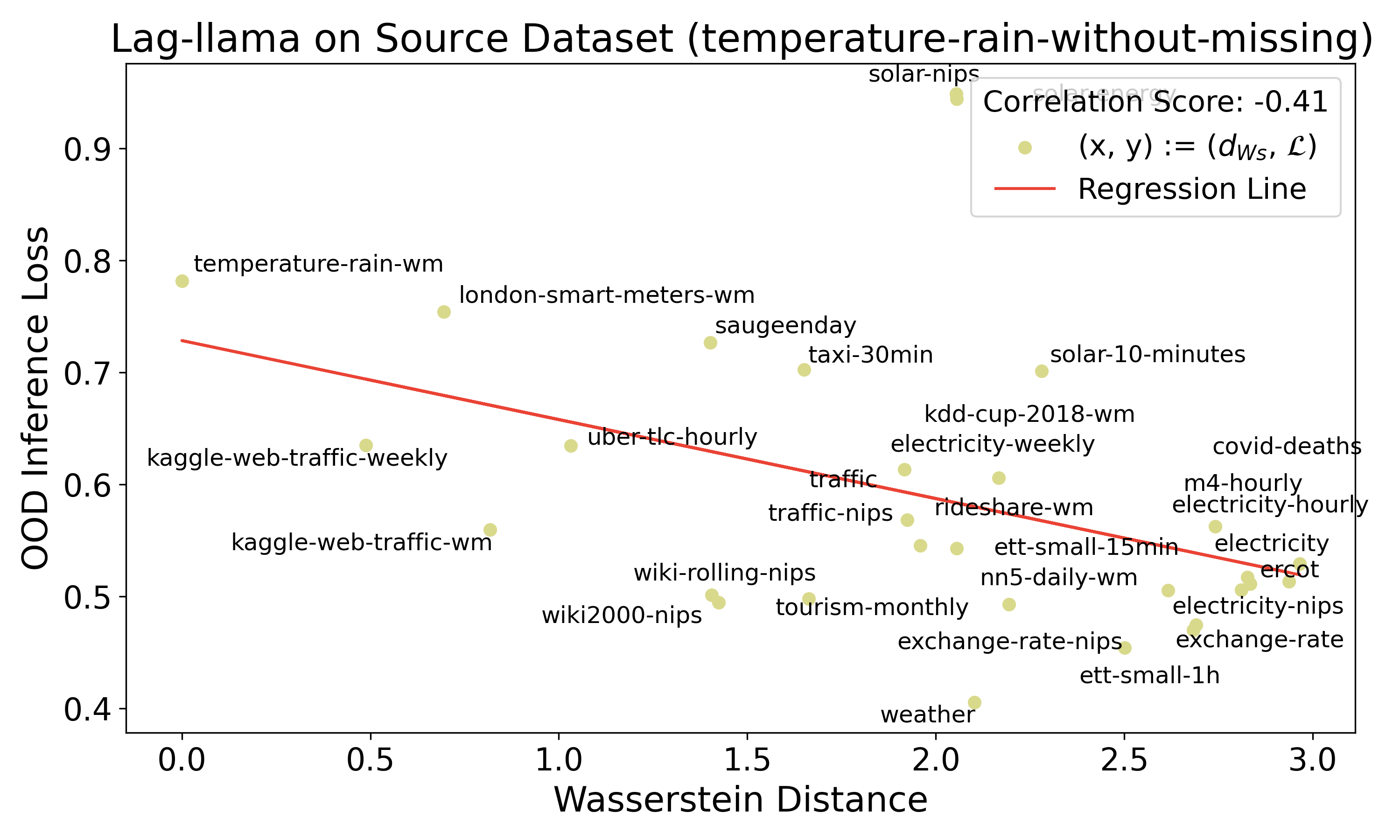}
\includegraphics[width=0.16\linewidth]{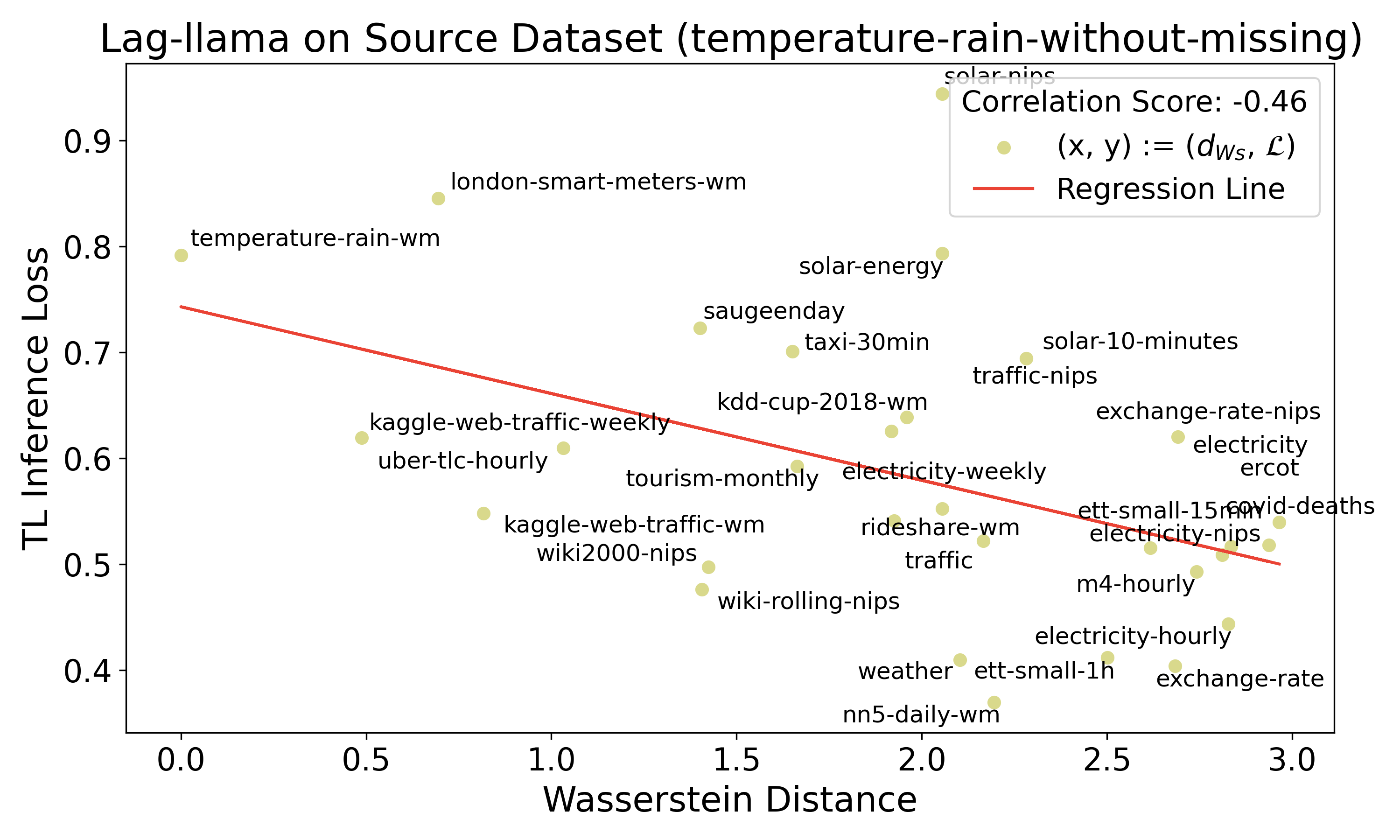}
\includegraphics[width=0.16\linewidth]{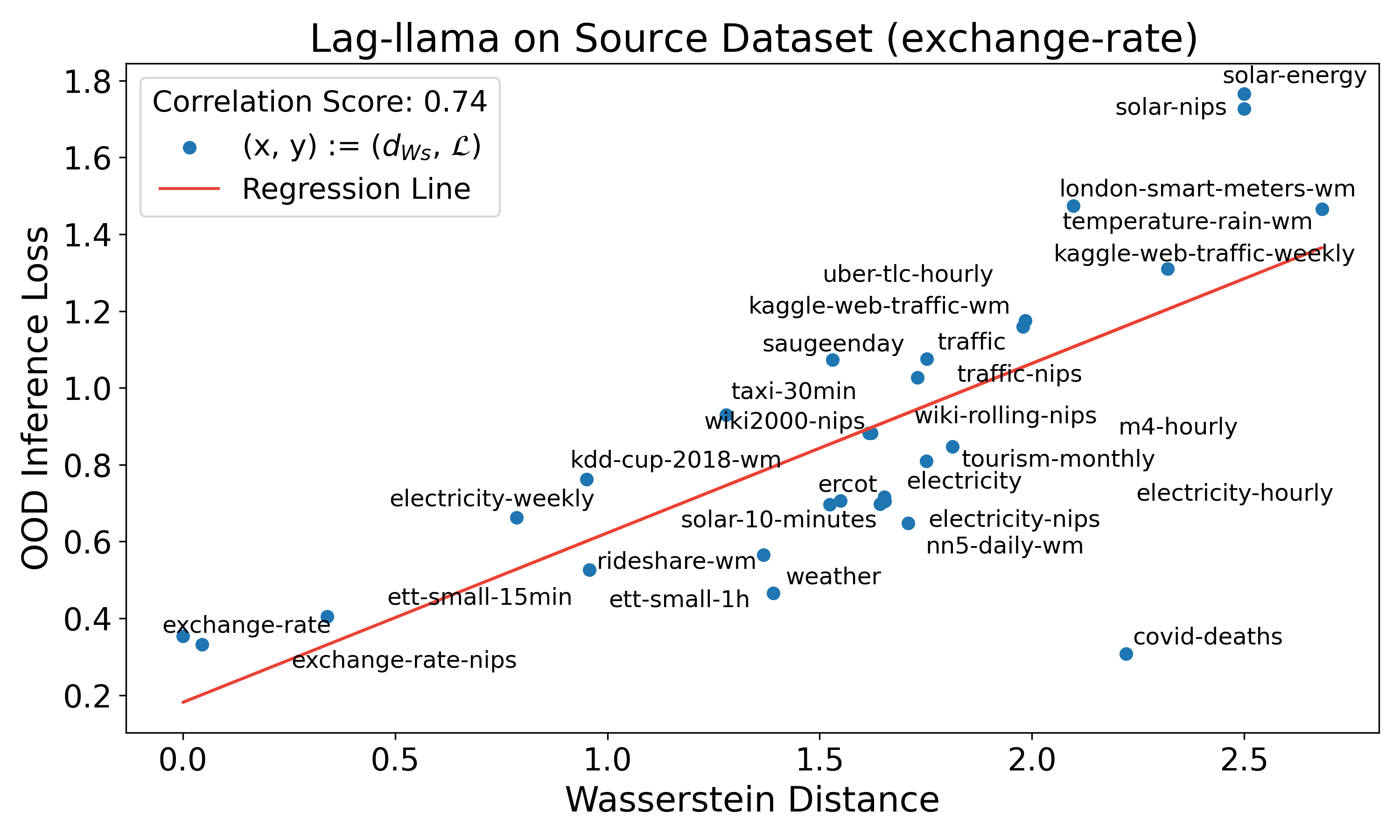}
\includegraphics[width=0.16\linewidth]{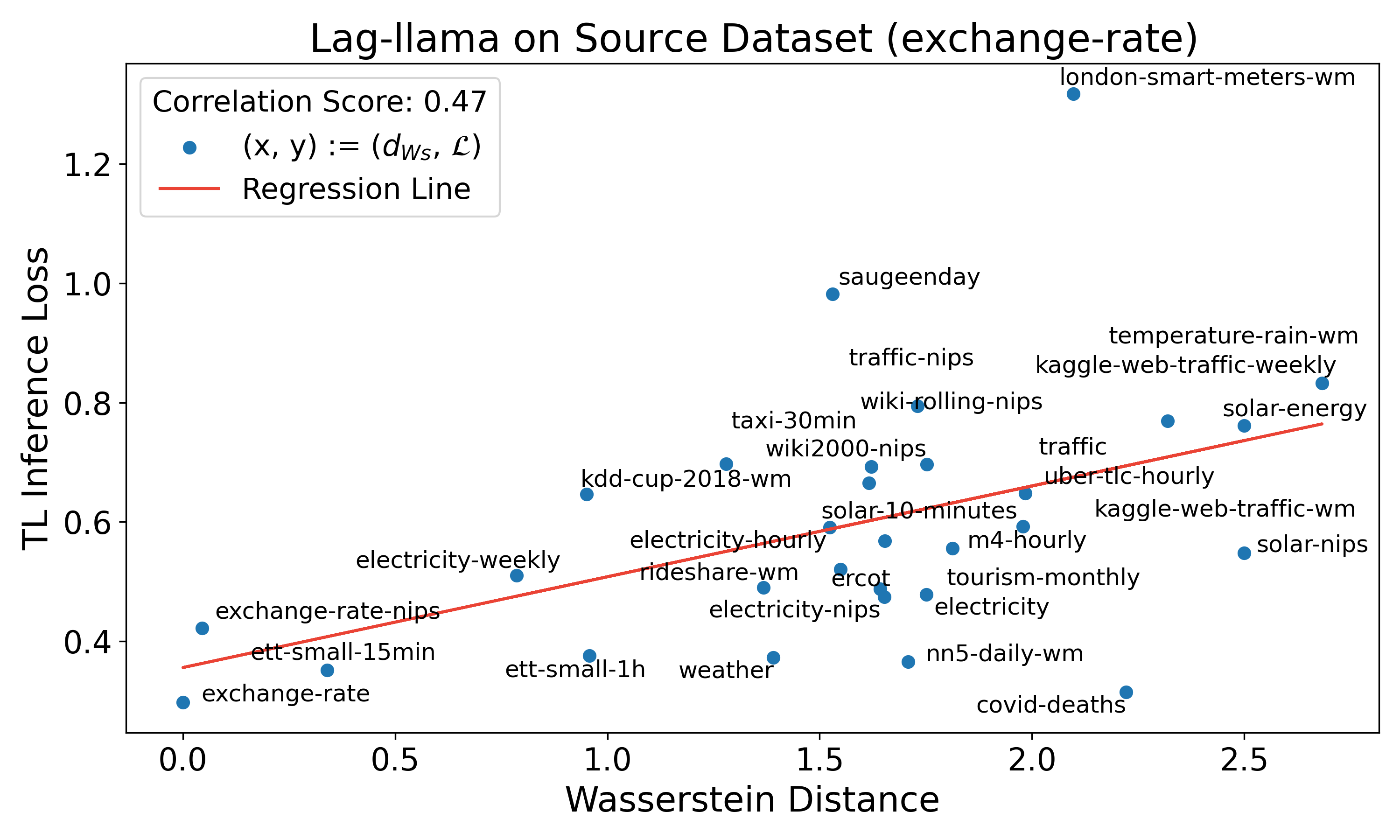}

\includegraphics[width=0.16\linewidth]{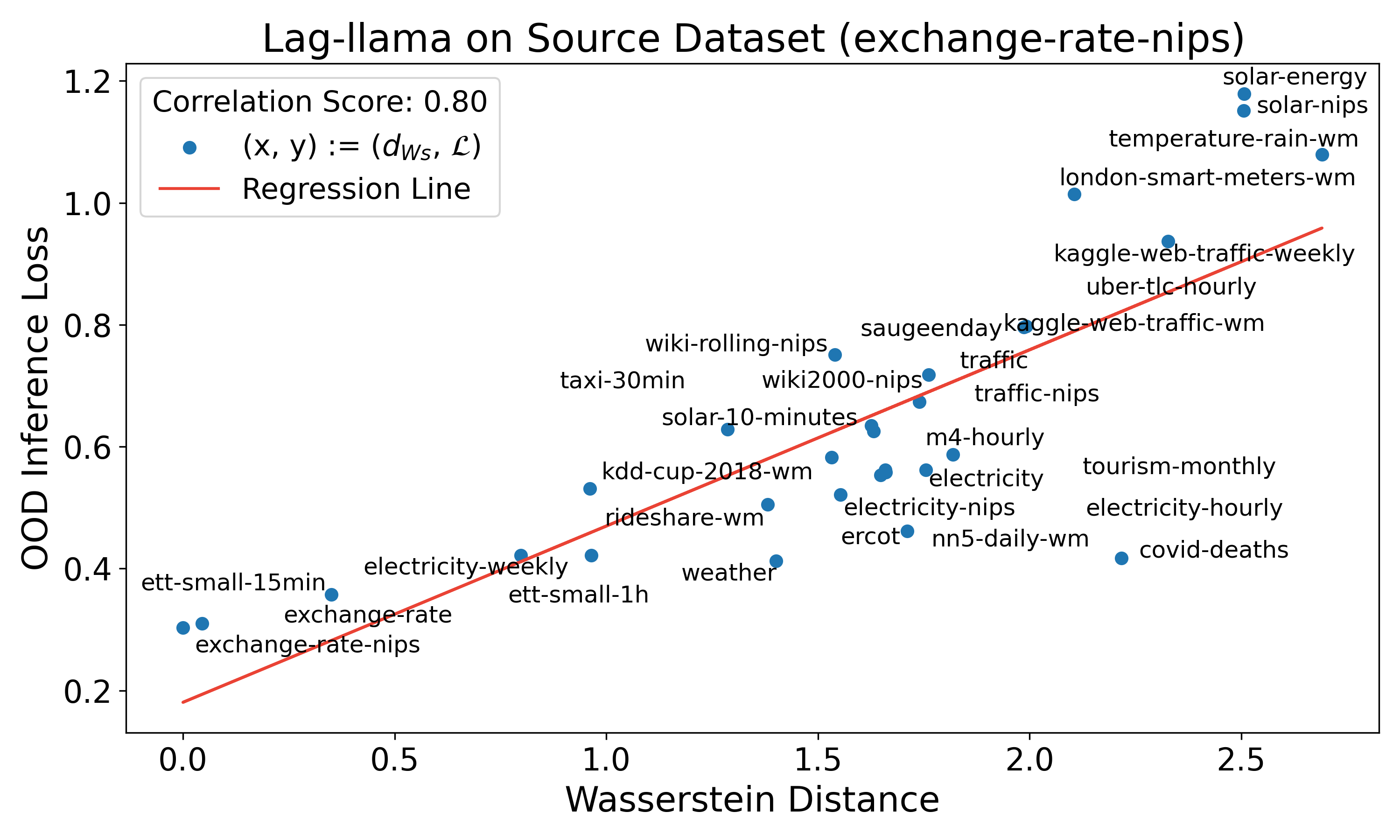}
\includegraphics[width=0.16\linewidth]{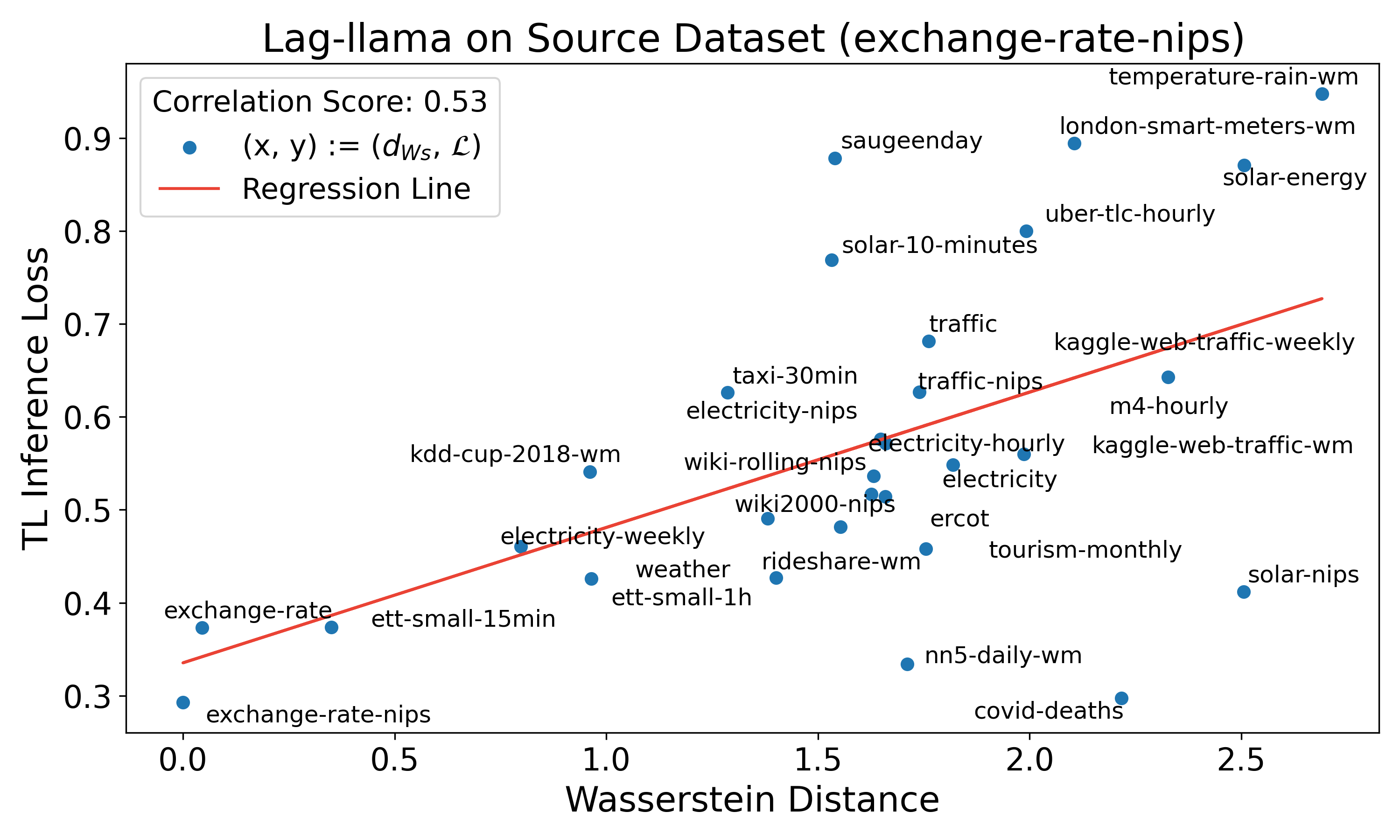}
\includegraphics[width=0.16\linewidth]{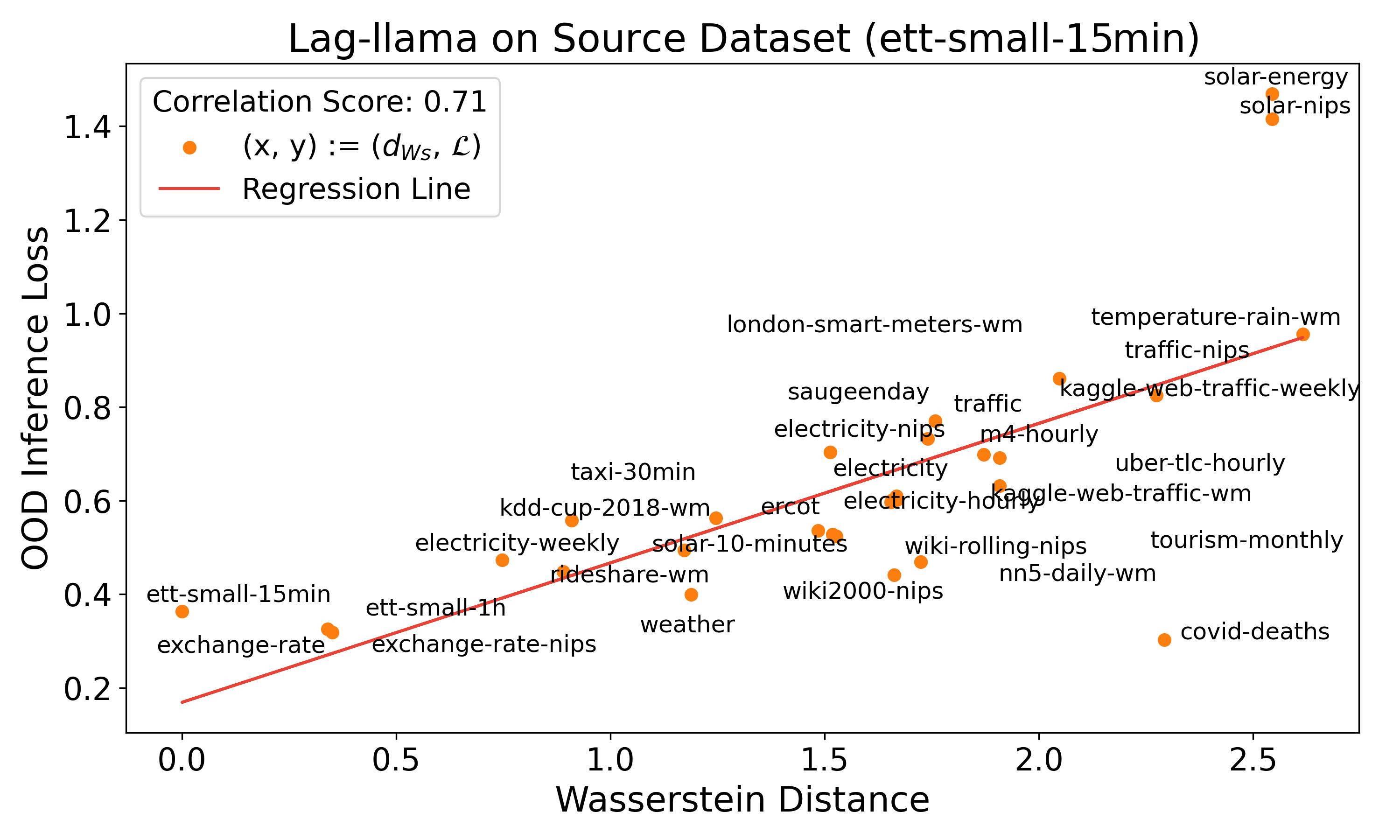}
\includegraphics[width=0.16\linewidth]{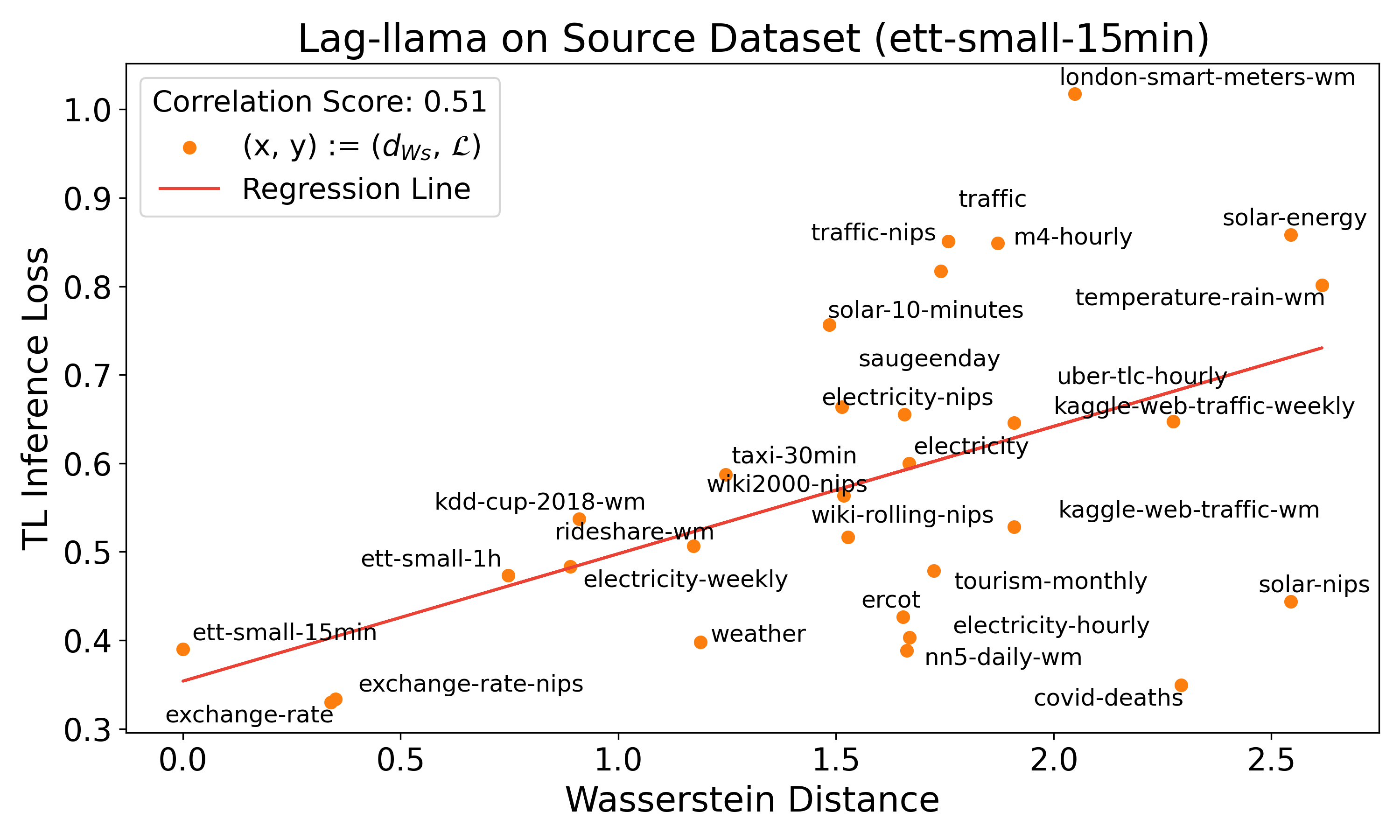}
\includegraphics[width=0.16\linewidth]{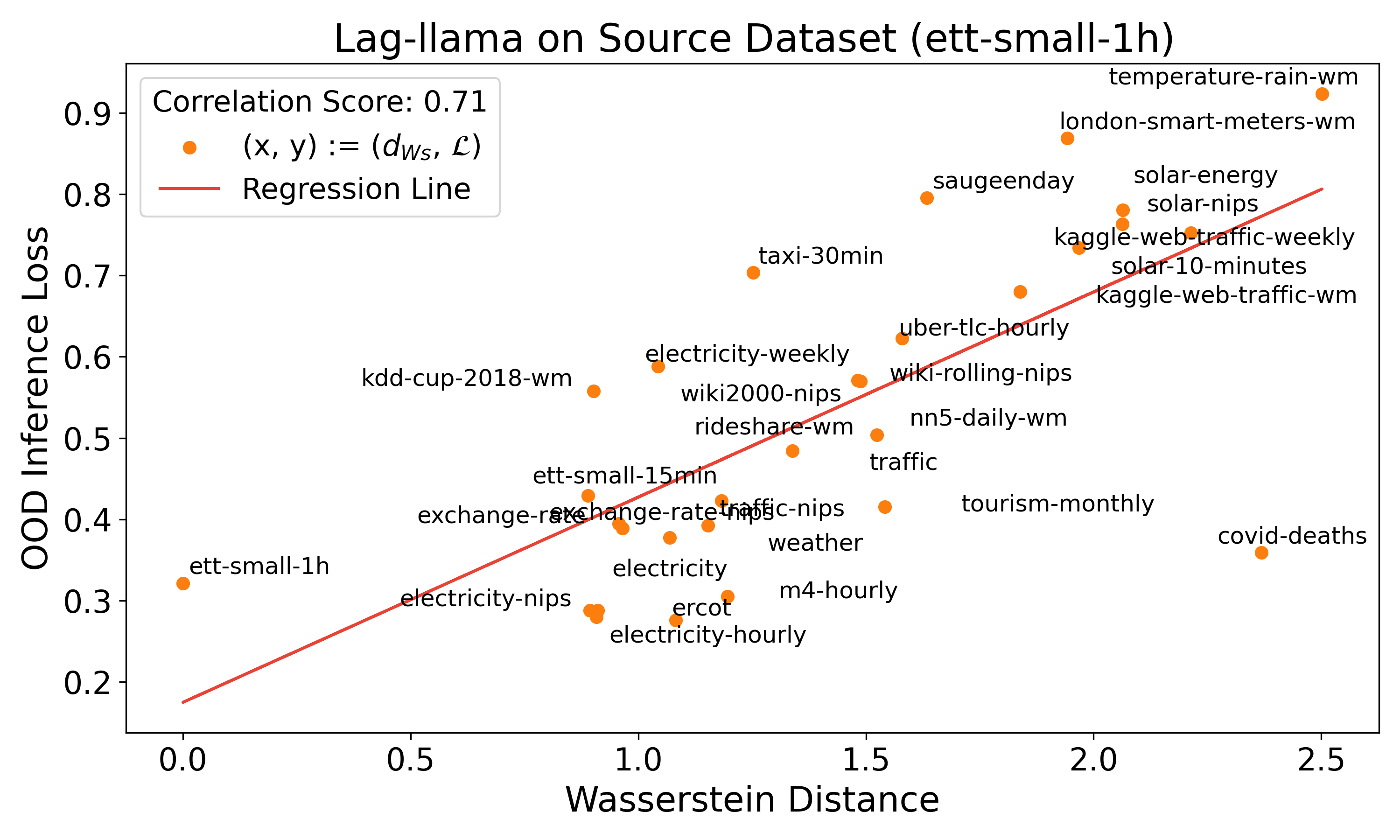}
\includegraphics[width=0.16\linewidth]{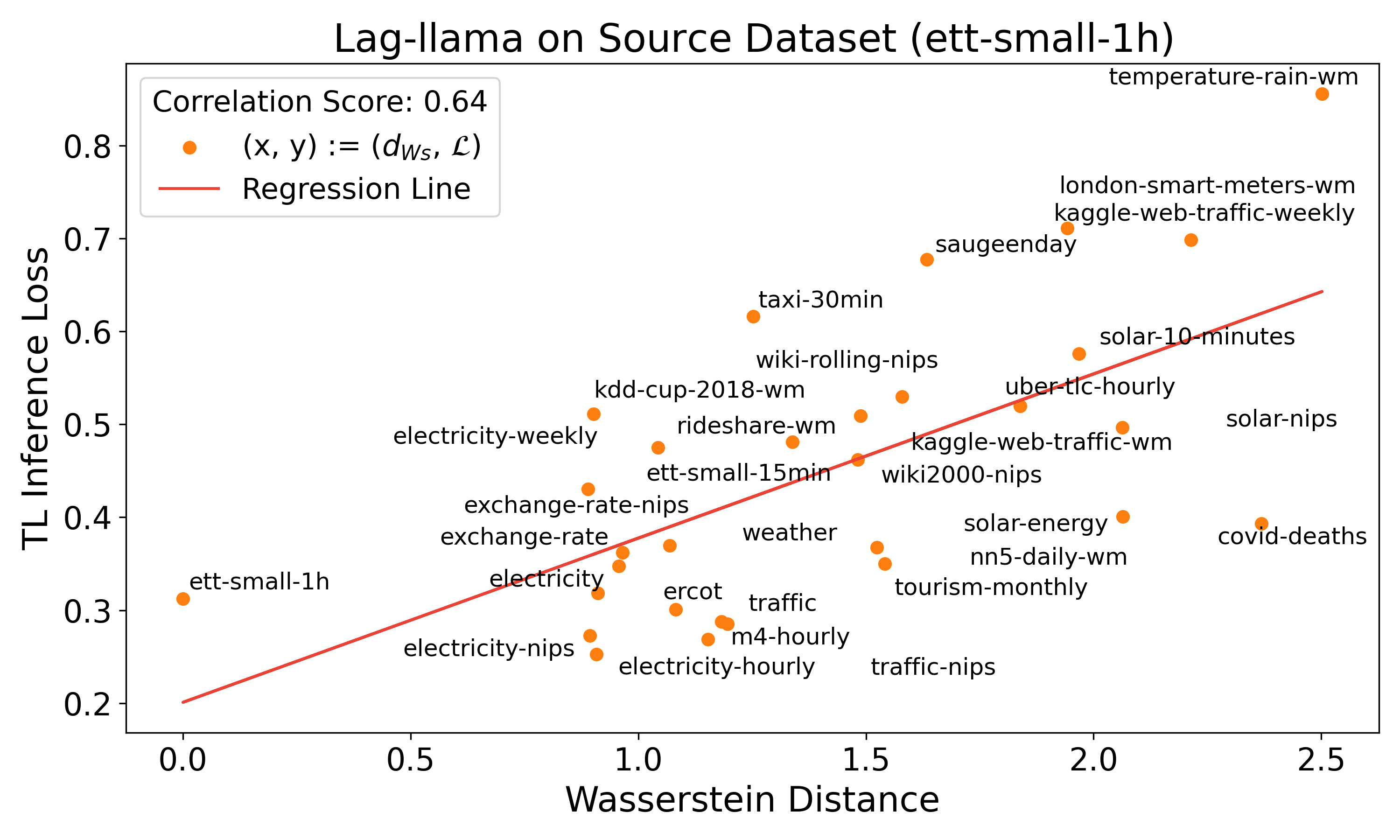}

\includegraphics[width=0.16\linewidth]{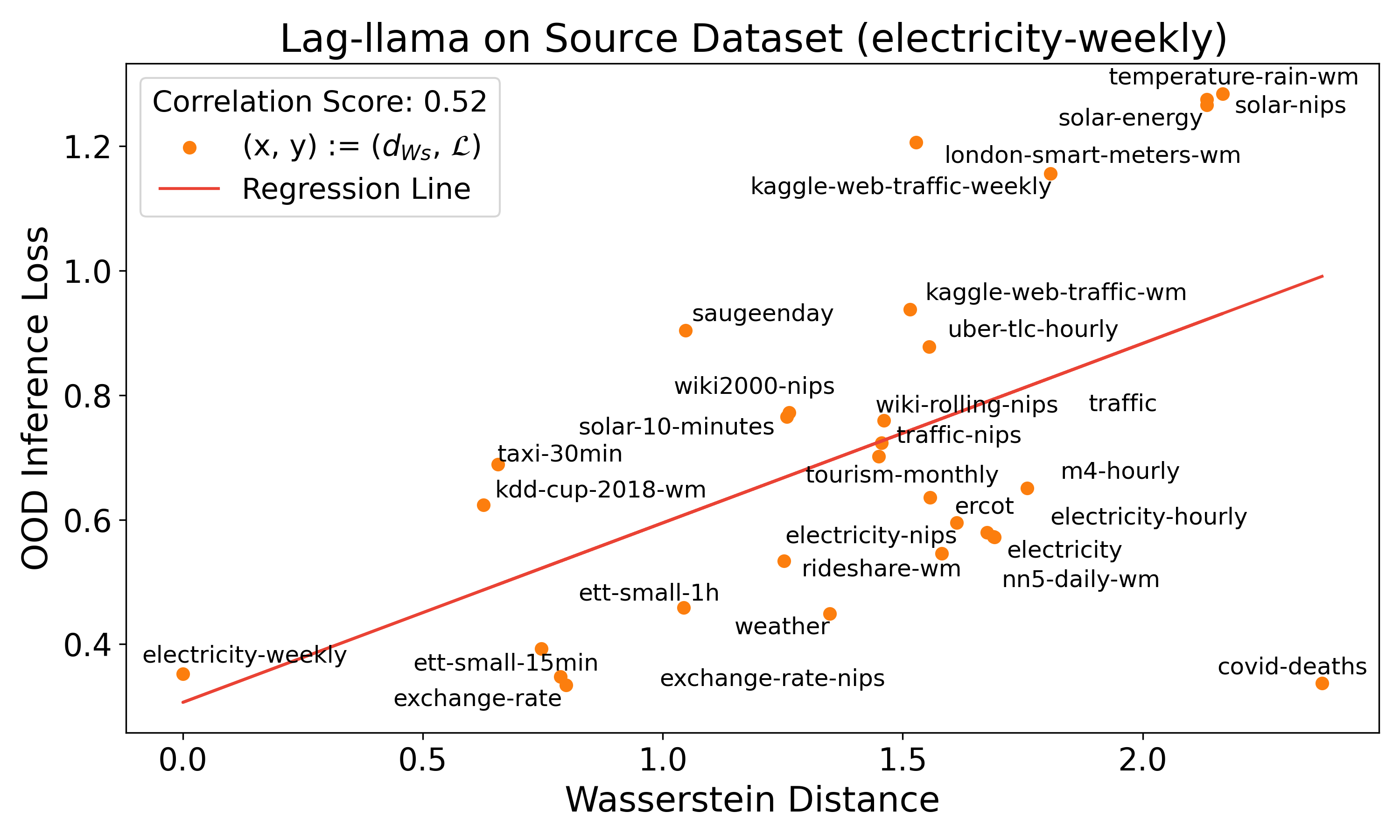}
\includegraphics[width=0.16\linewidth]{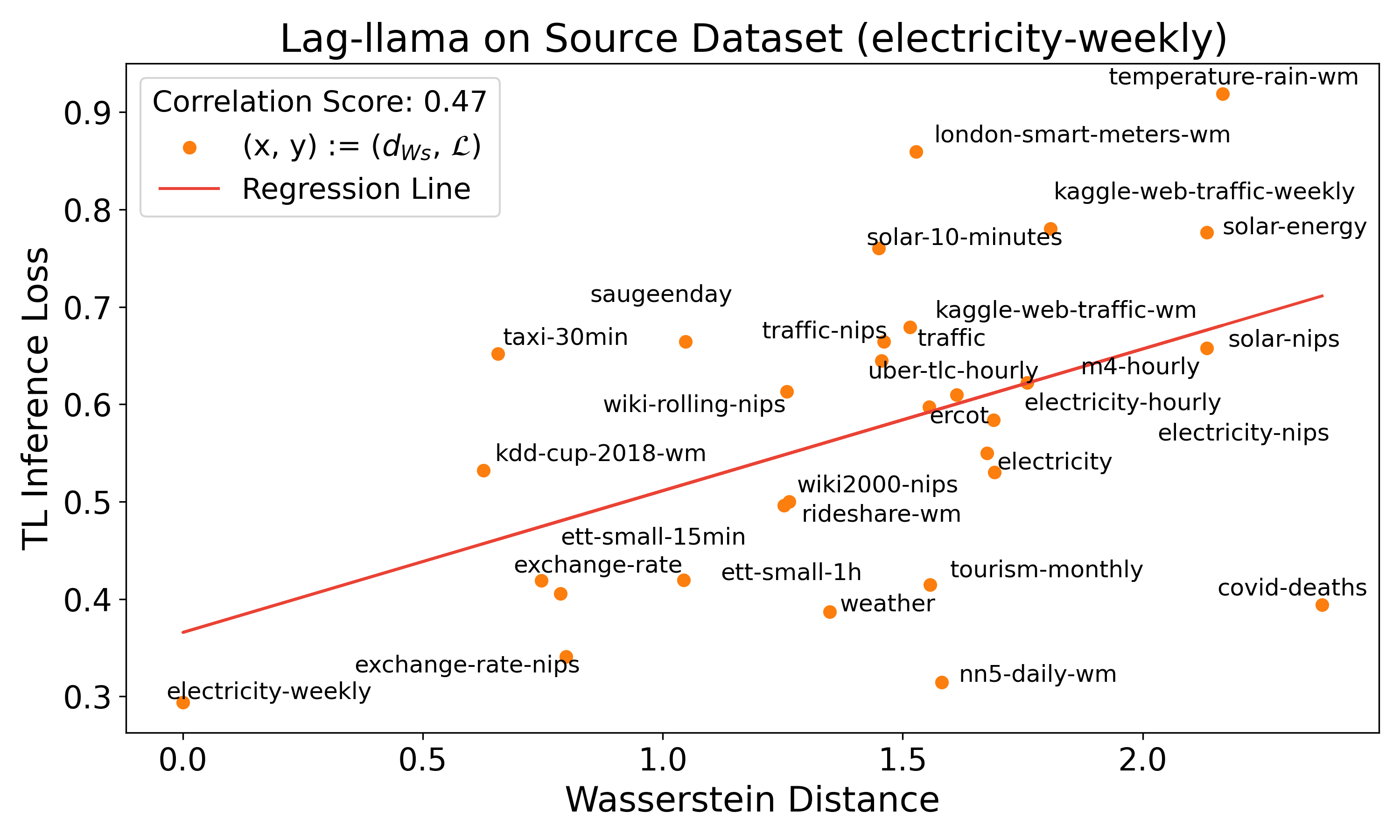}
\includegraphics[width=0.16\linewidth]{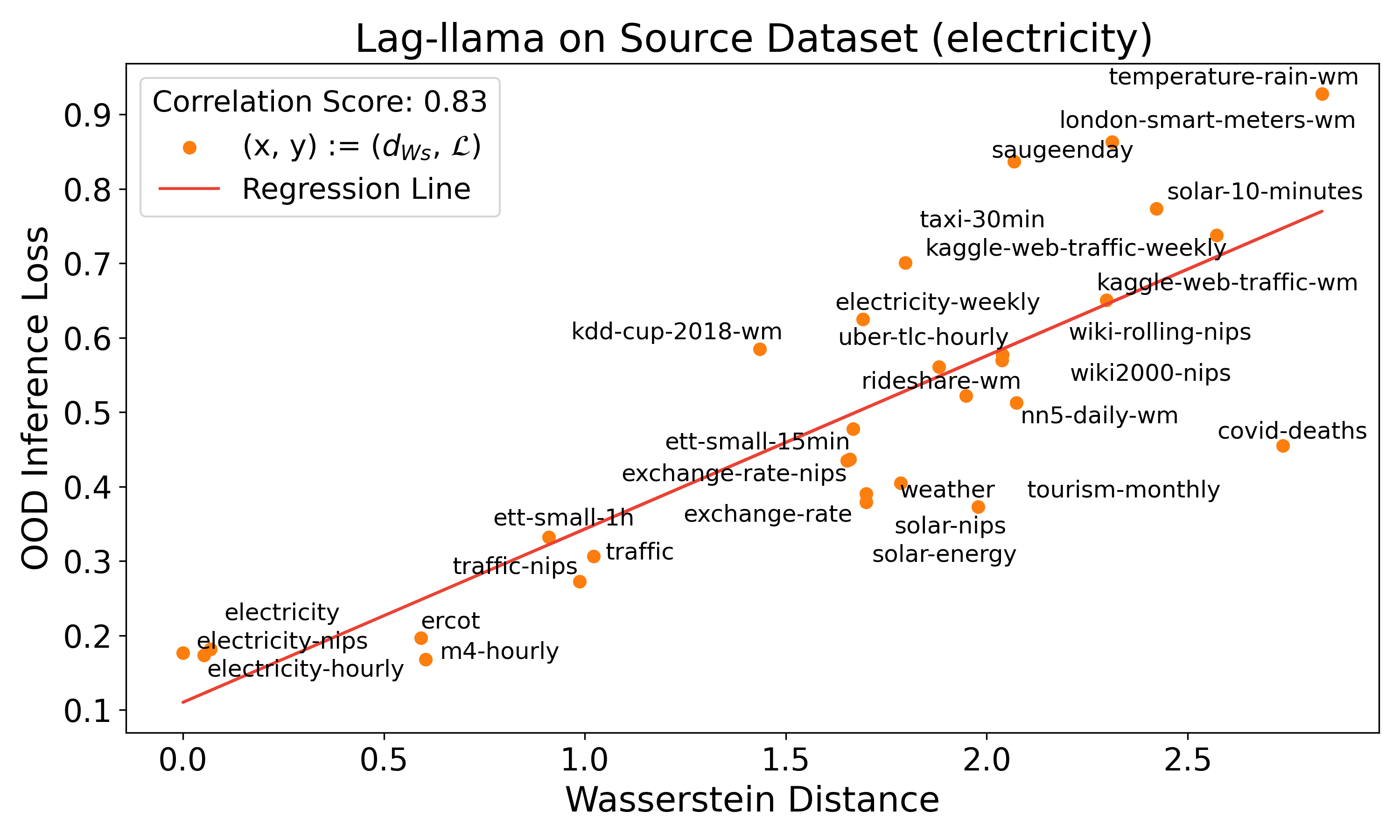}
\includegraphics[width=0.16\linewidth]{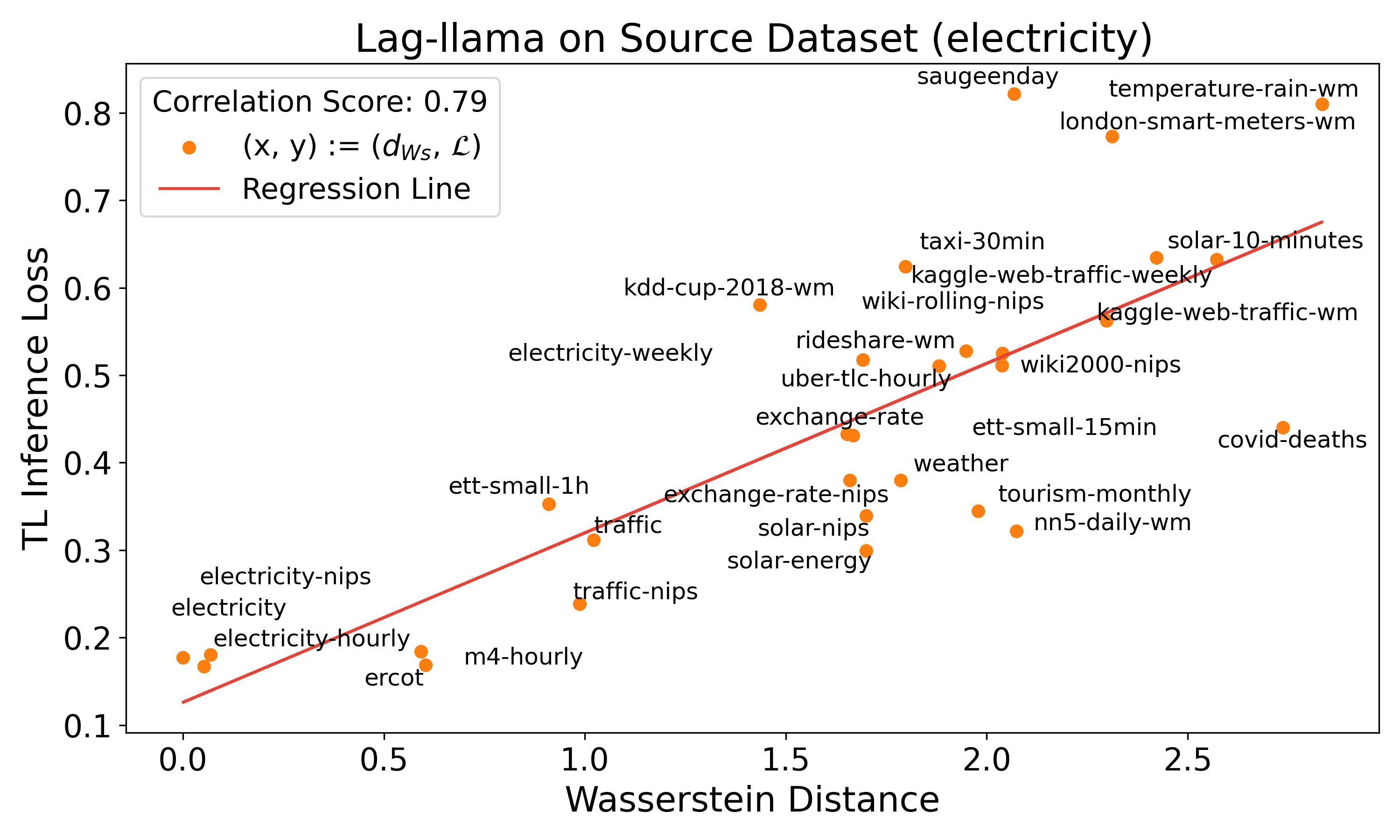}
\includegraphics[width=0.16\linewidth]{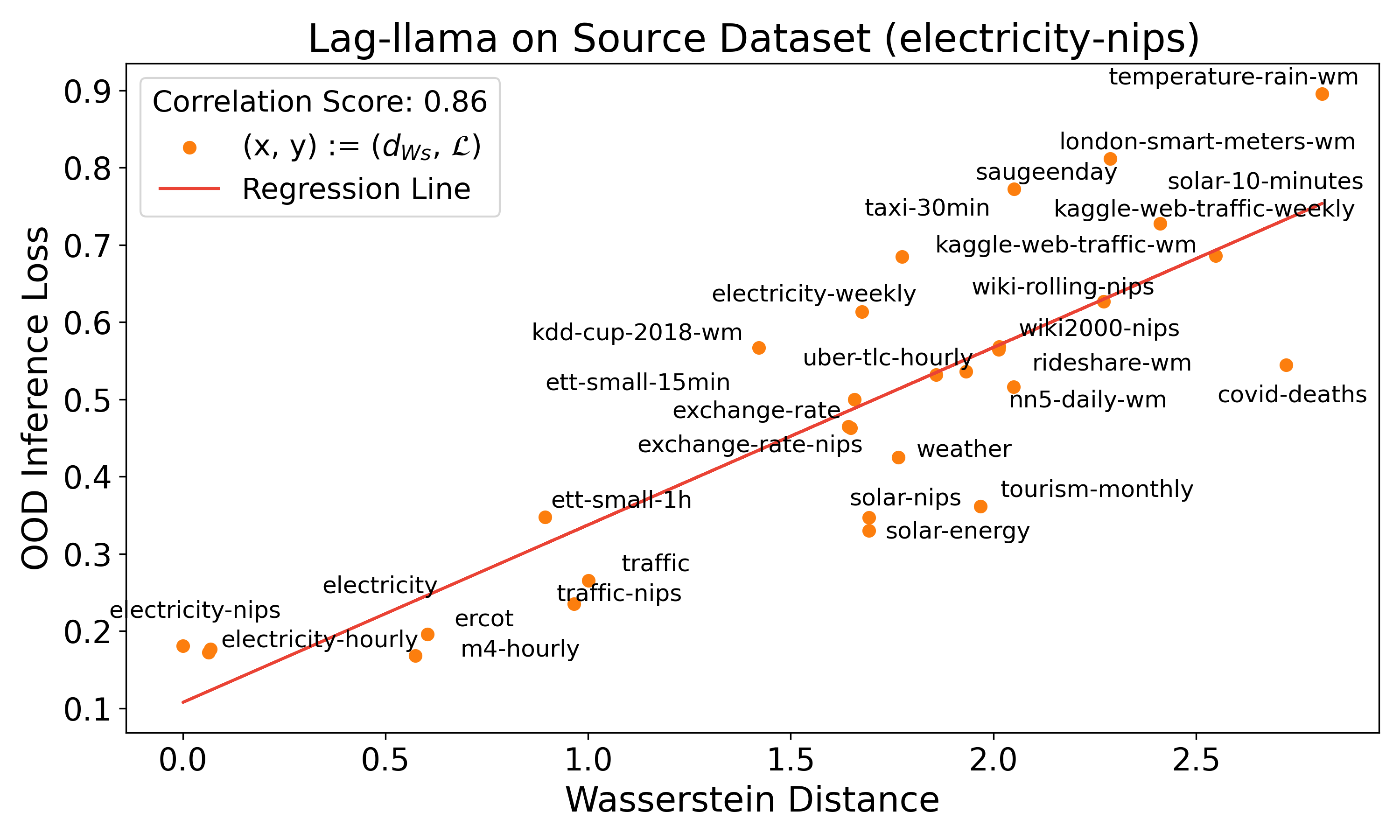}
\includegraphics[width=0.16\linewidth]{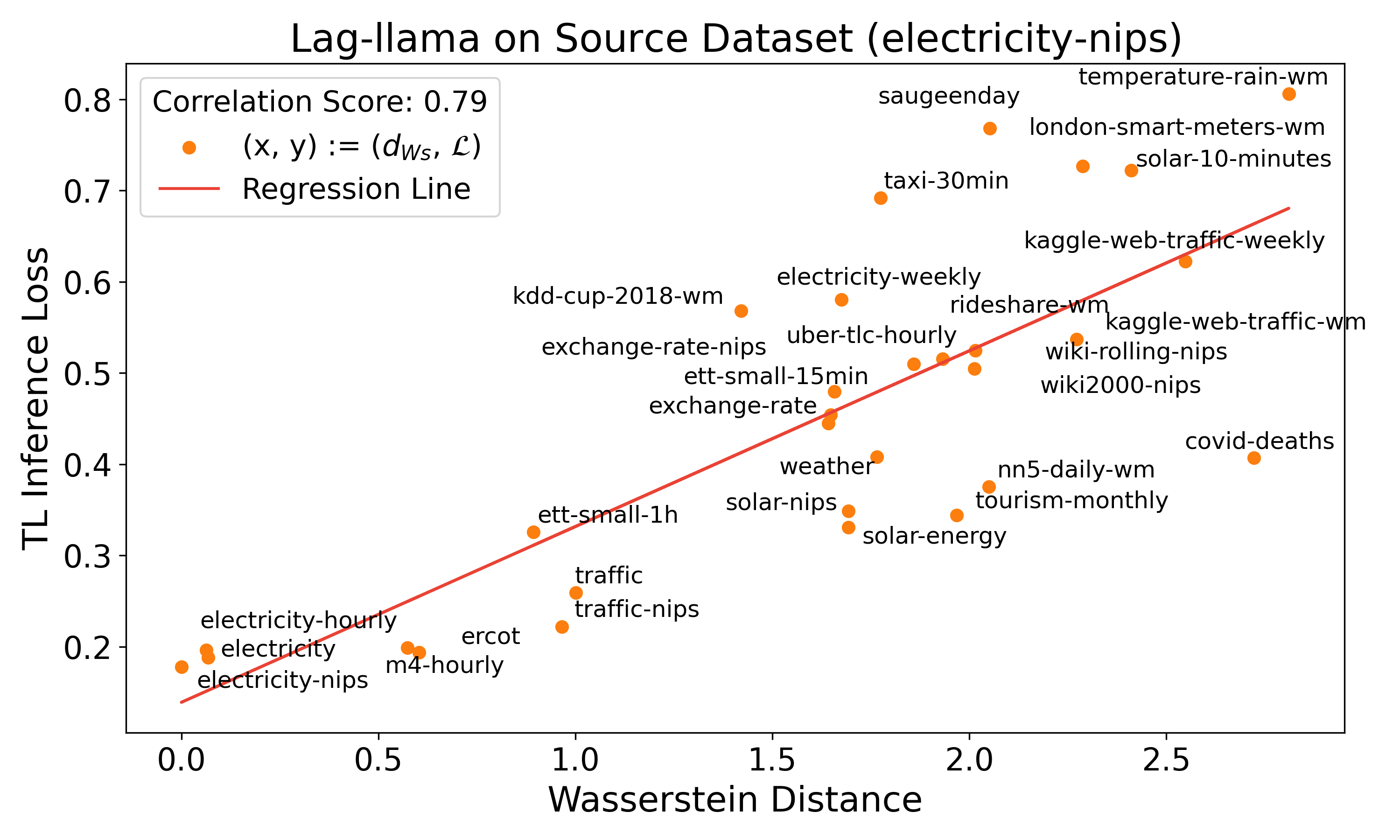}

\includegraphics[width=0.16\linewidth]{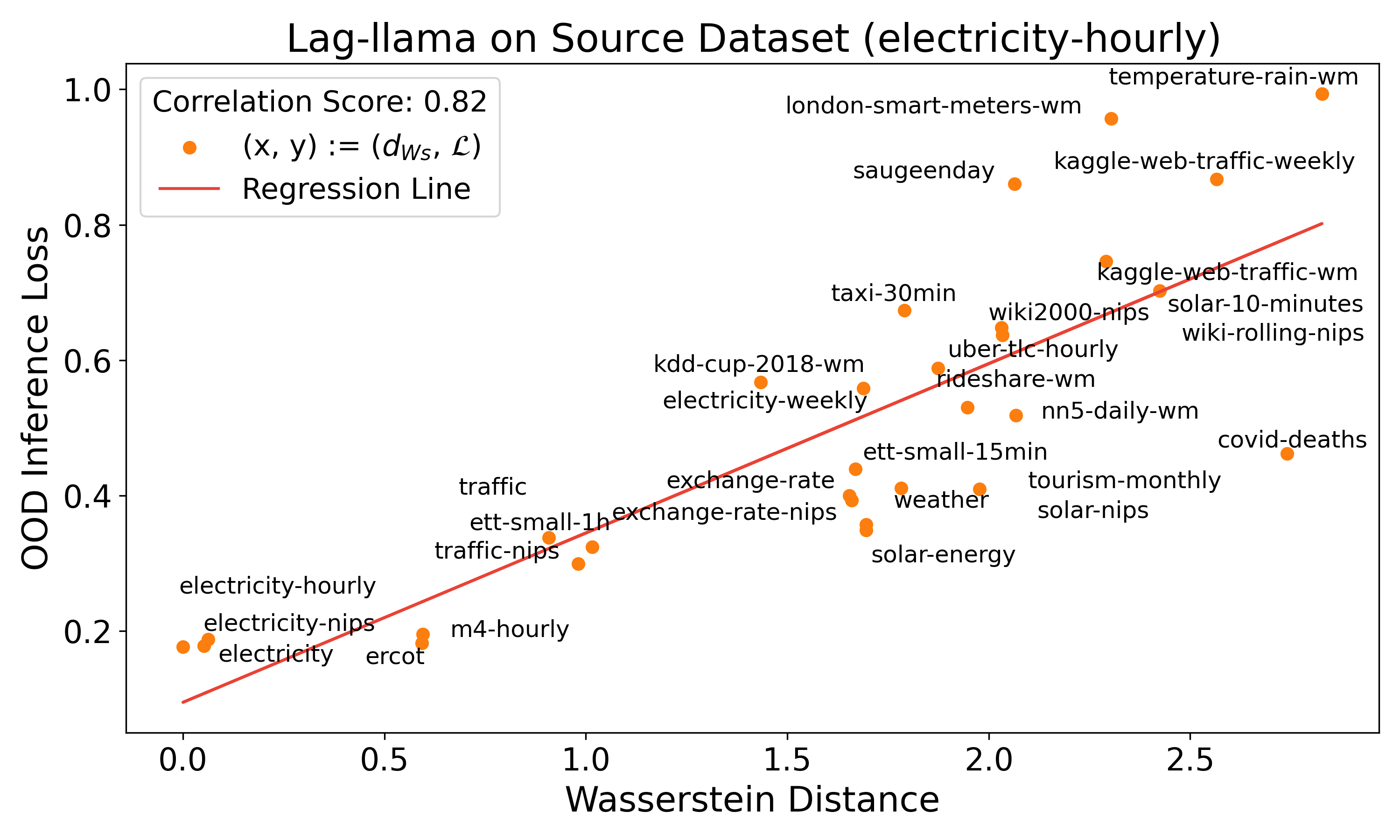}
\includegraphics[width=0.16\linewidth]{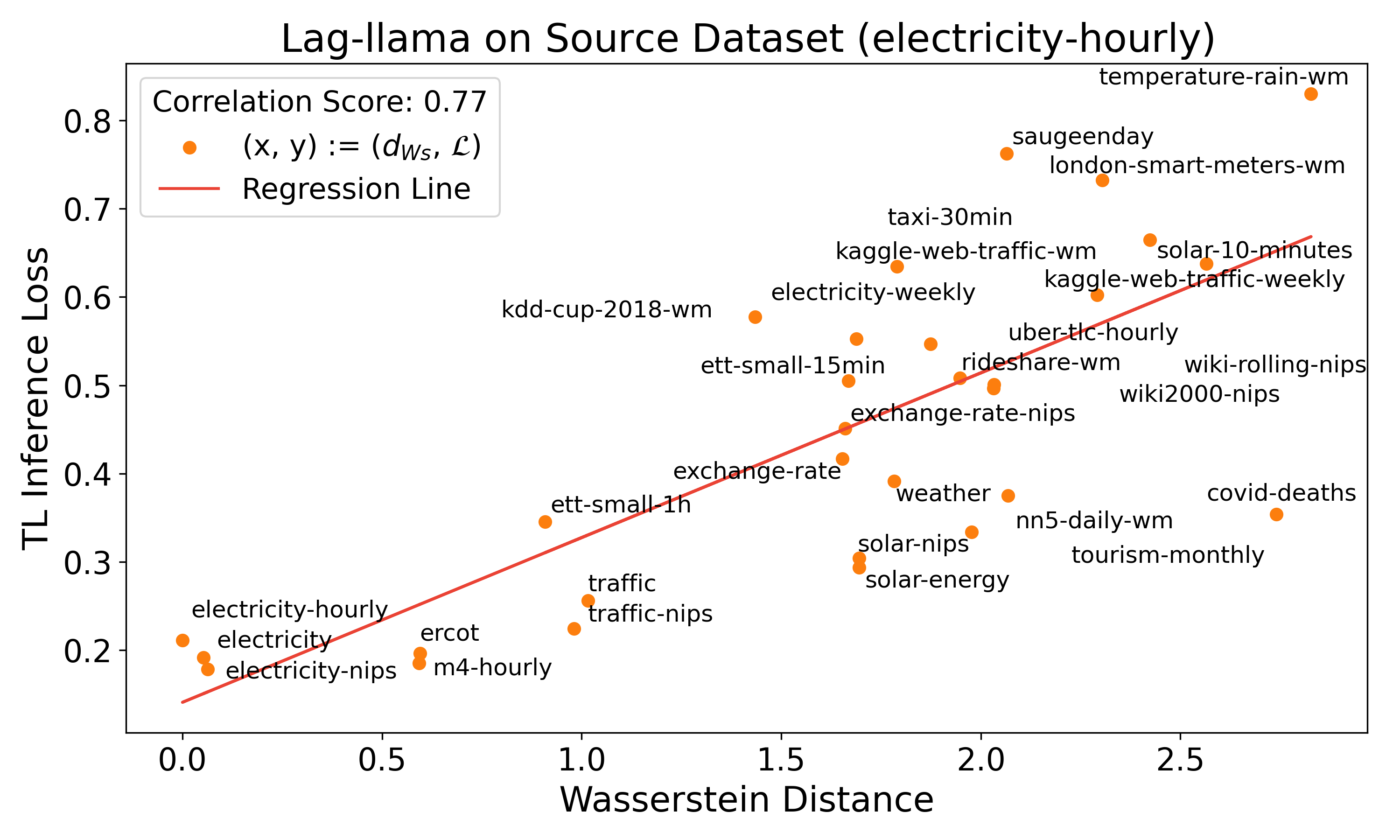}
\includegraphics[width=0.16\linewidth]{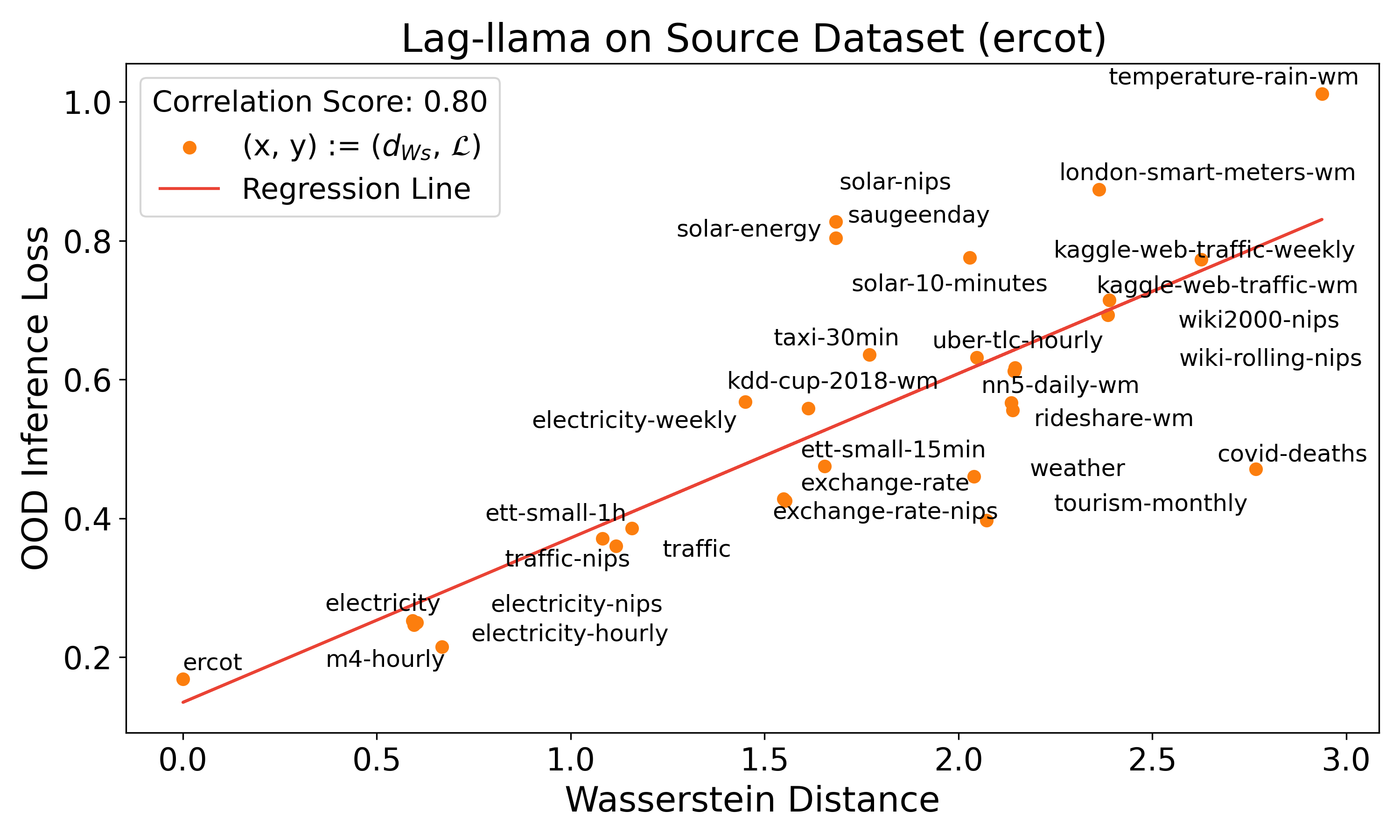}
\includegraphics[width=0.16\linewidth]{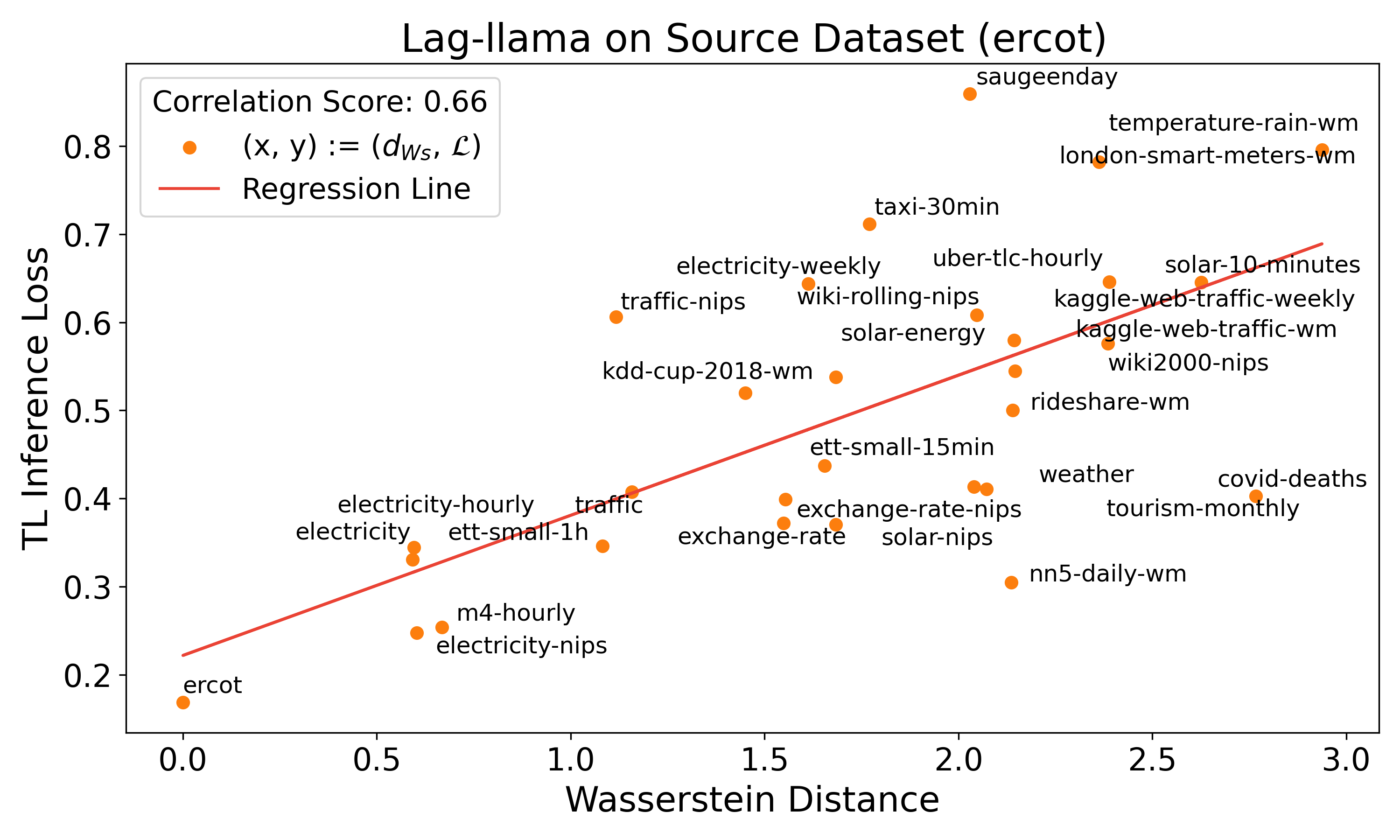}
\includegraphics[width=0.16\linewidth]{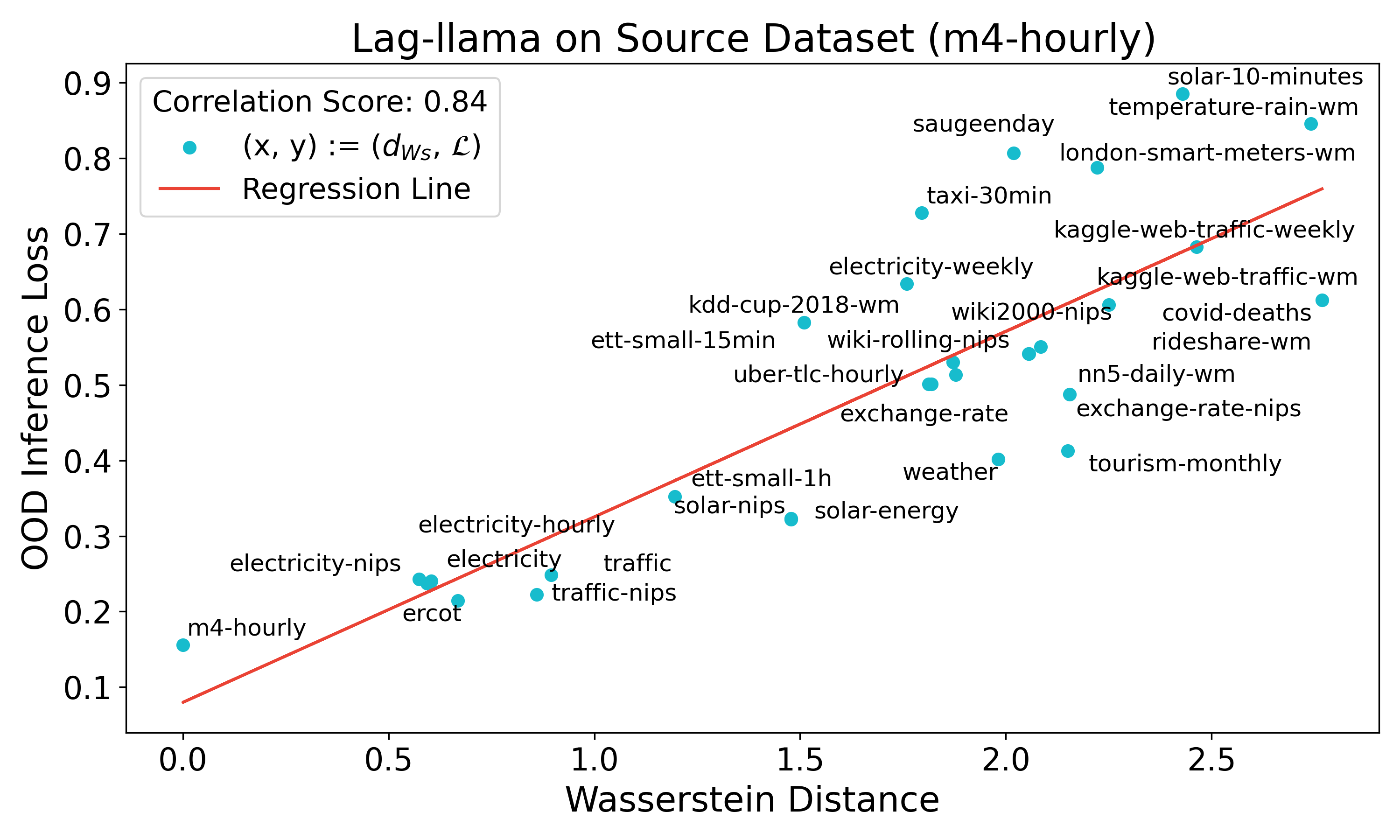}
\includegraphics[width=0.16\linewidth]{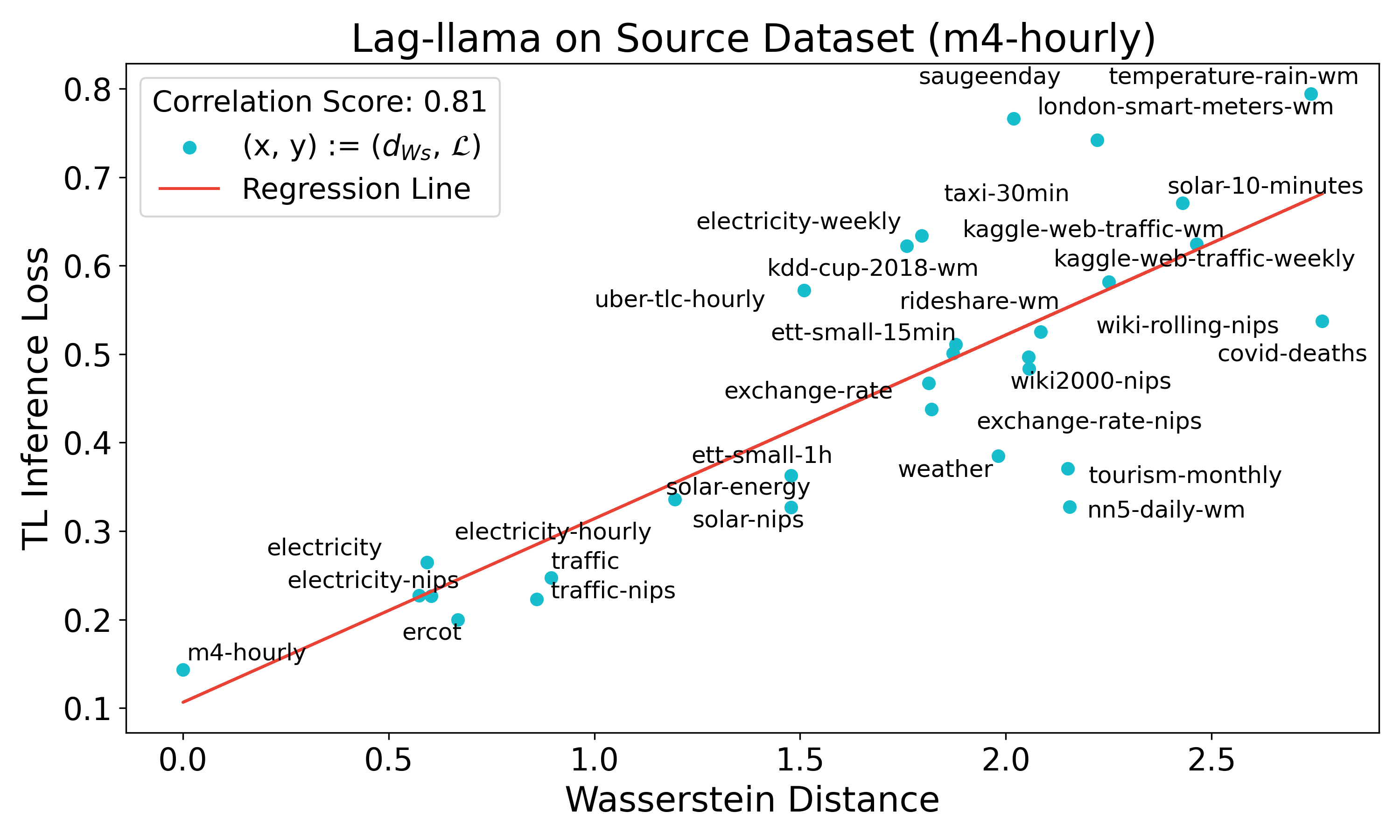}

\includegraphics[width=0.16\linewidth]{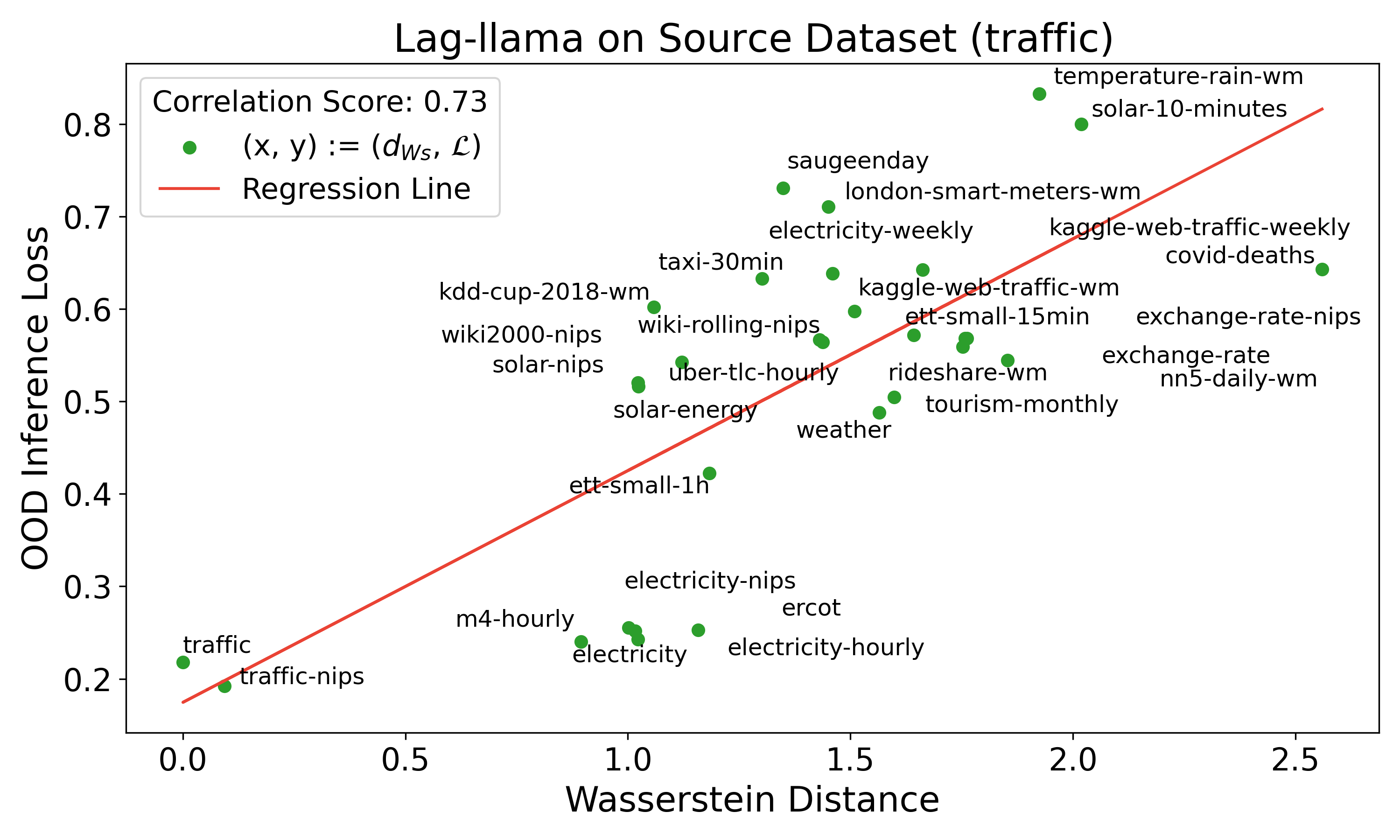}
\includegraphics[width=0.16\linewidth]{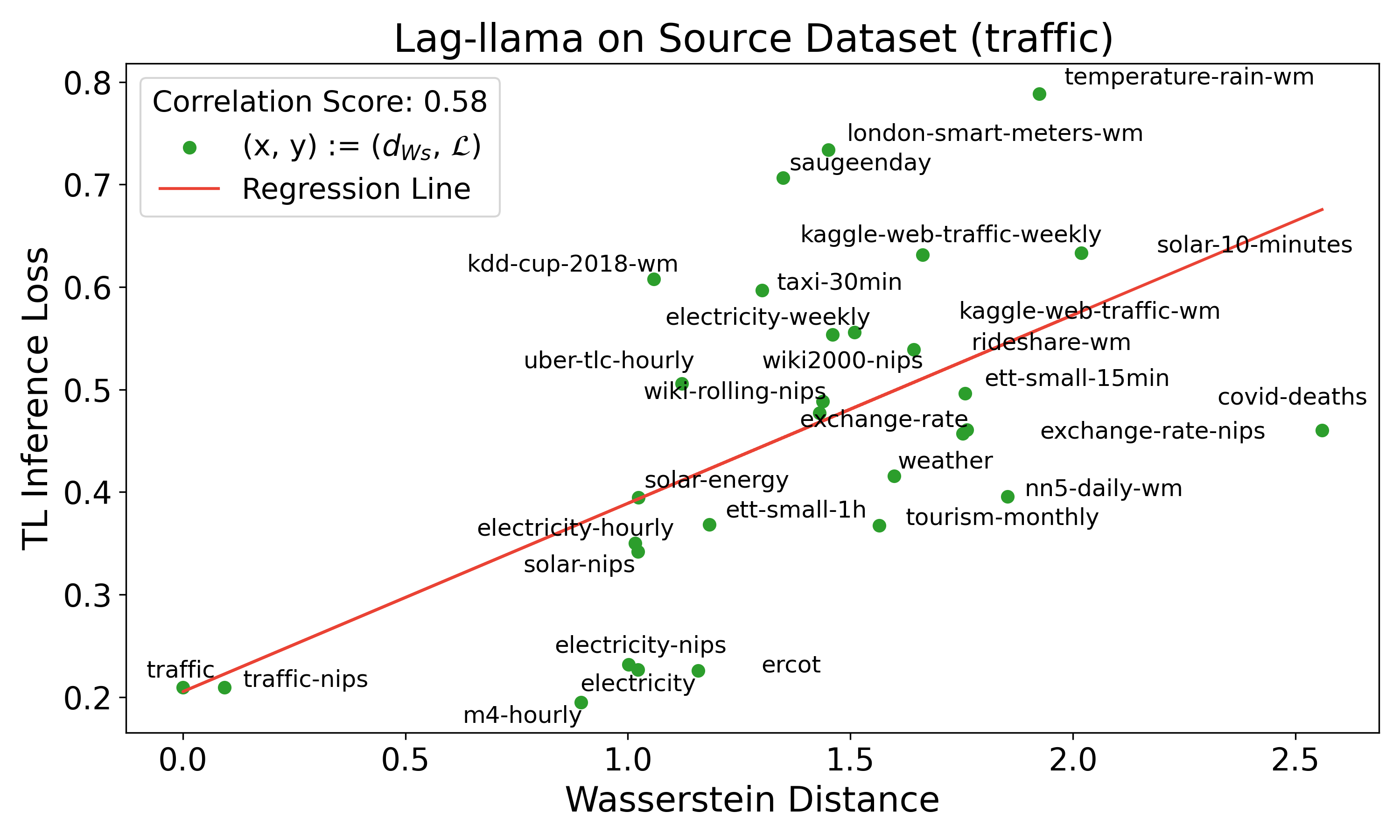}
\includegraphics[width=0.16\linewidth]{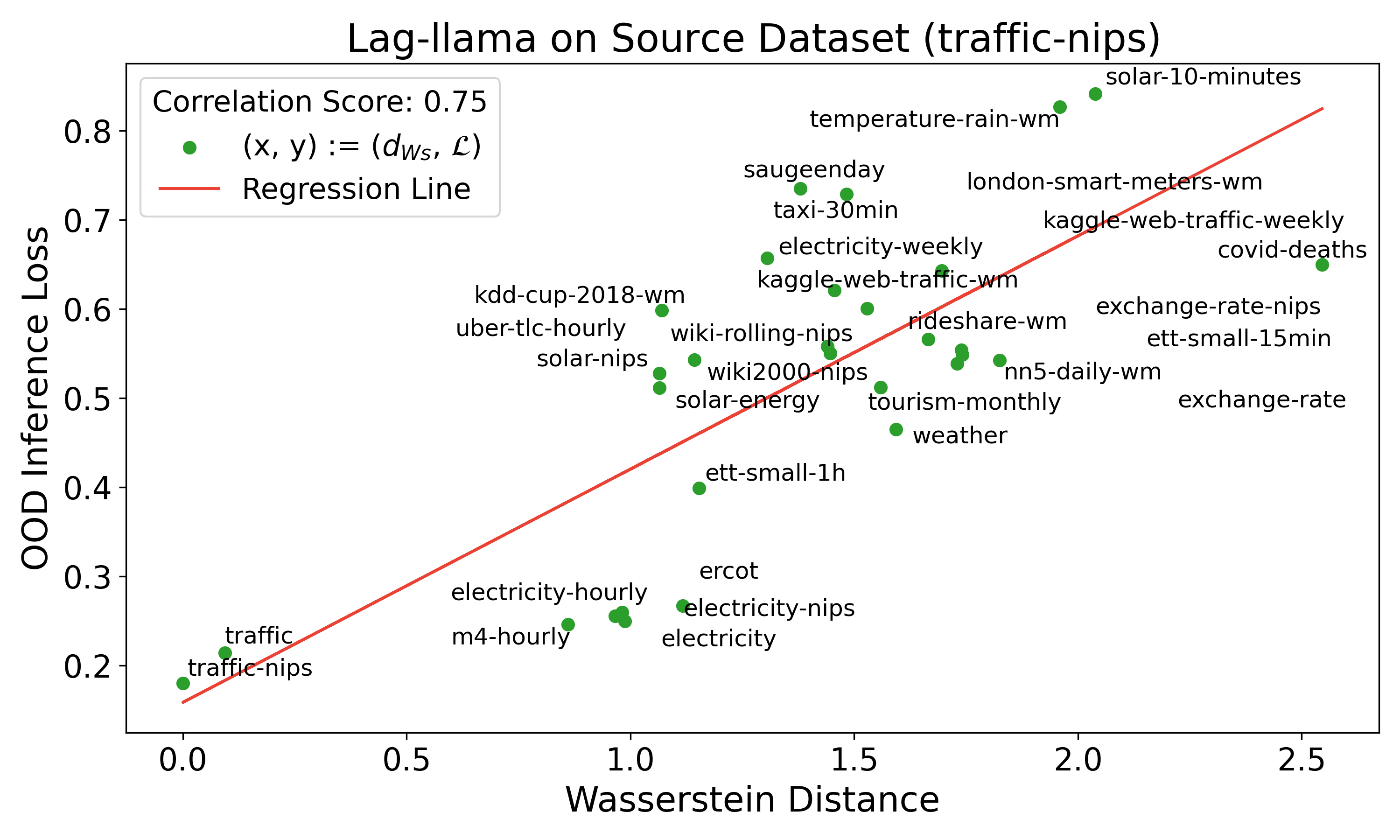}
\includegraphics[width=0.16\linewidth]{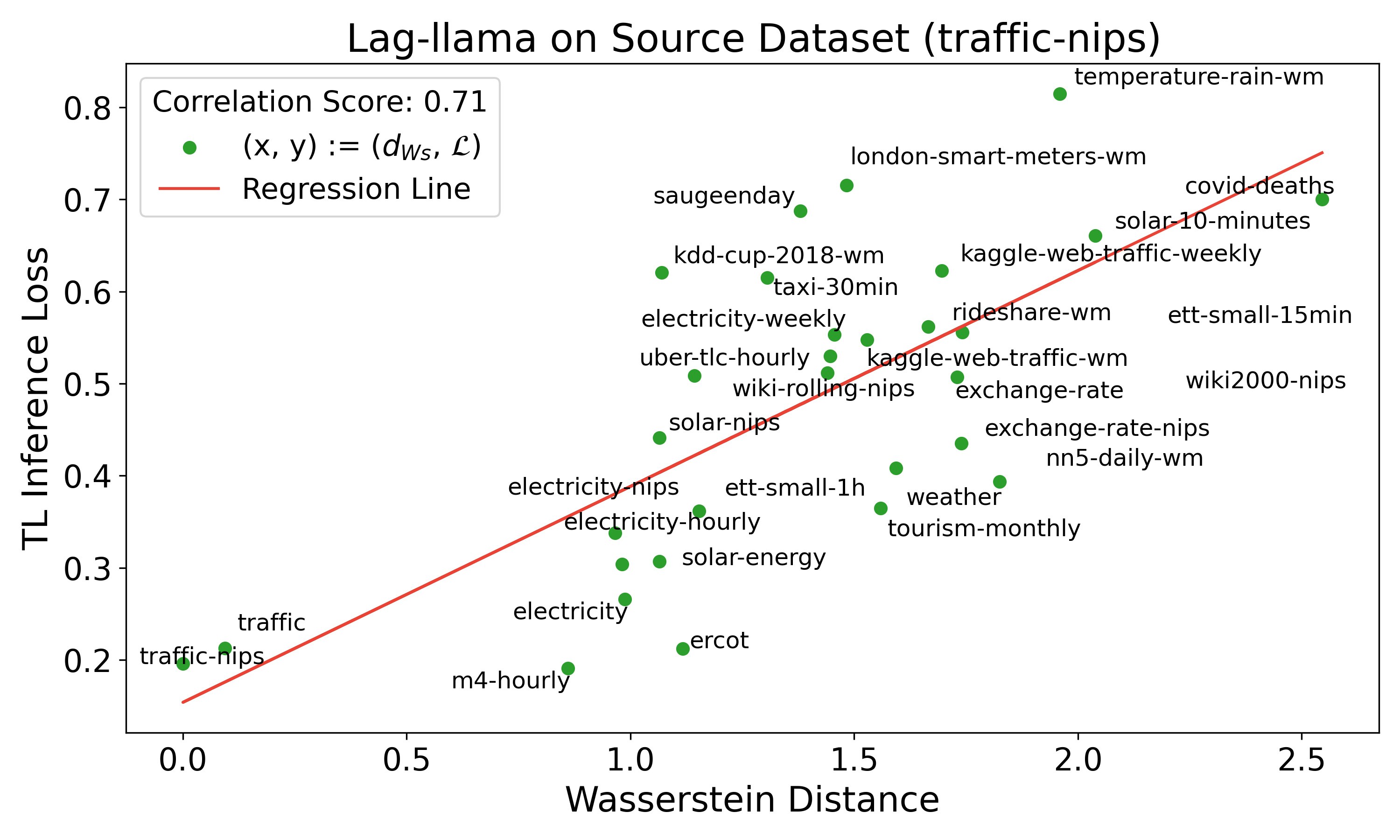}
\includegraphics[width=0.16\linewidth]{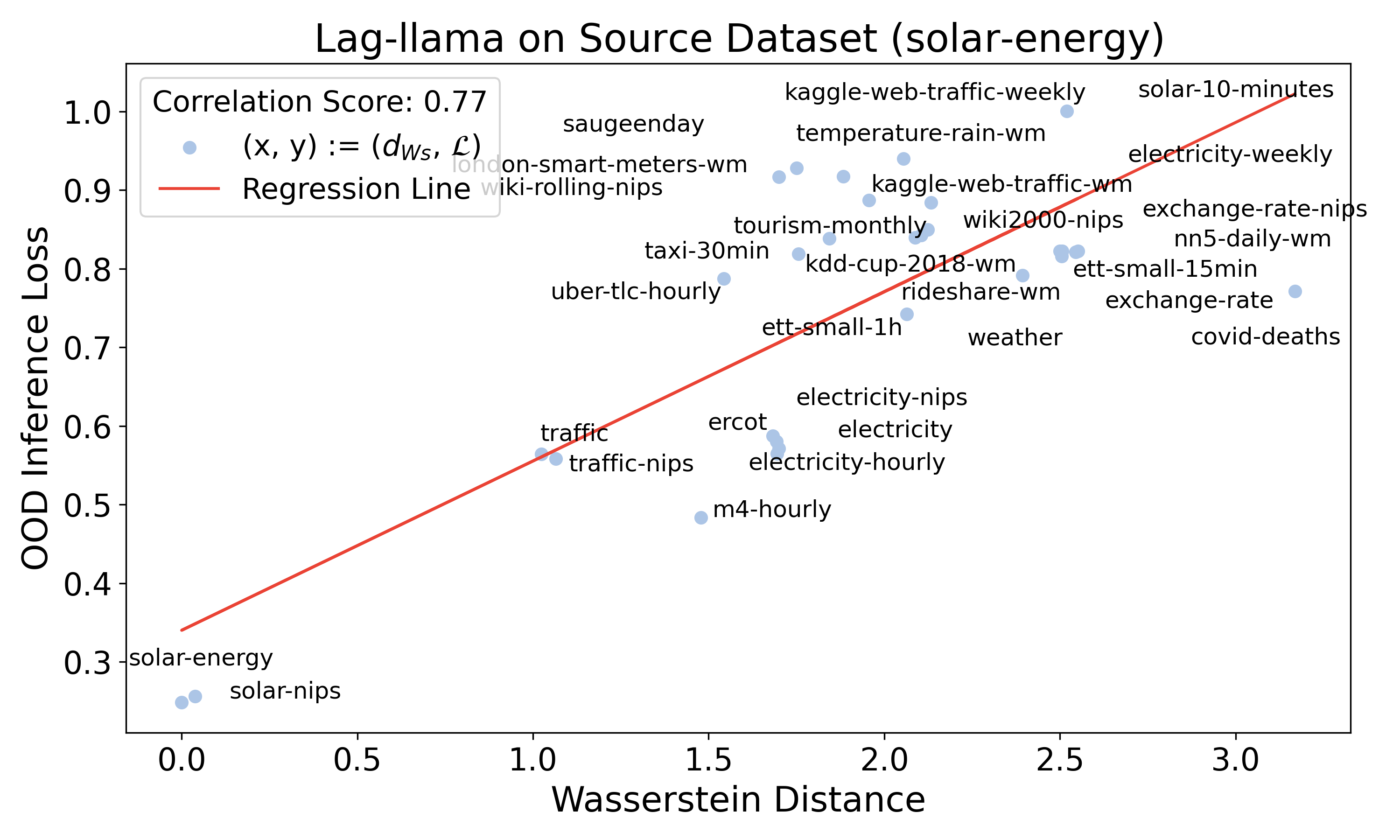}
\includegraphics[width=0.16\linewidth]{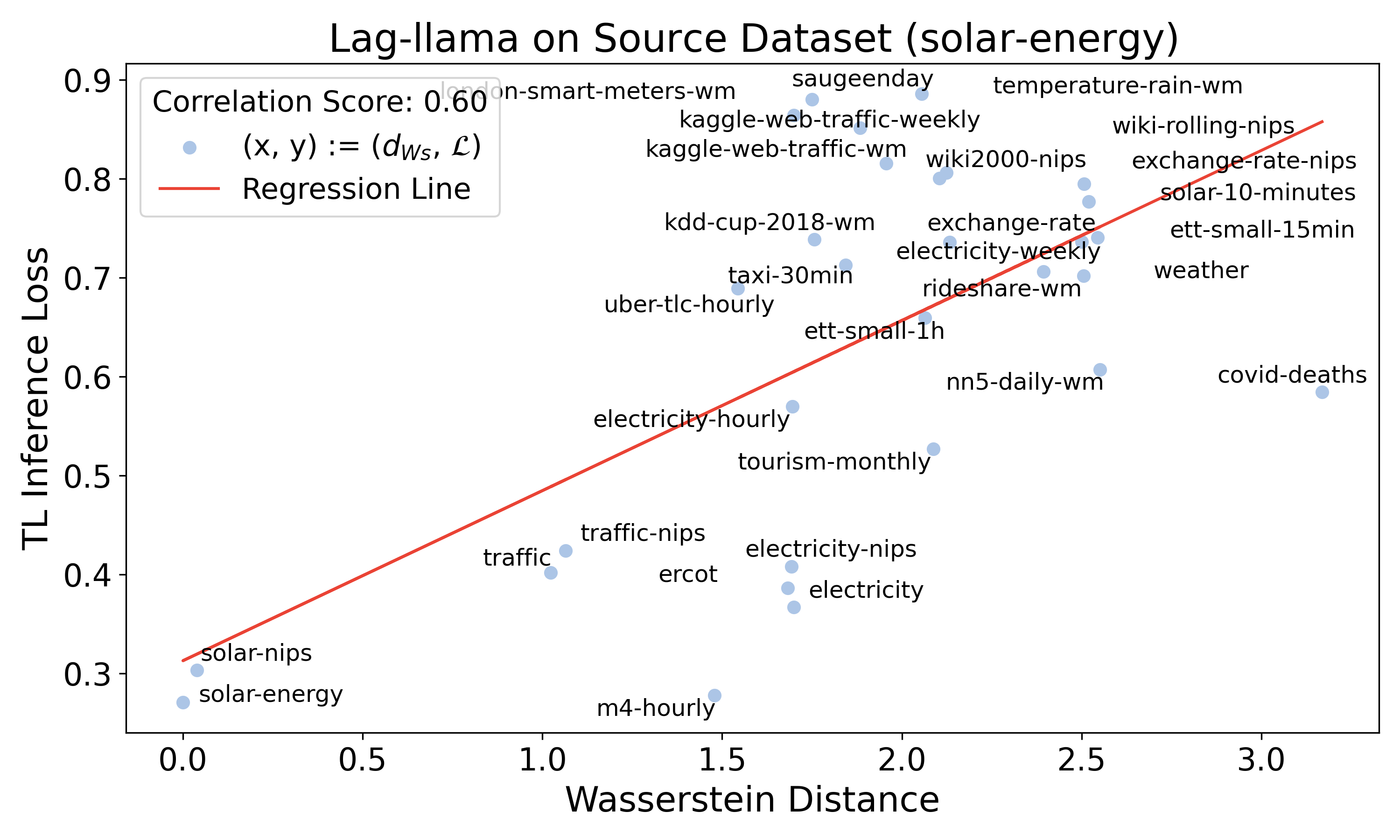}

\includegraphics[width=0.16\linewidth]{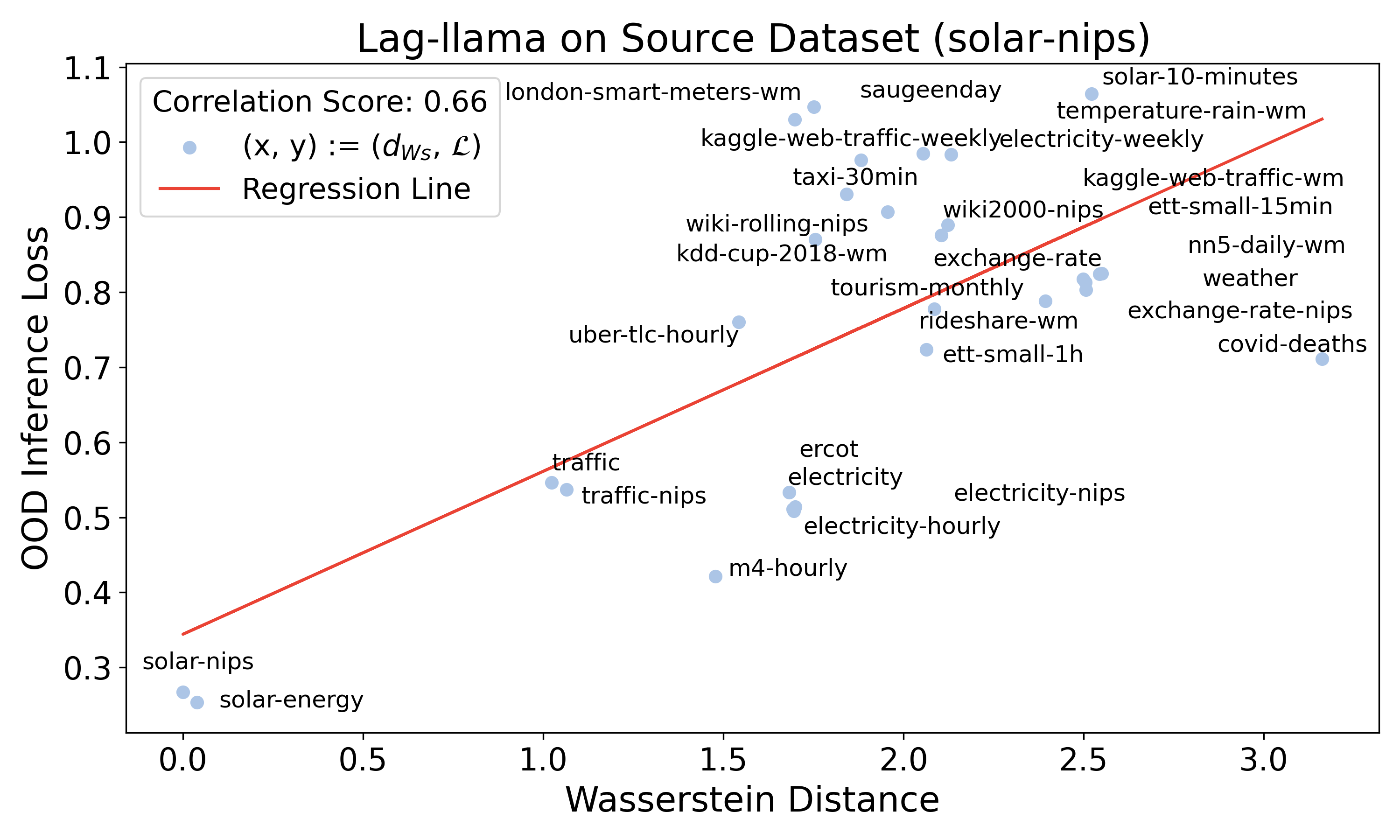}
\includegraphics[width=0.16\linewidth]{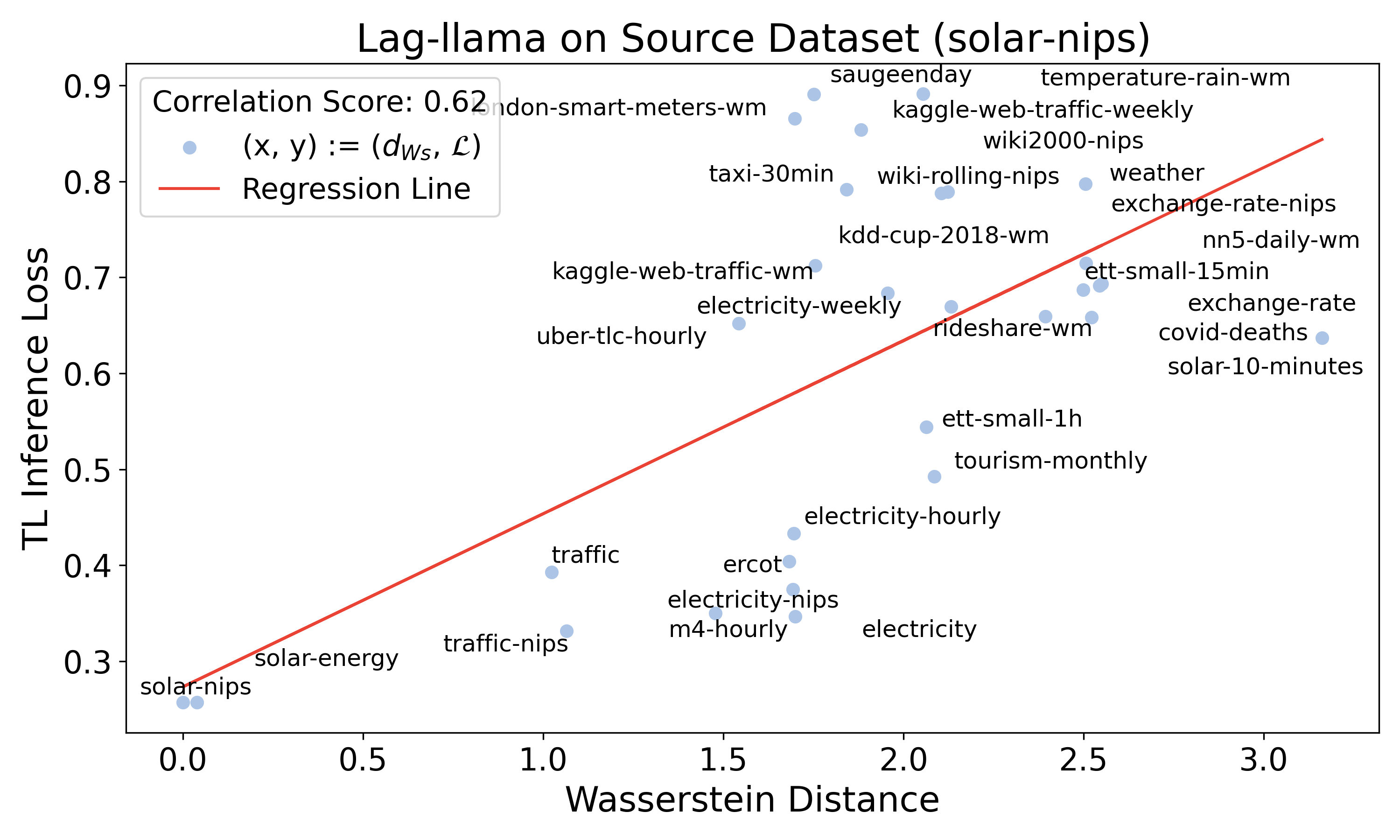}
\includegraphics[width=0.16\linewidth]{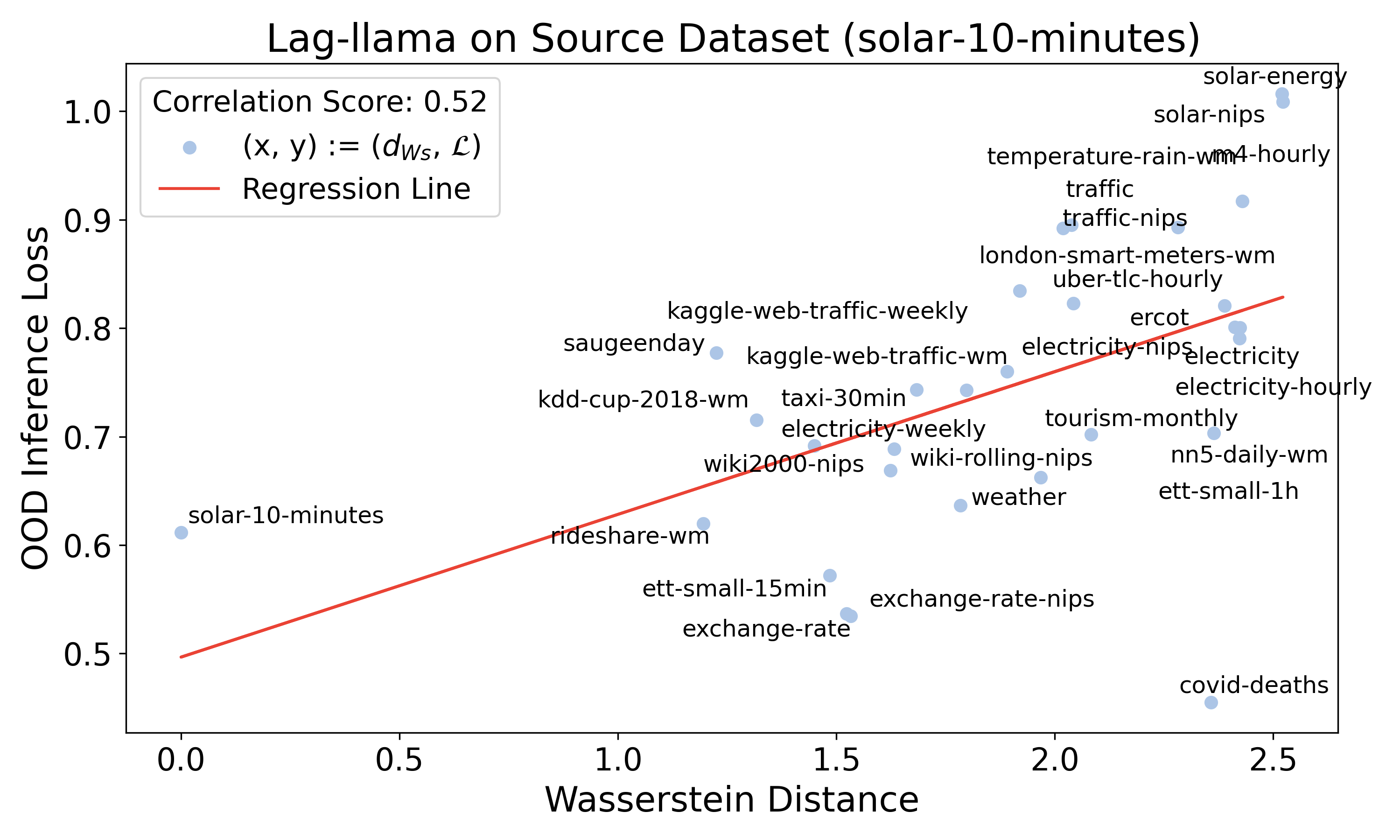}
\includegraphics[width=0.16\linewidth]{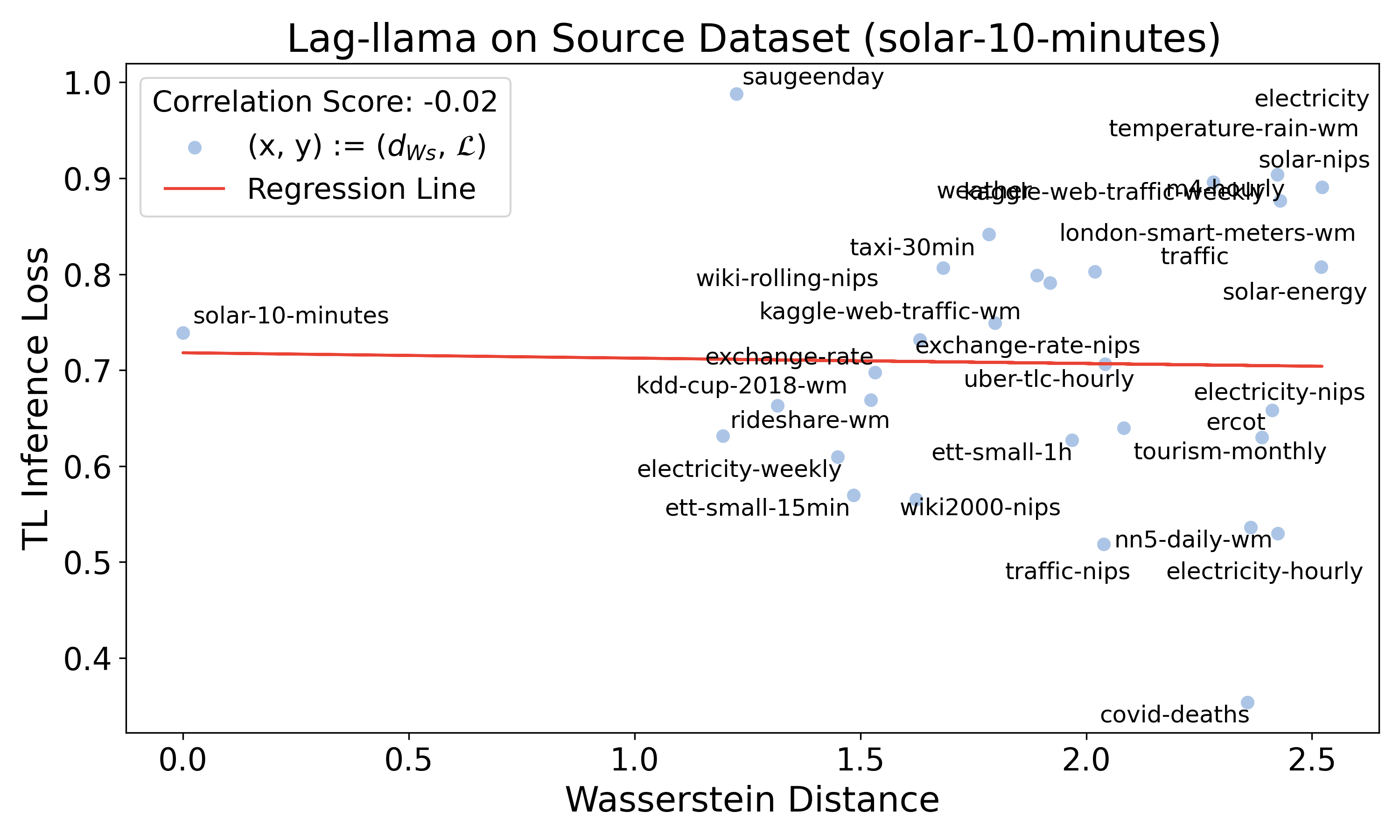}
\includegraphics[width=0.16\linewidth]{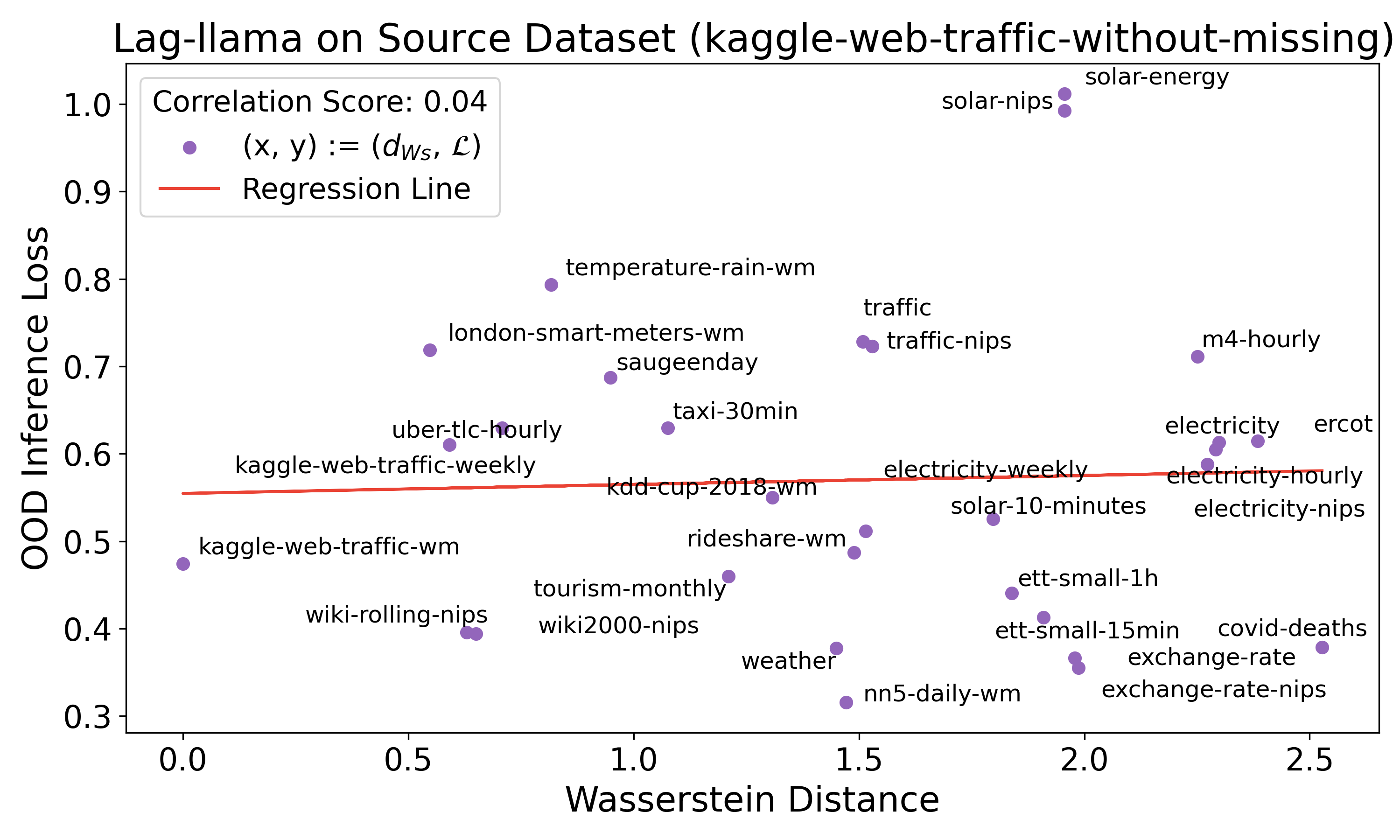}
\includegraphics[width=0.16\linewidth]{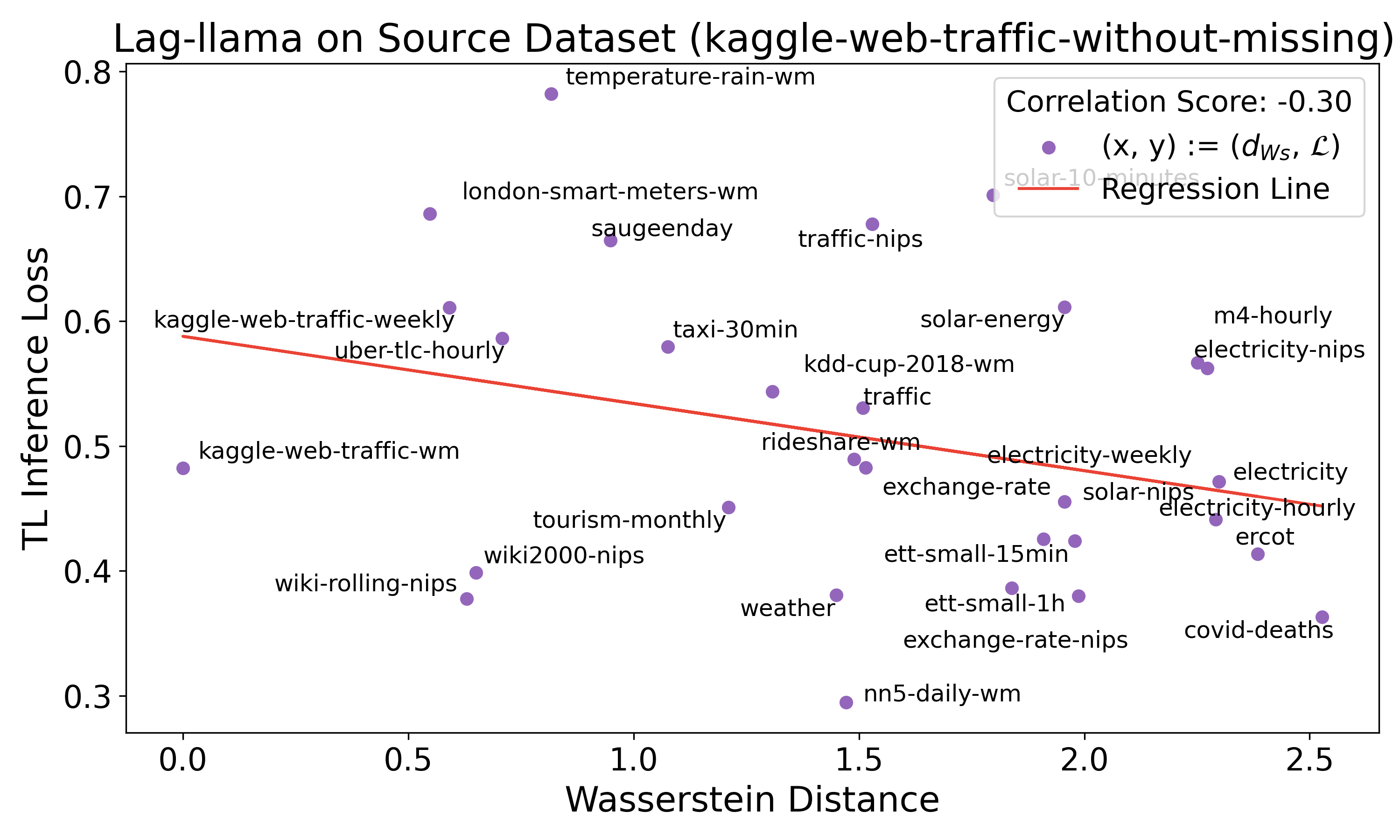}

\includegraphics[width=0.16\linewidth]{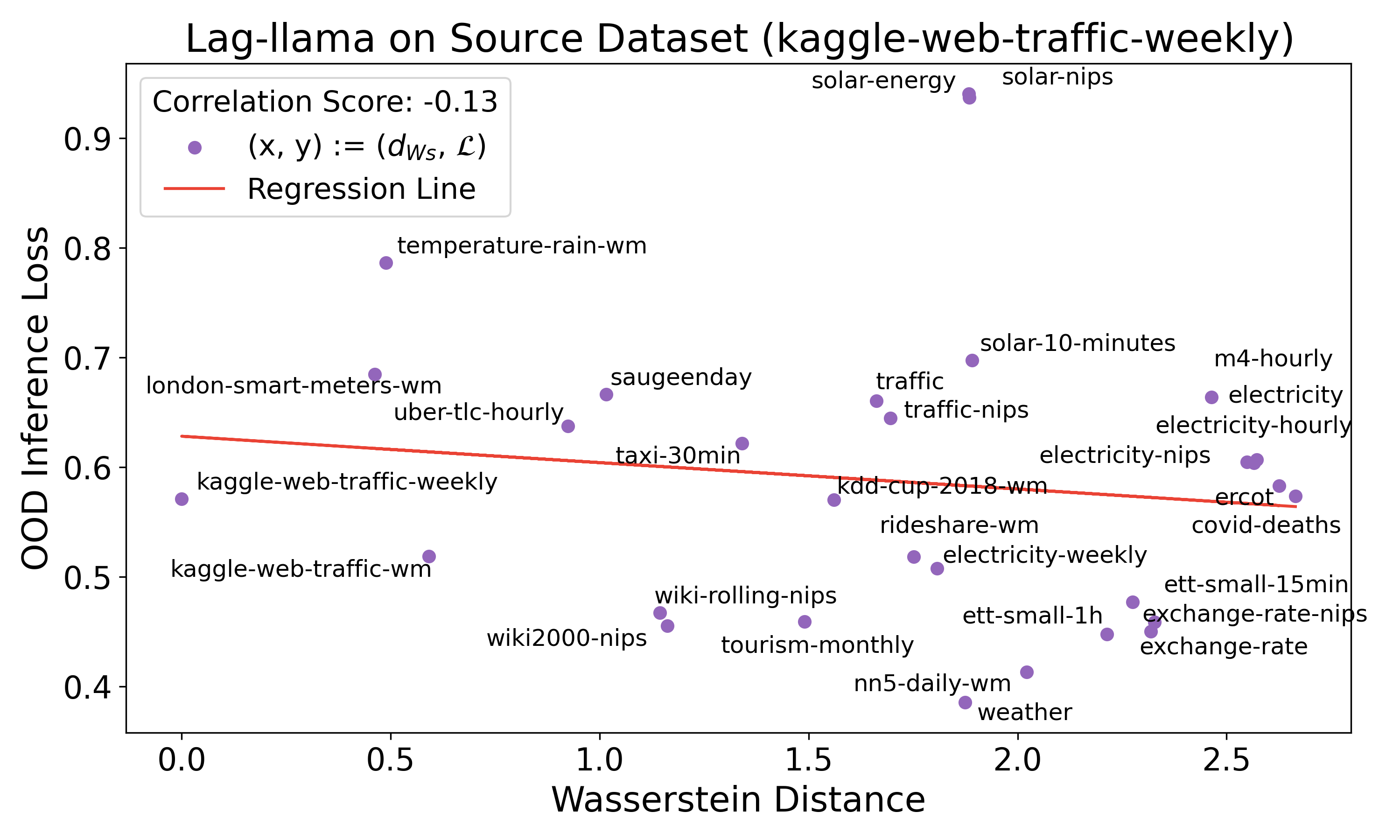}
\includegraphics[width=0.16\linewidth]{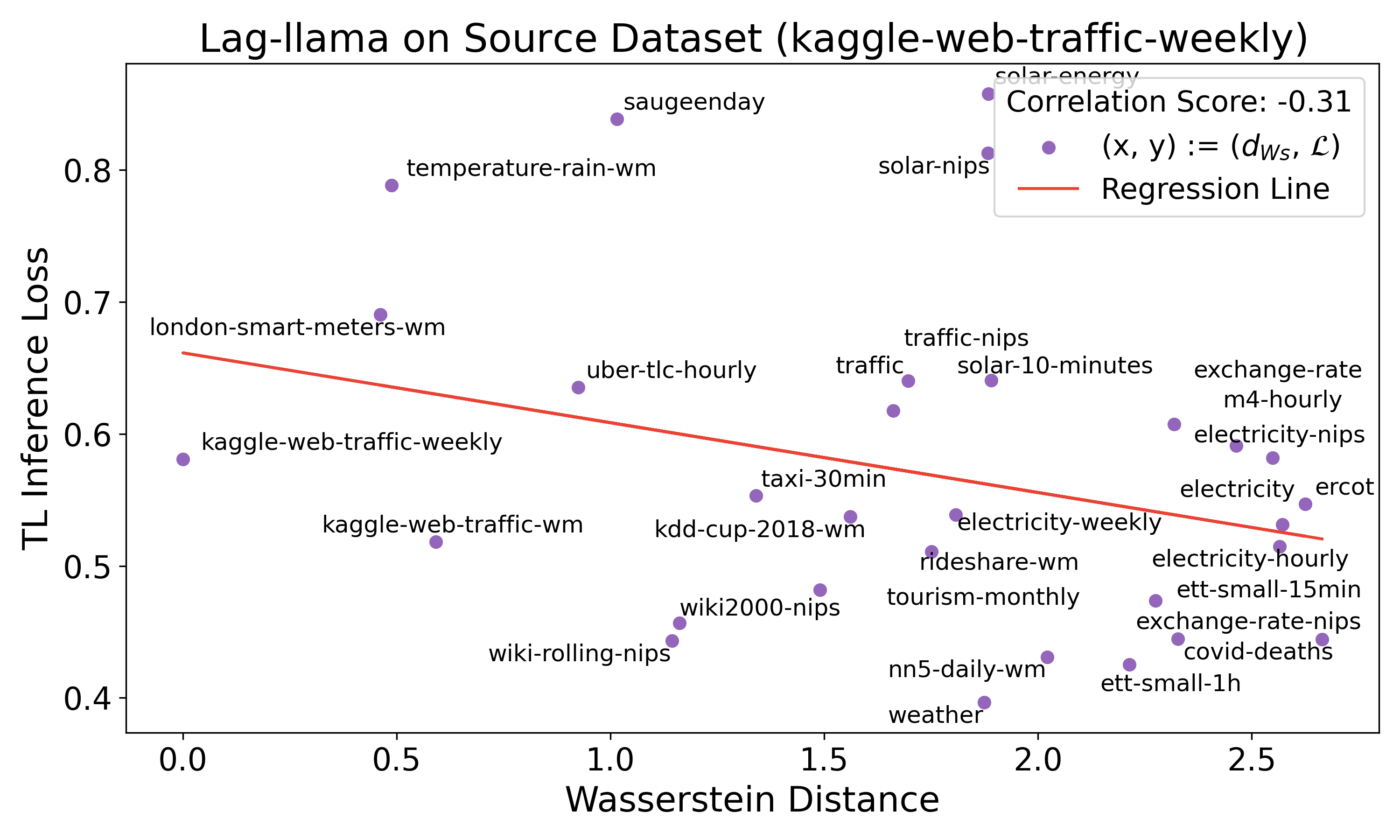}
\includegraphics[width=0.16\linewidth]{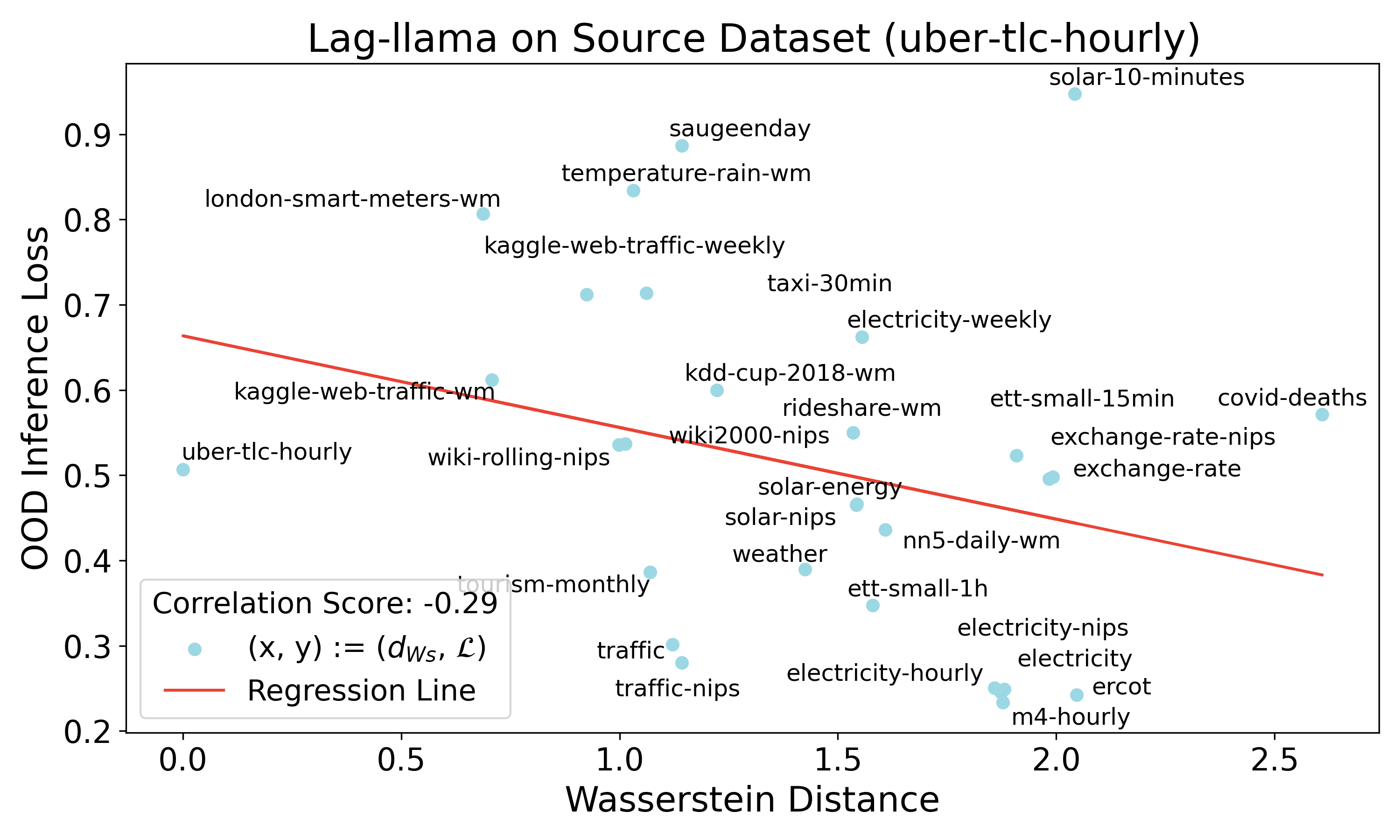}
\includegraphics[width=0.16\linewidth]{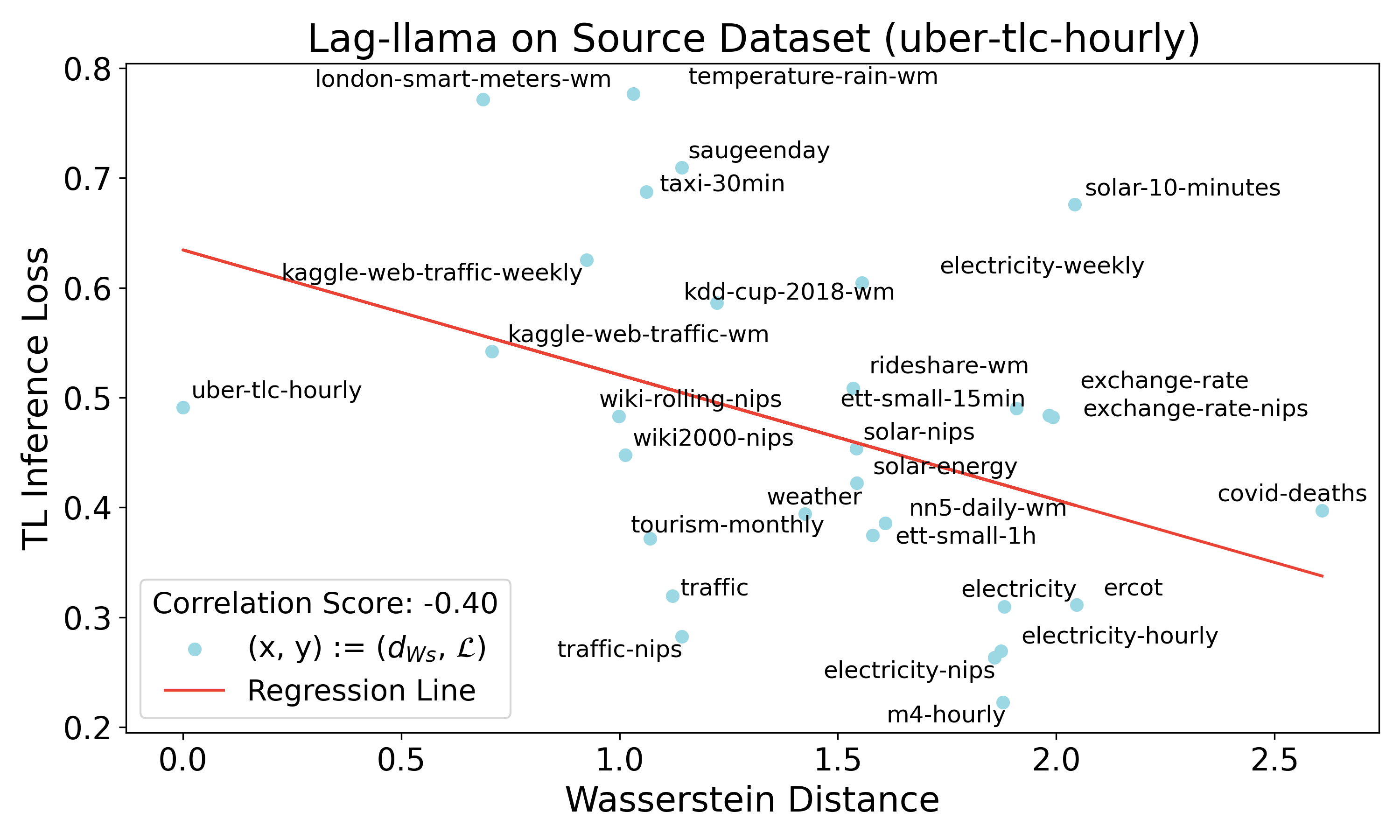}
\includegraphics[width=0.16\linewidth]{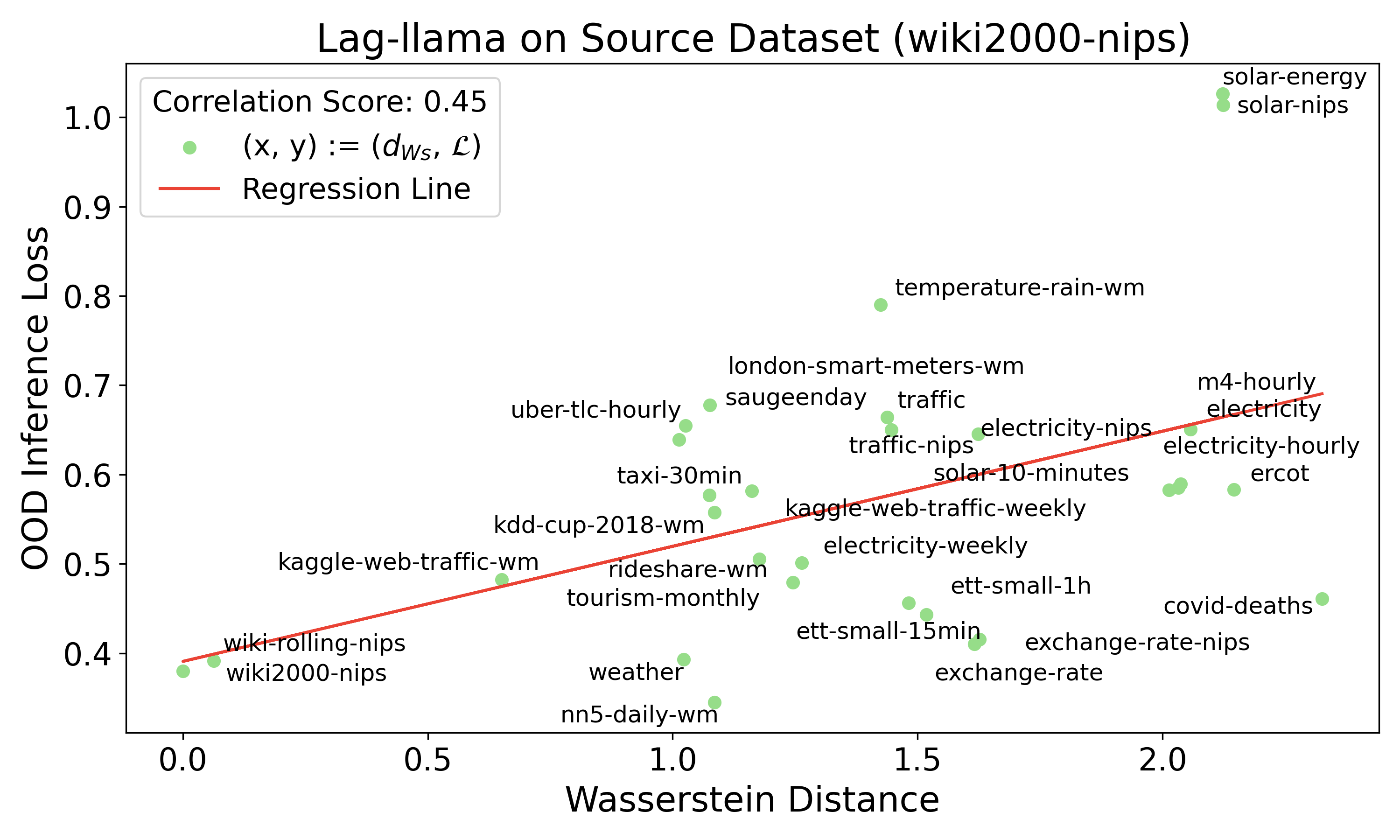}
\includegraphics[width=0.16\linewidth]{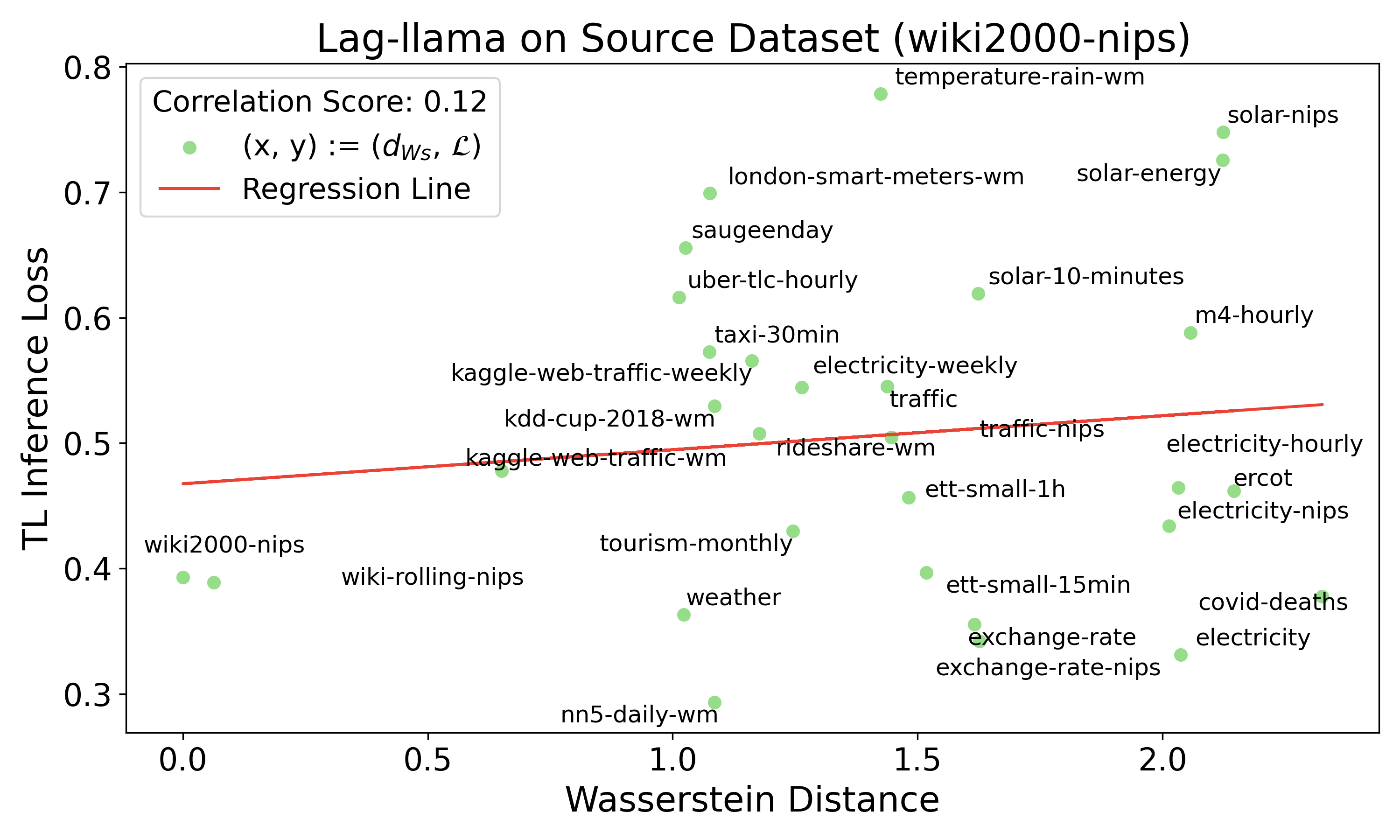}

\includegraphics[width=0.16\linewidth]{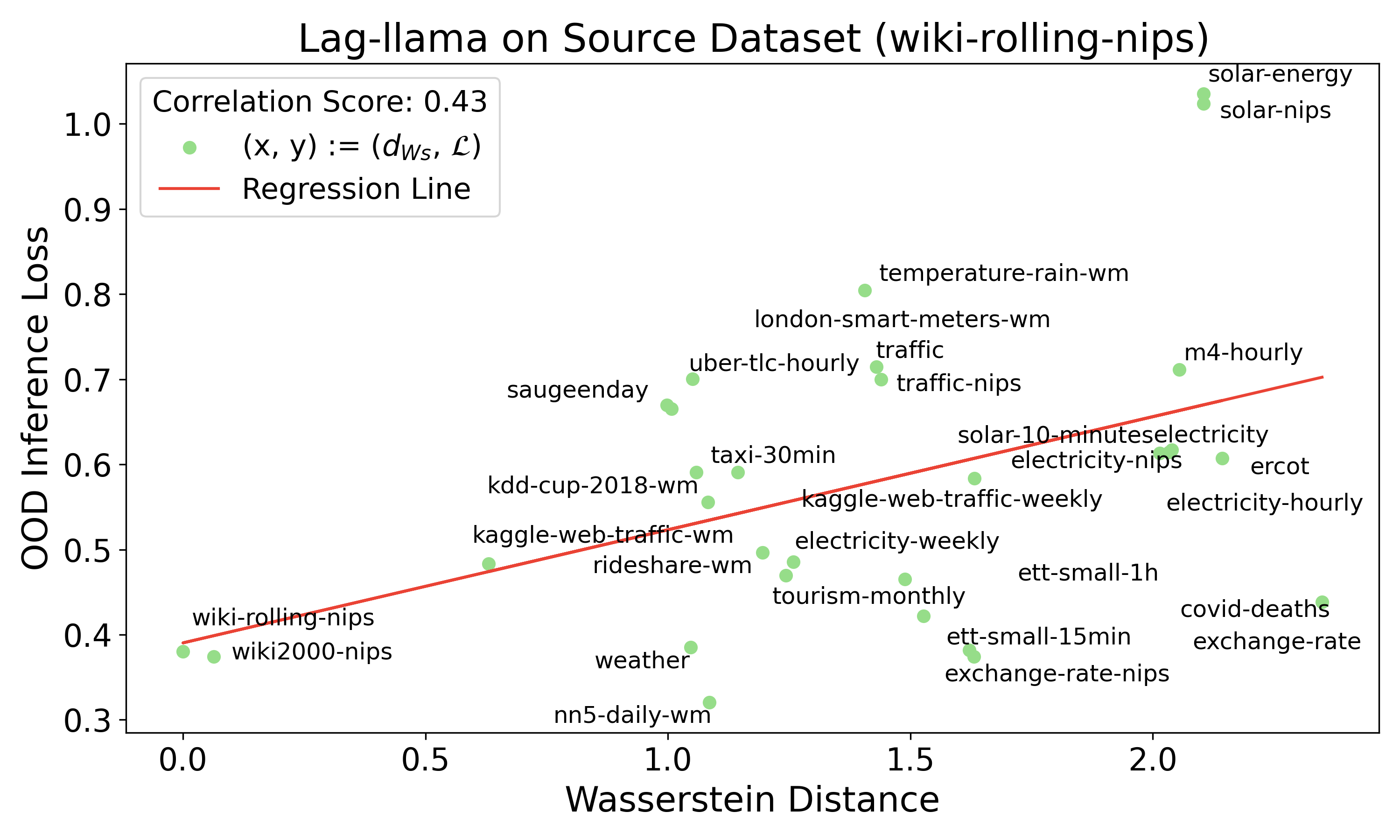}
\includegraphics[width=0.16\linewidth]{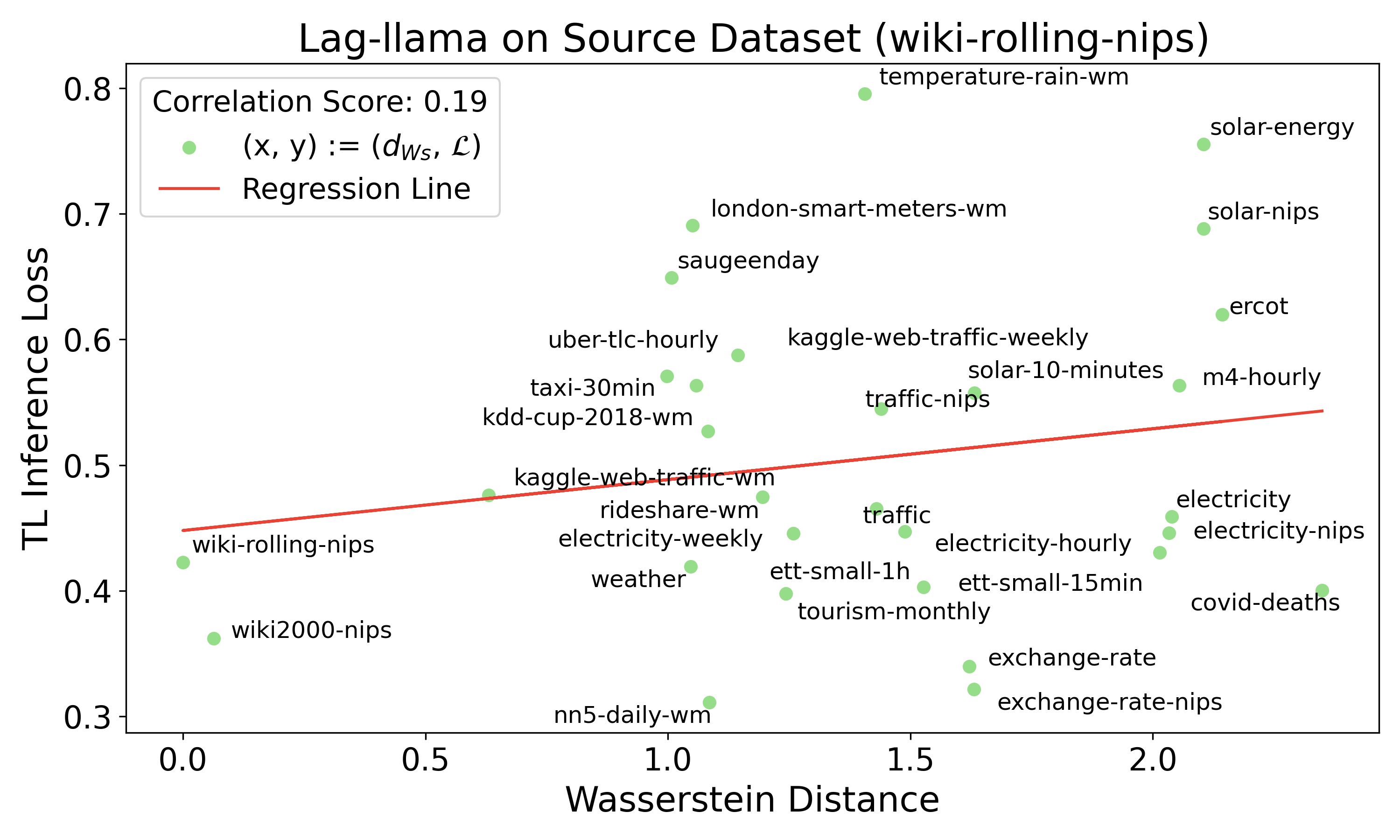}
\includegraphics[width=0.16\linewidth]{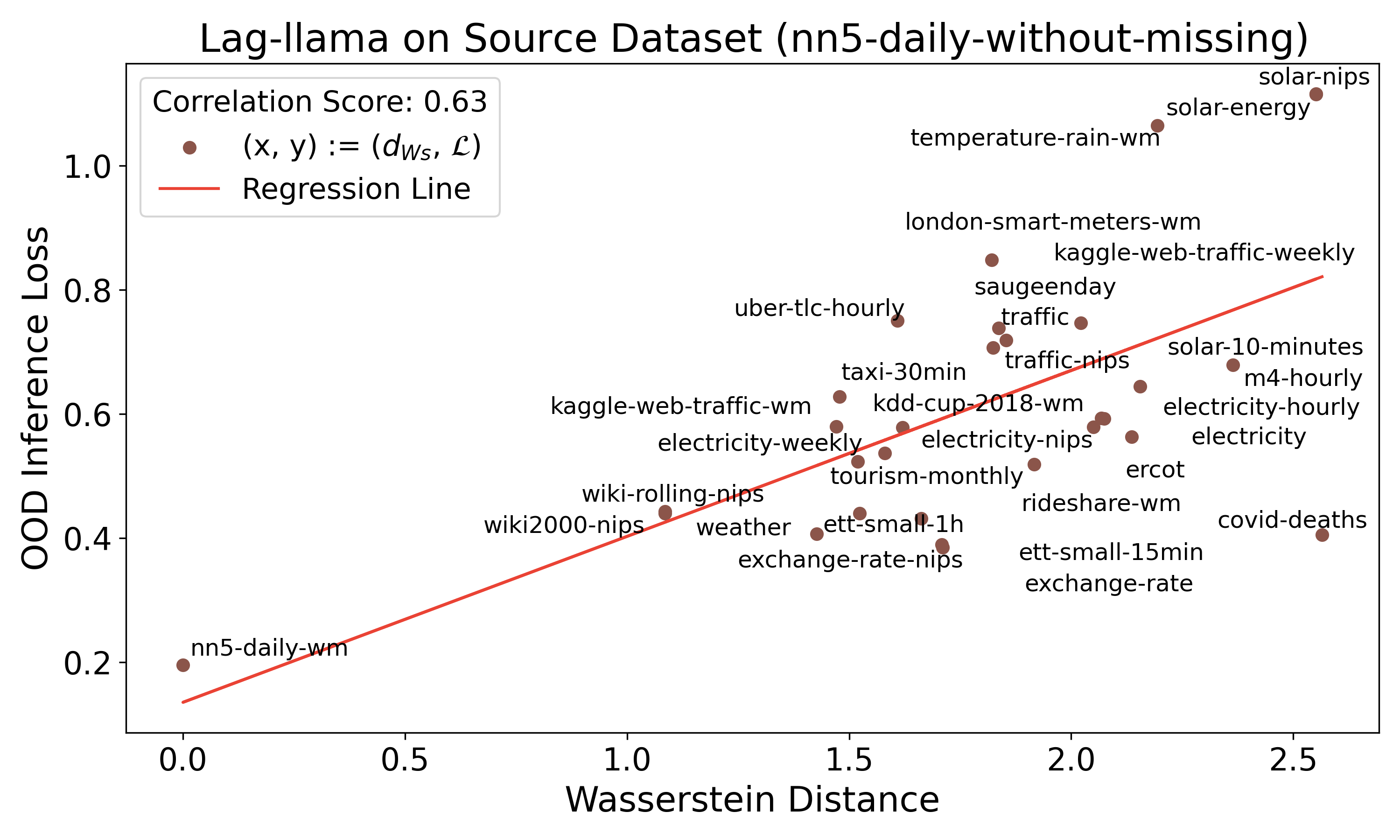}
\includegraphics[width=0.16\linewidth]{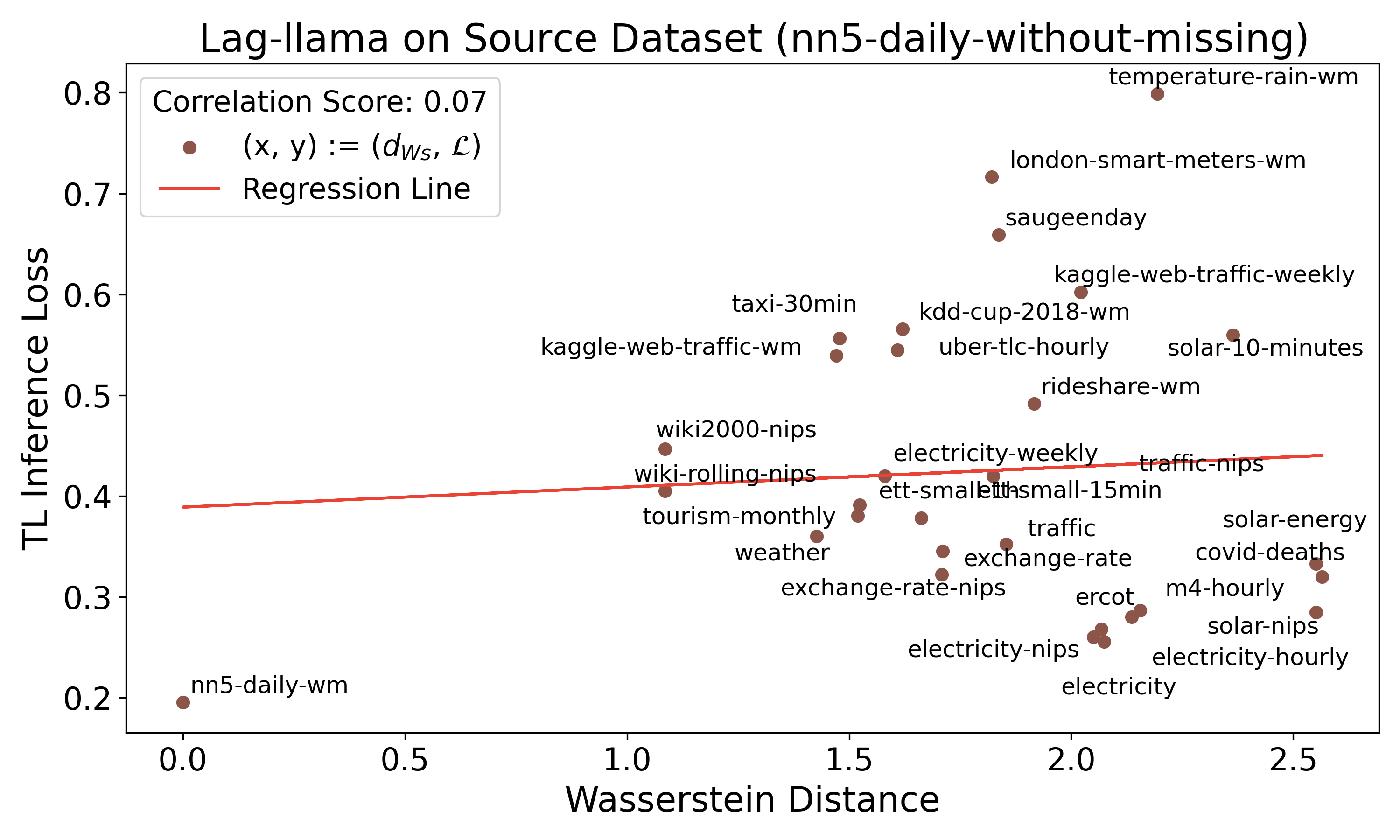}
\includegraphics[width=0.16\linewidth]{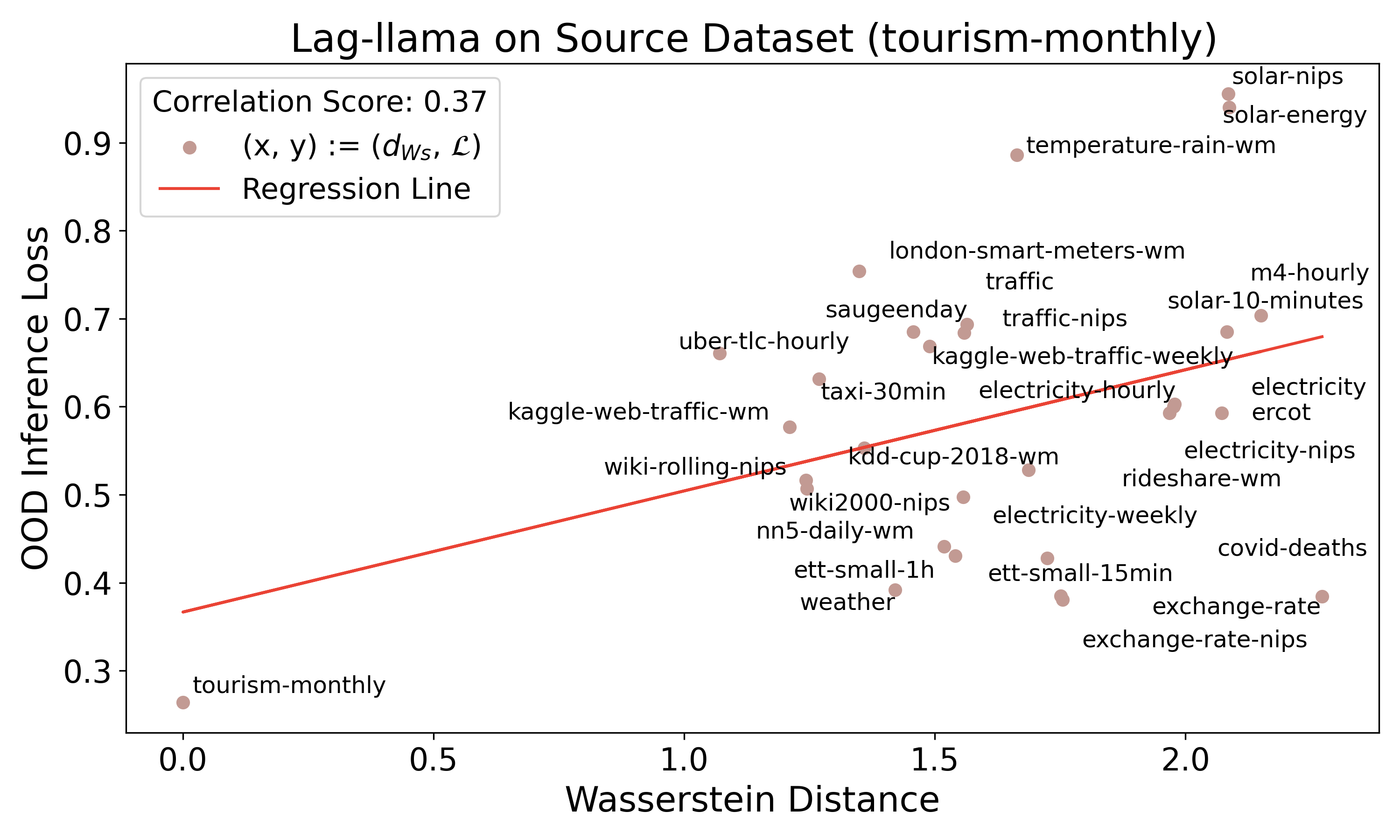}
\includegraphics[width=0.16\linewidth]{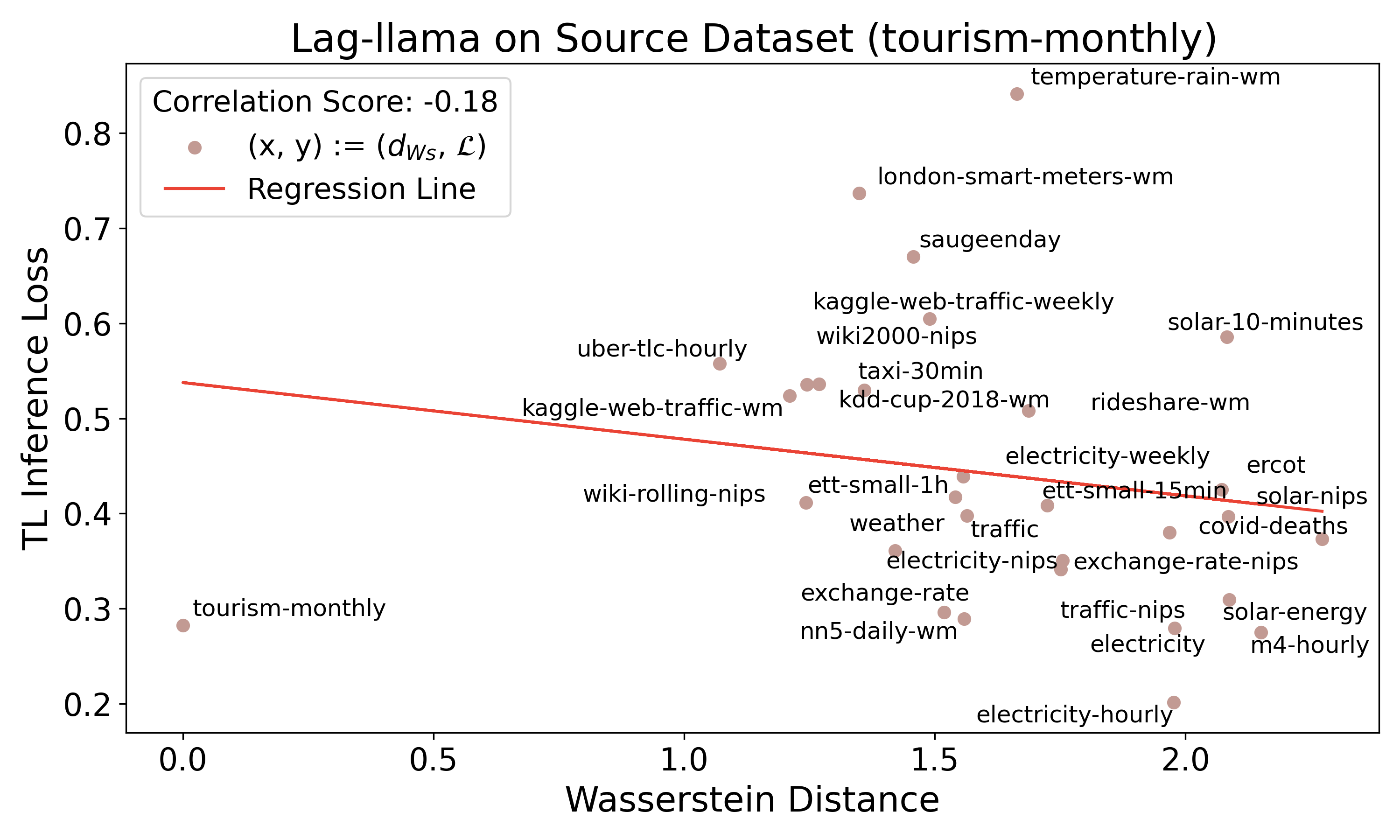}

\includegraphics[width=0.16\linewidth]{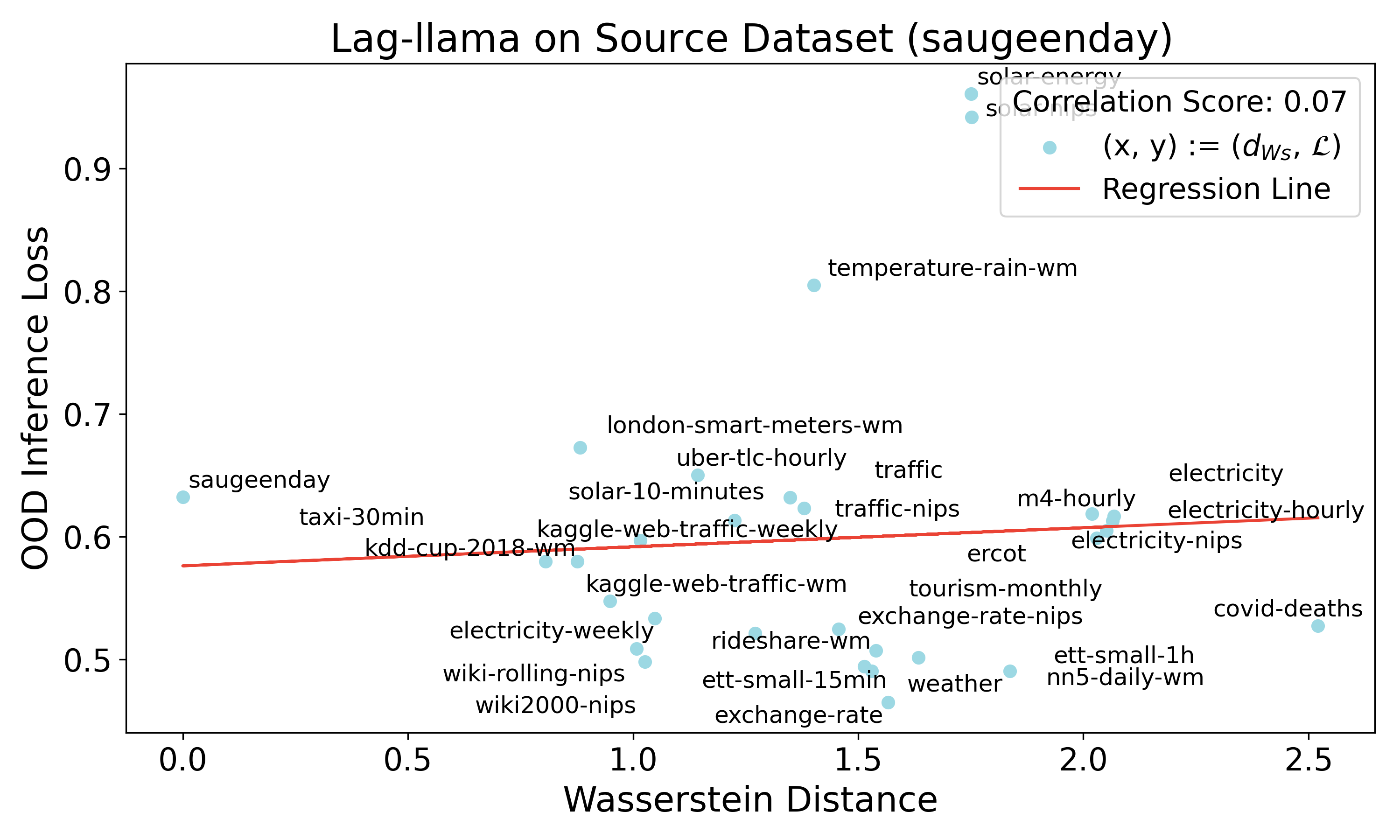}
\includegraphics[width=0.16\linewidth]{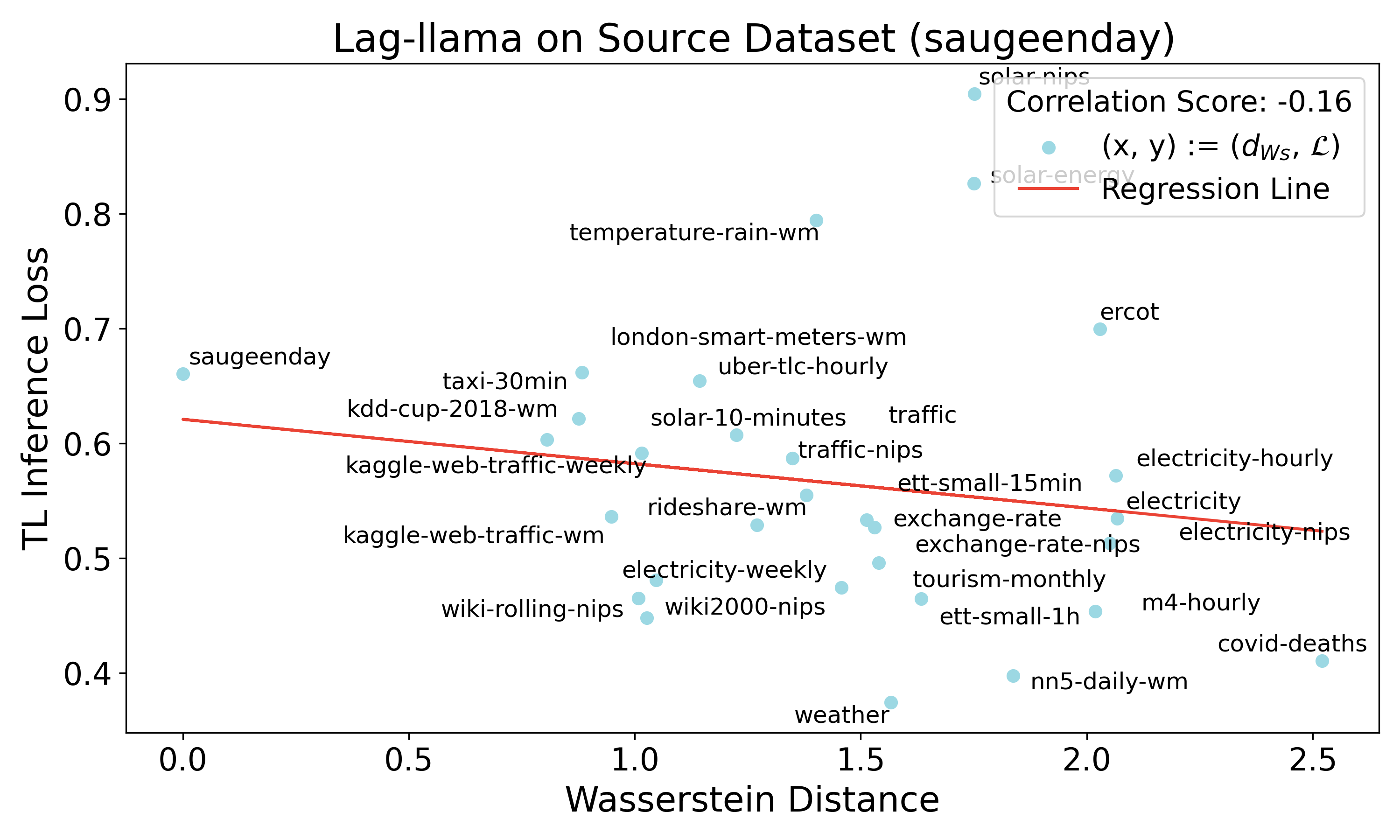}
\includegraphics[width=0.16\linewidth]{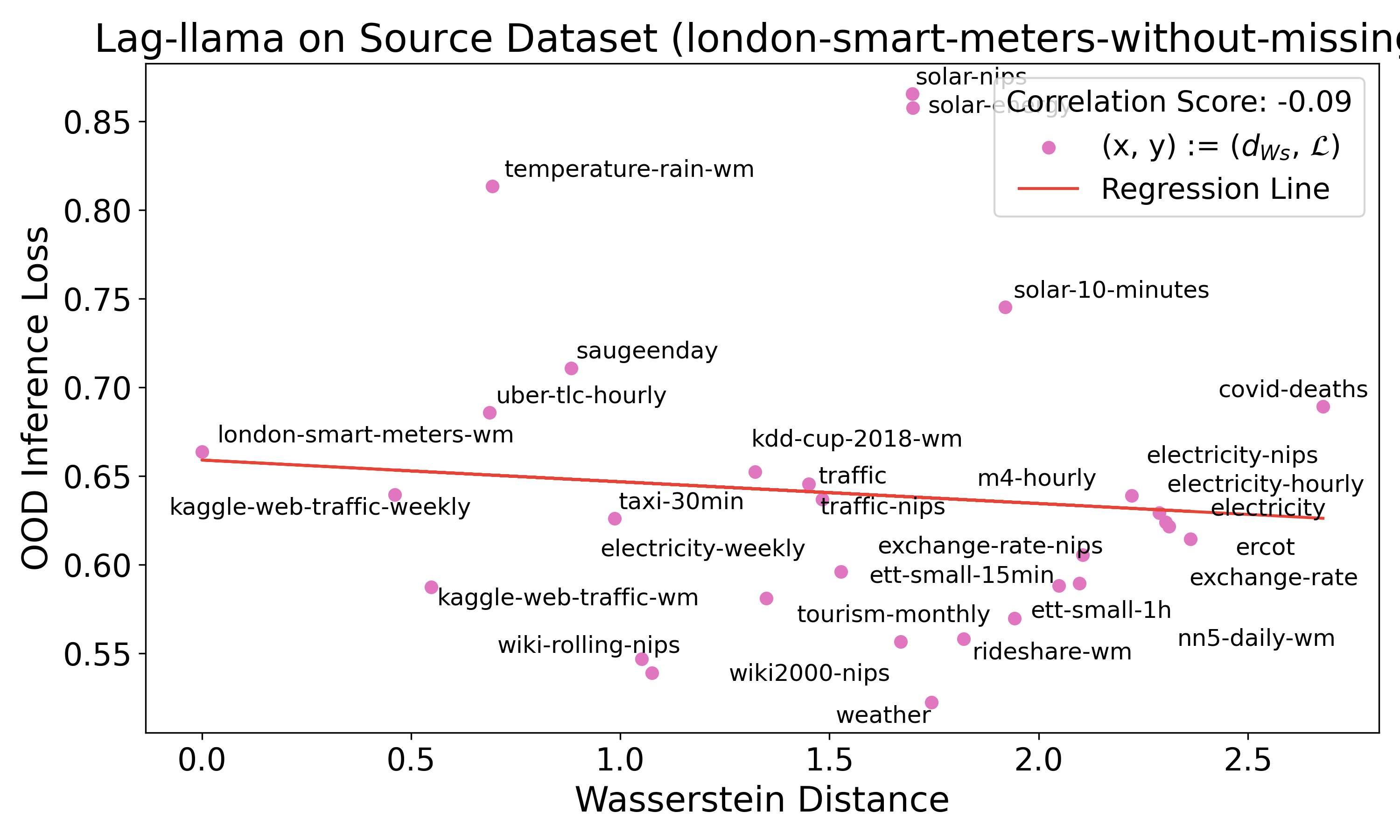}
\includegraphics[width=0.16\linewidth]{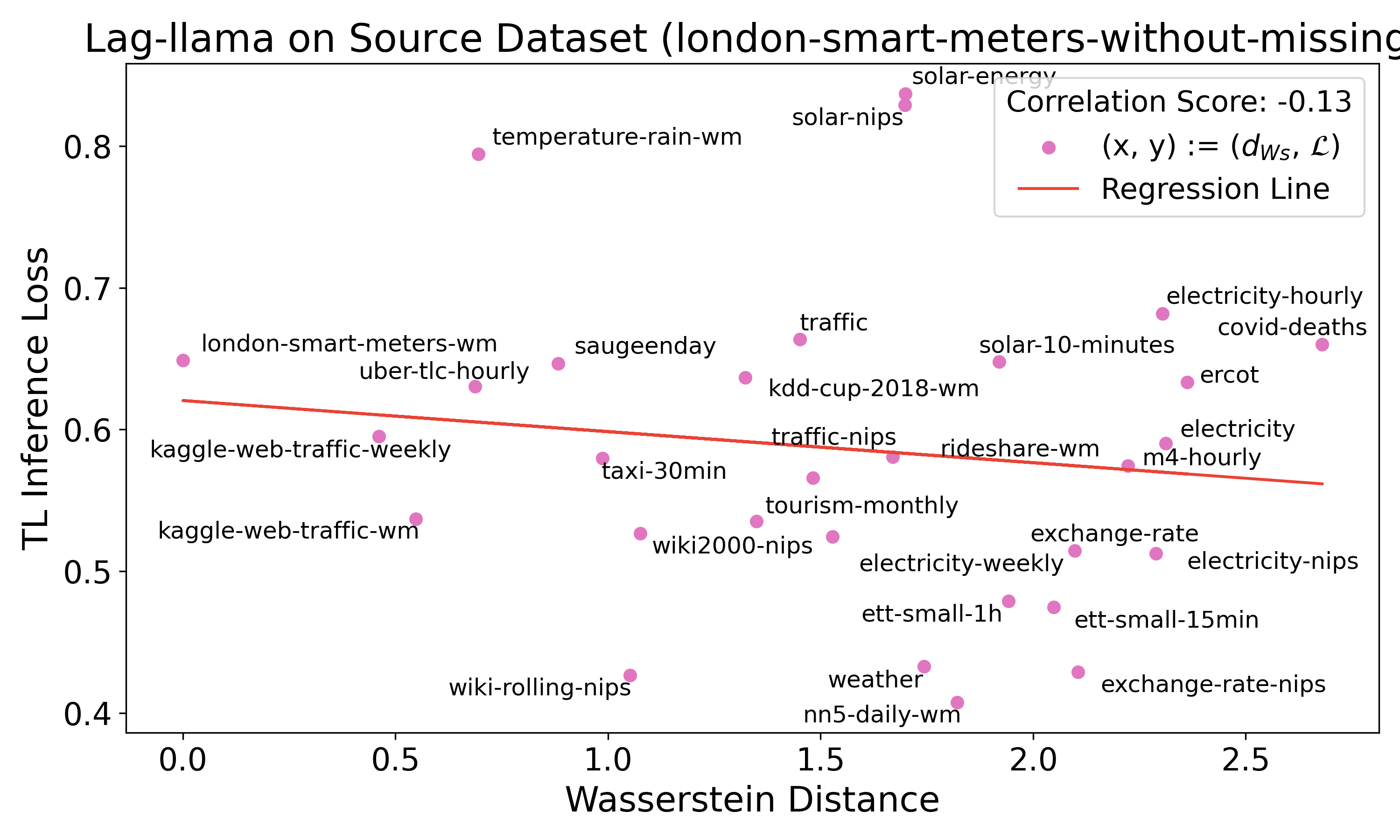}
\includegraphics[width=0.16\linewidth]{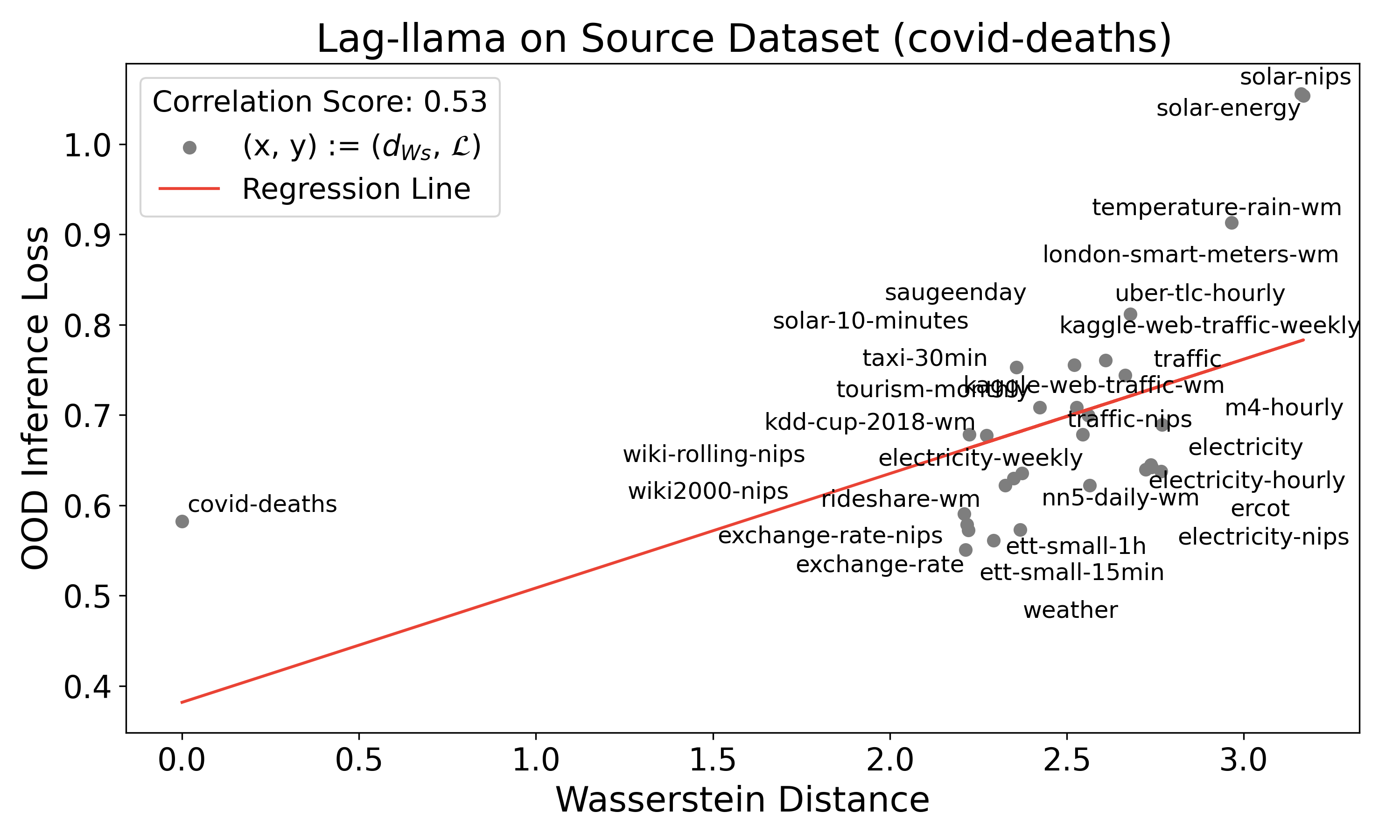}
\includegraphics[width=0.16\linewidth]{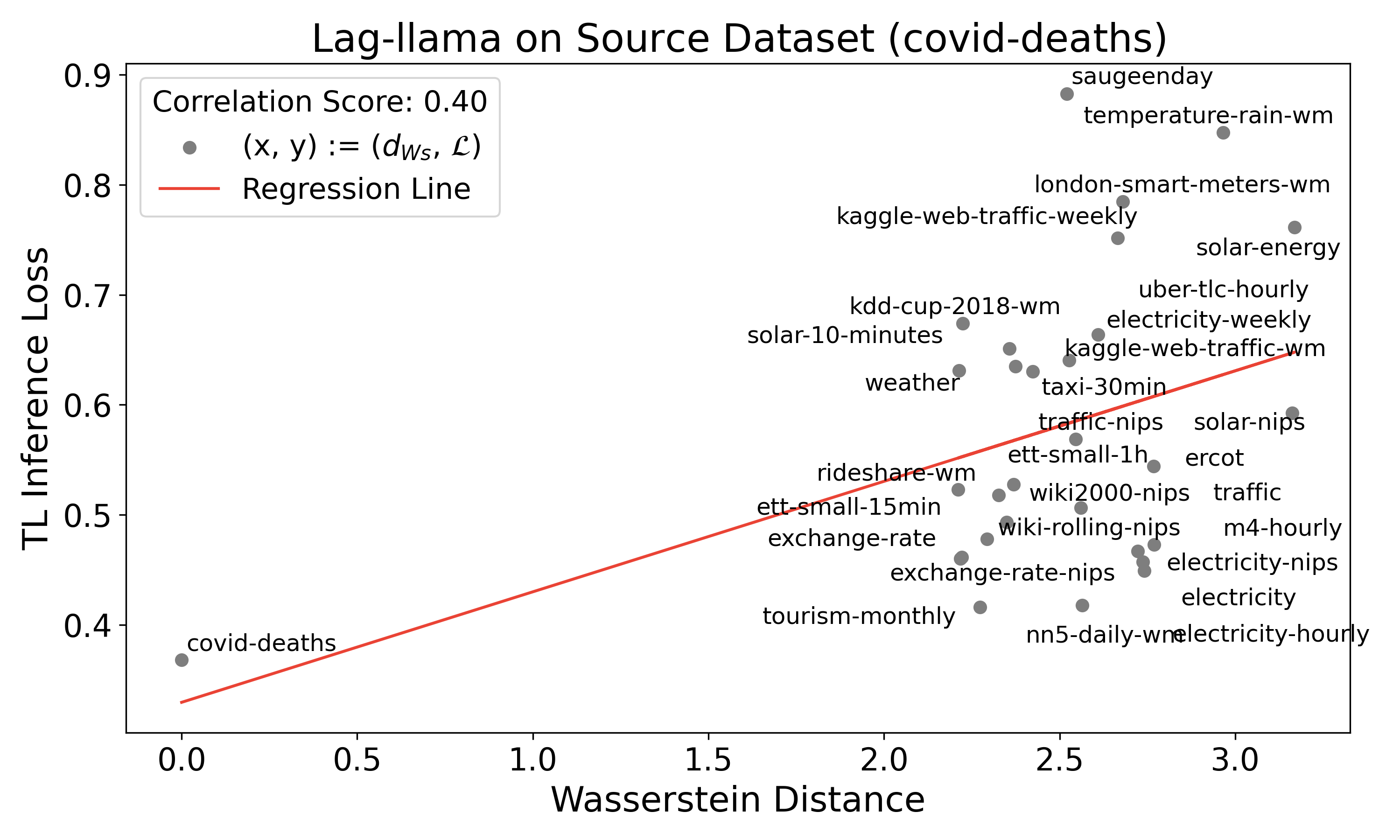}
\caption{
Out-Of-Distribution (OOD) and Transfer Learning (TL) inference losses of \textit{Lag-llama} trained on each individual dataset.
Each dot represents a target dataset, where the x-axis represents the Wasserstein distance between the source dataset and the target dataset, while the y-axis represents the OOD or TL loss on the target dataset.
}
\label{fig:complete-lagllama-OOD-TL-pairs}

\end{figure*}
\begin{figure*}[t]
\centering
\includegraphics[width=0.16\linewidth]{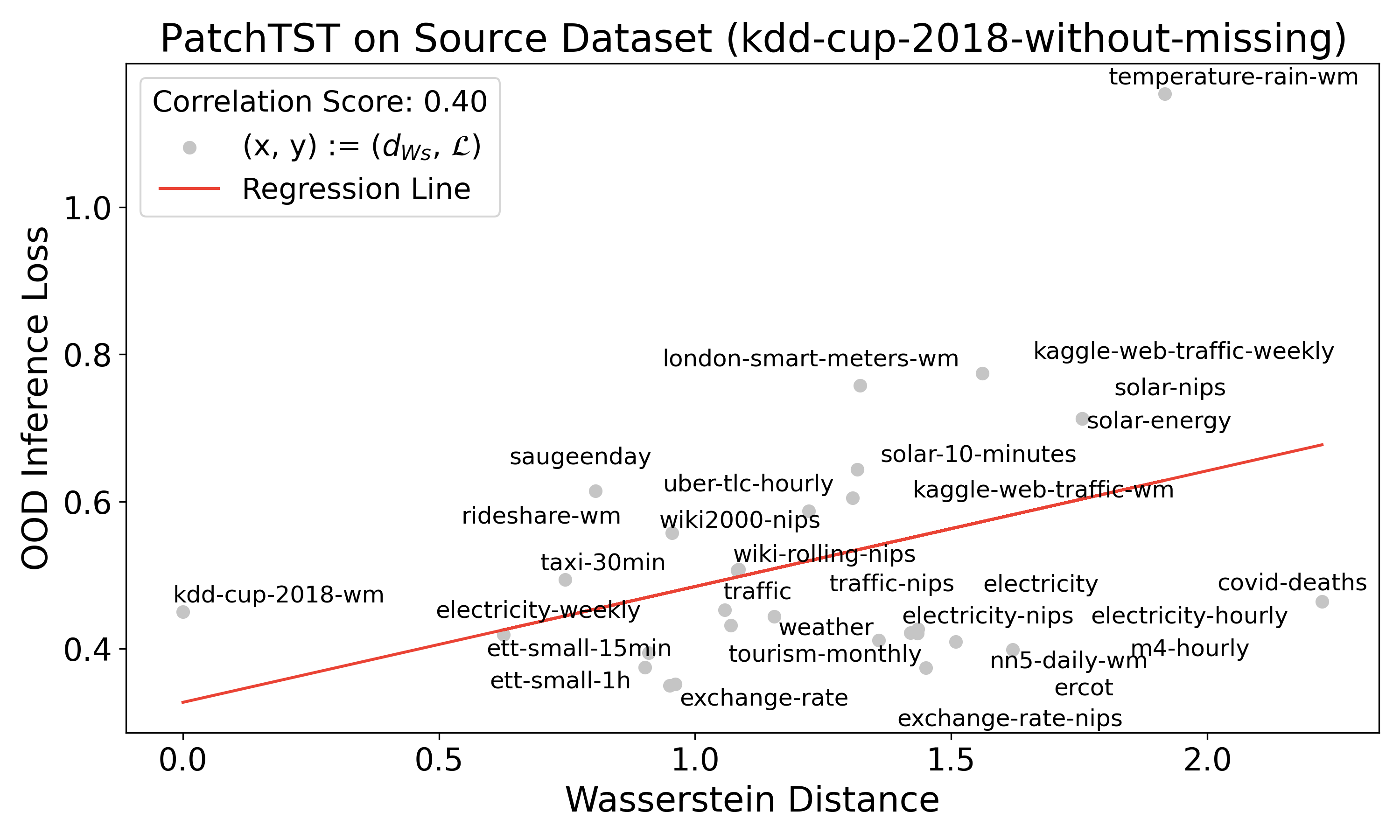}
\includegraphics[width=0.16\linewidth]{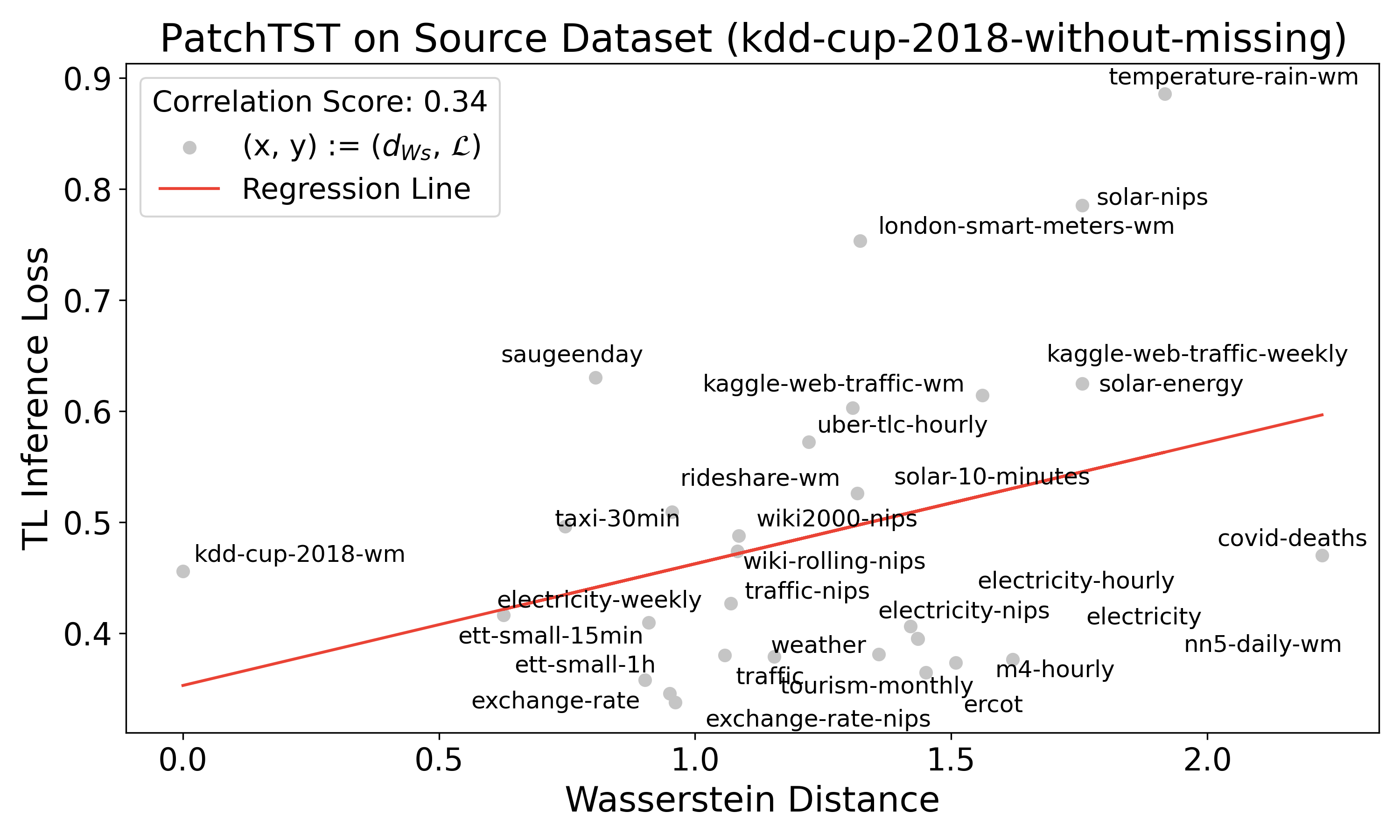}
\includegraphics[width=0.16\linewidth]{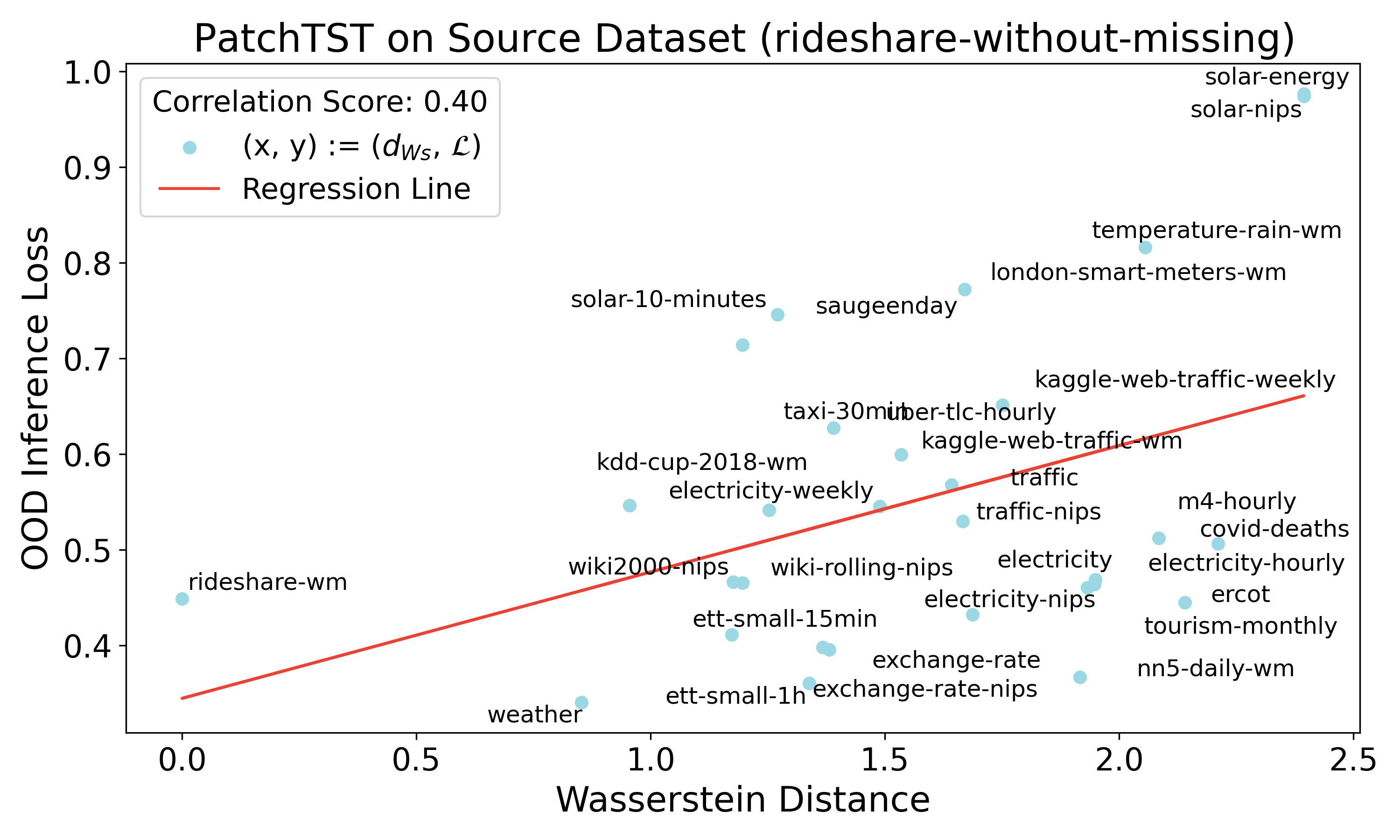}
\includegraphics[width=0.16\linewidth]{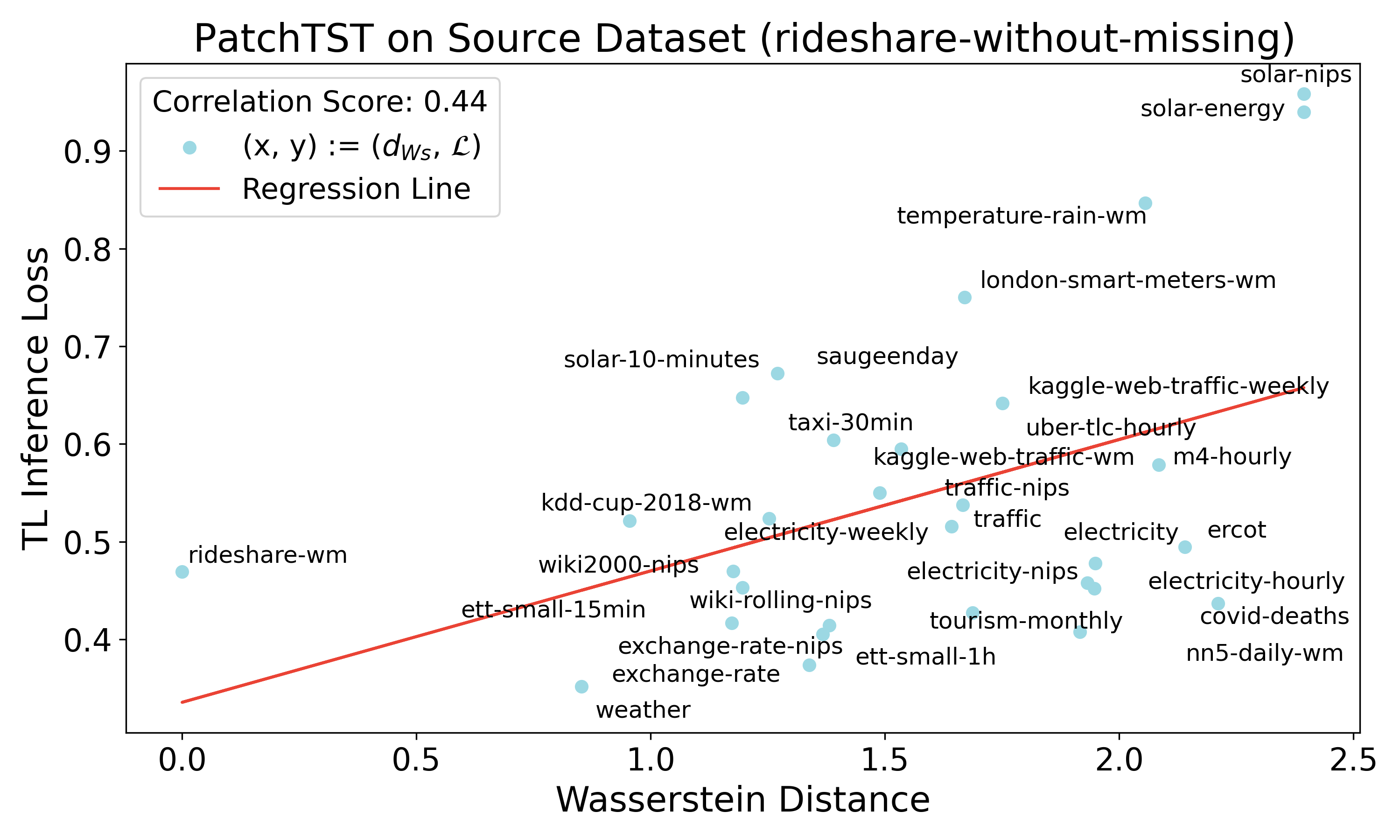}
\includegraphics[width=0.16\linewidth]{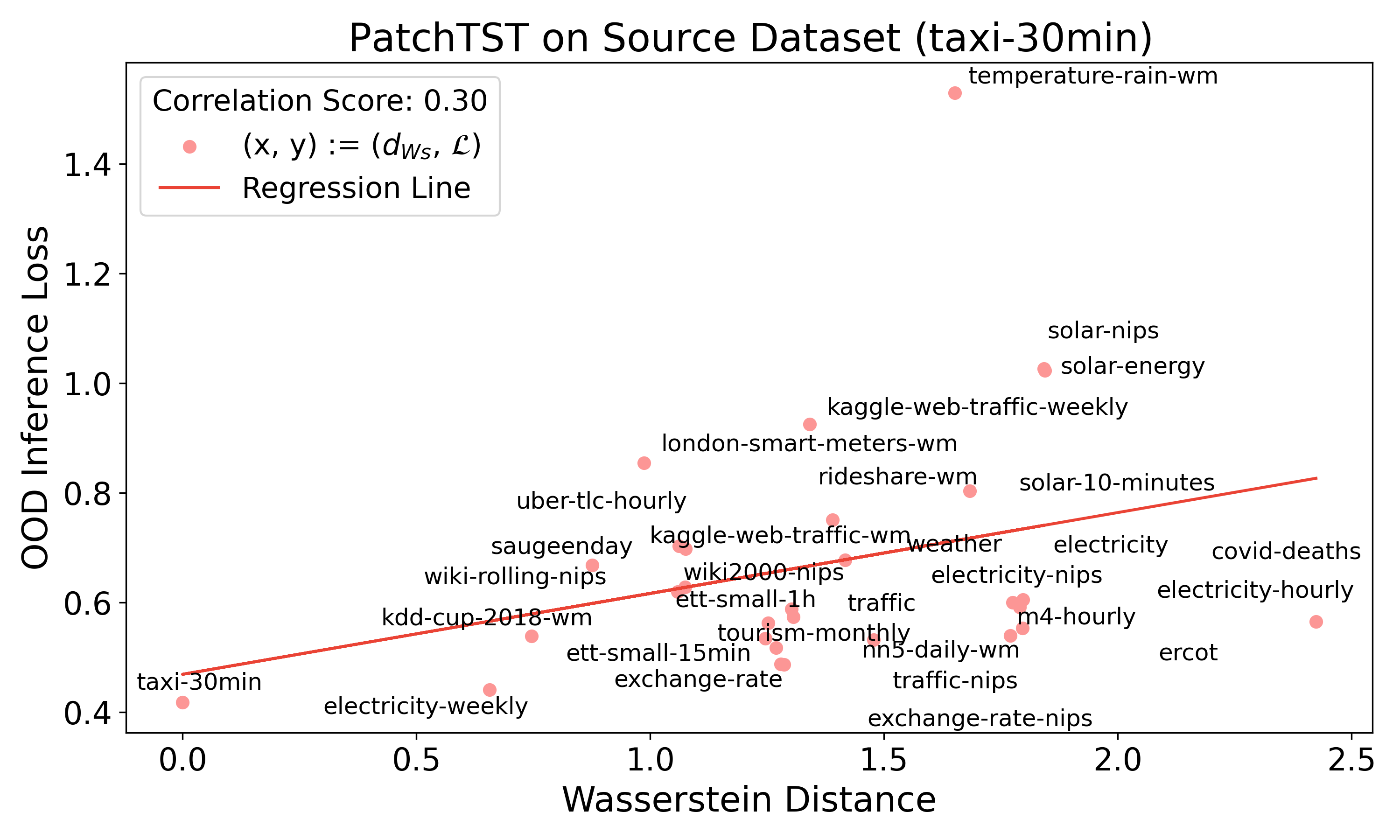}
\includegraphics[width=0.16\linewidth]{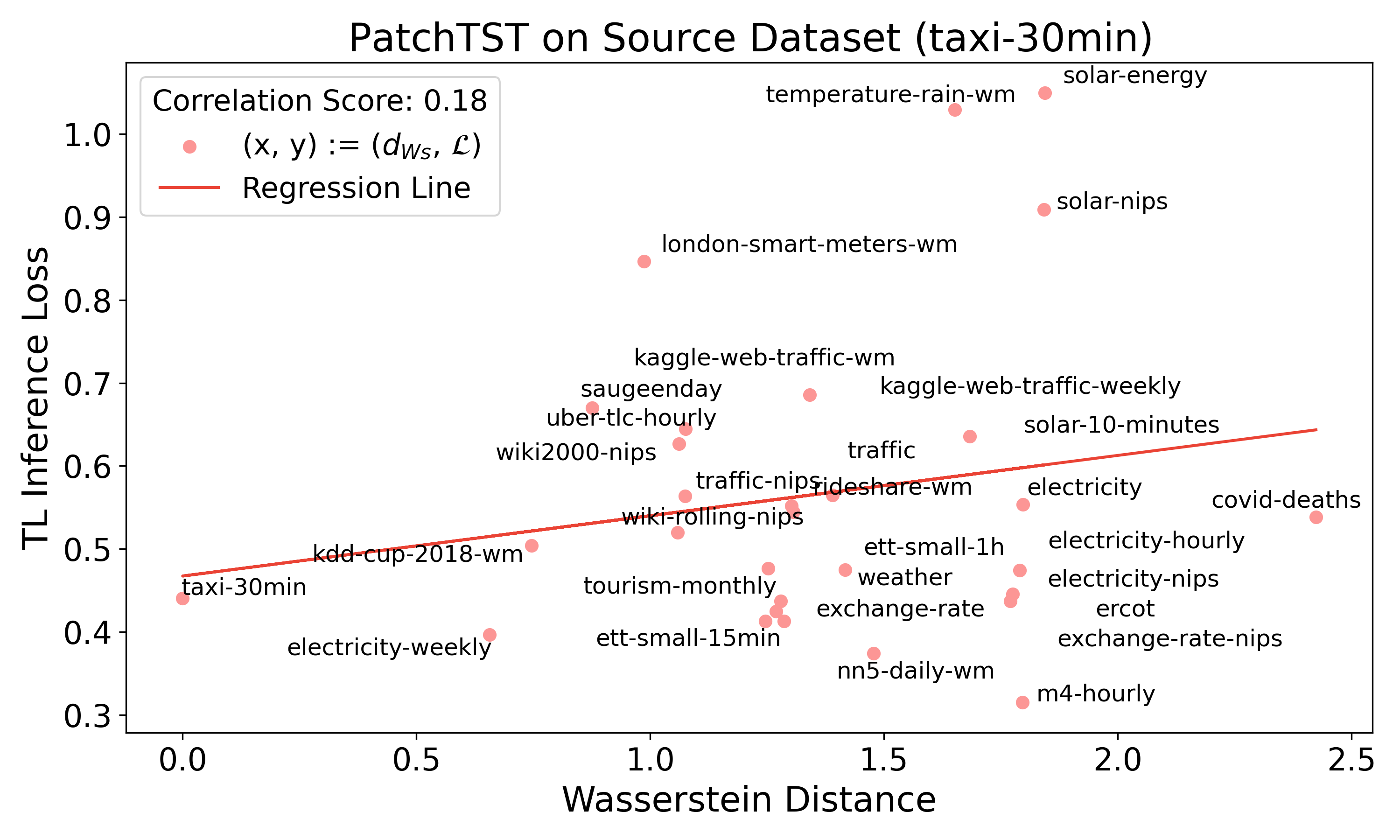}

\includegraphics[width=0.16\linewidth]{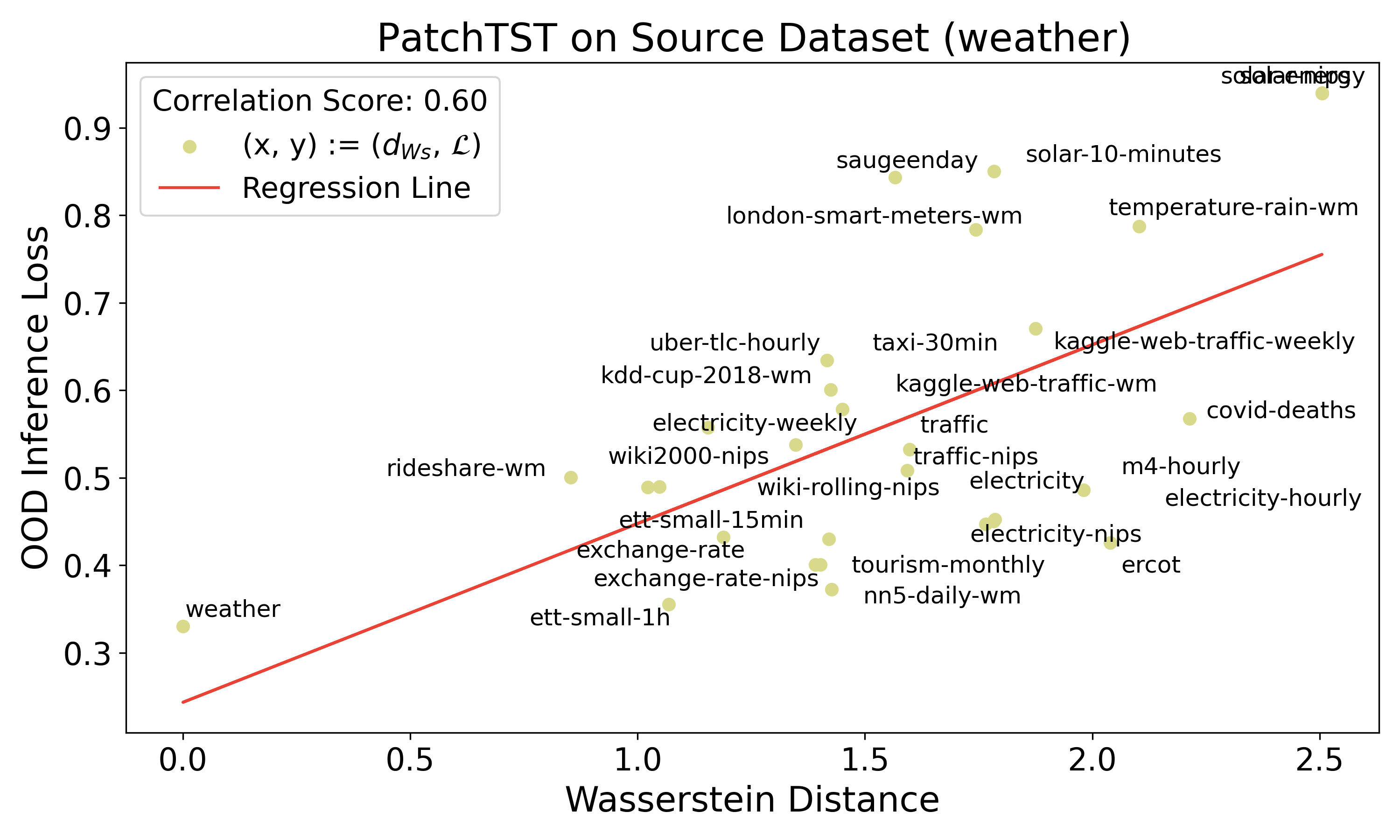}
\includegraphics[width=0.16\linewidth]{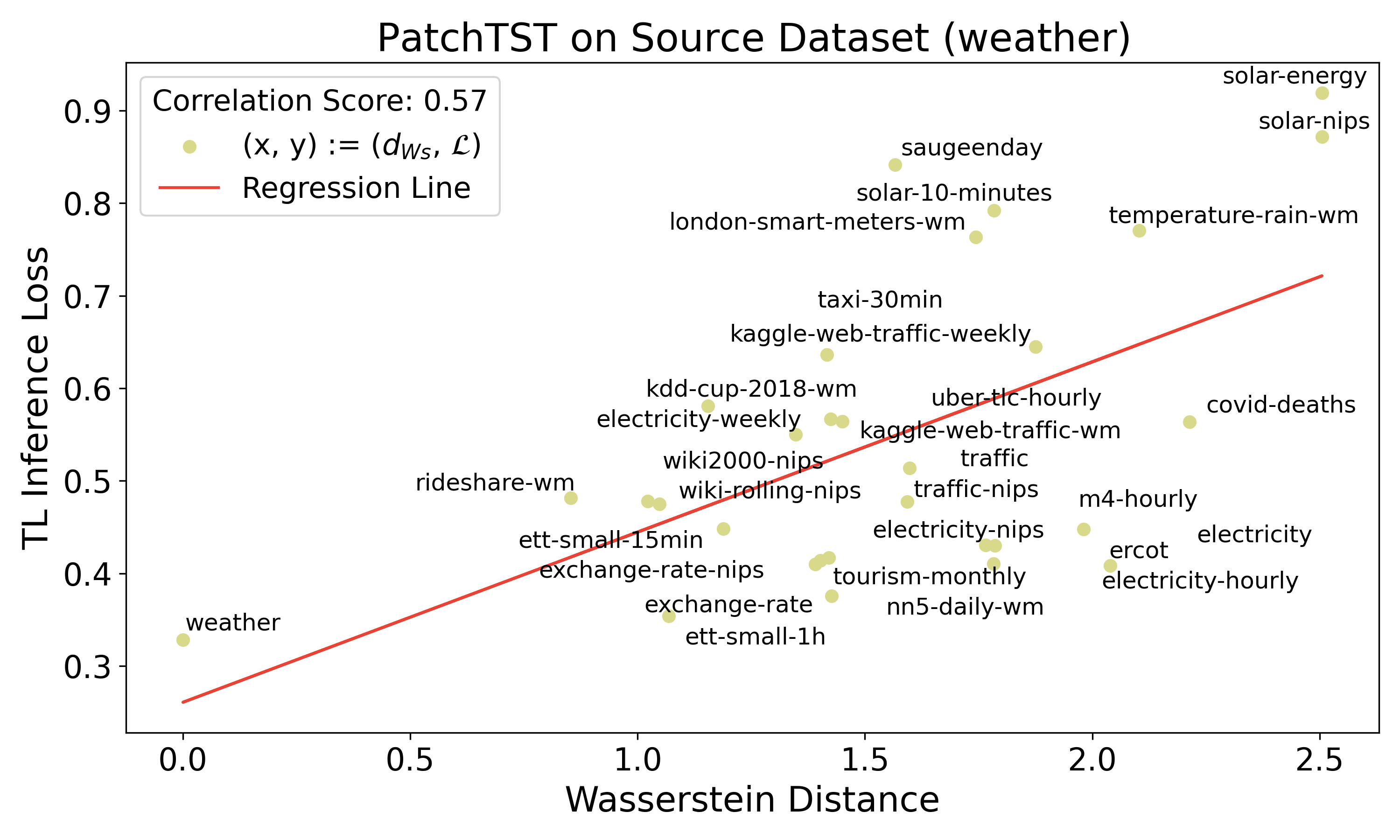}
\includegraphics[width=0.16\linewidth]{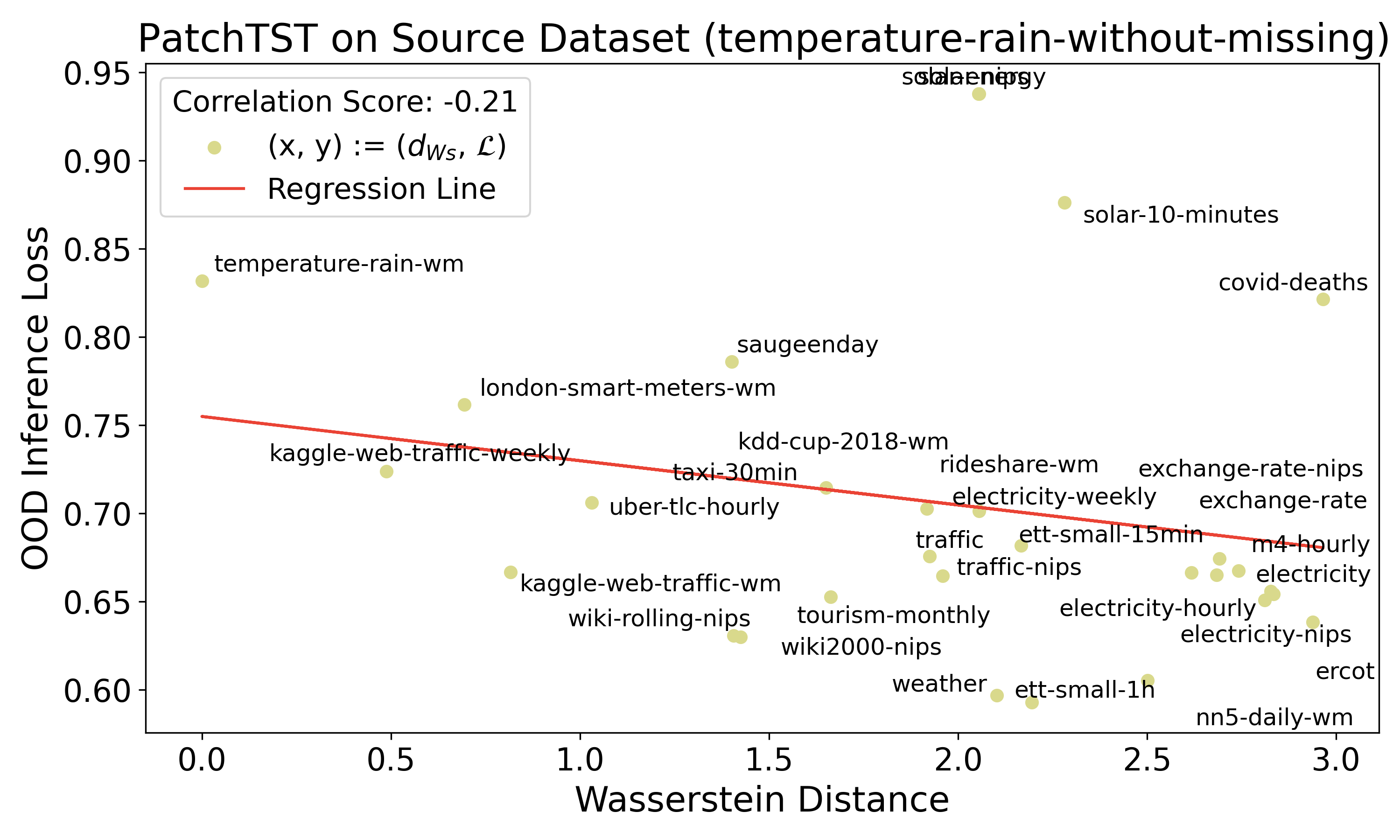}
\includegraphics[width=0.16\linewidth]{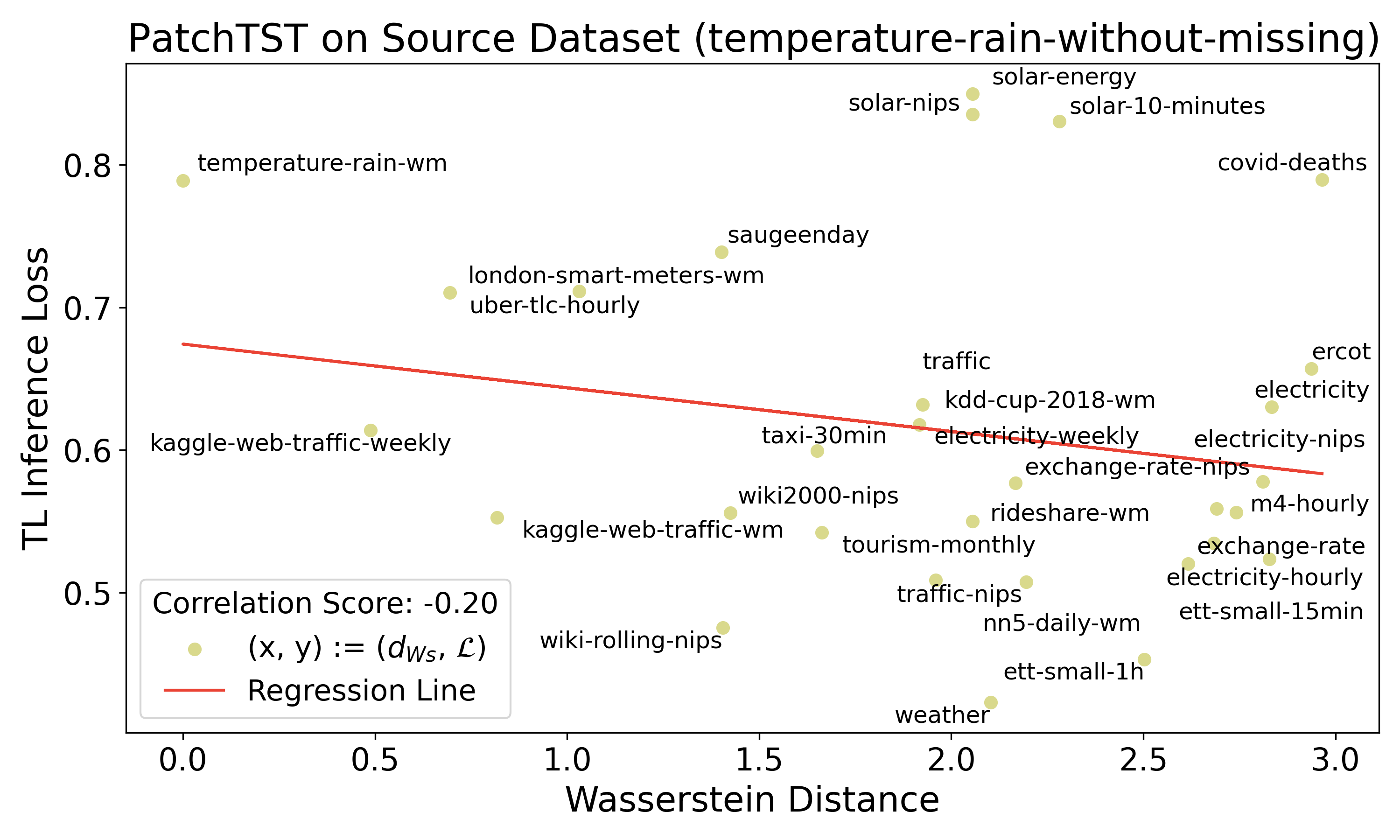}
\includegraphics[width=0.16\linewidth]{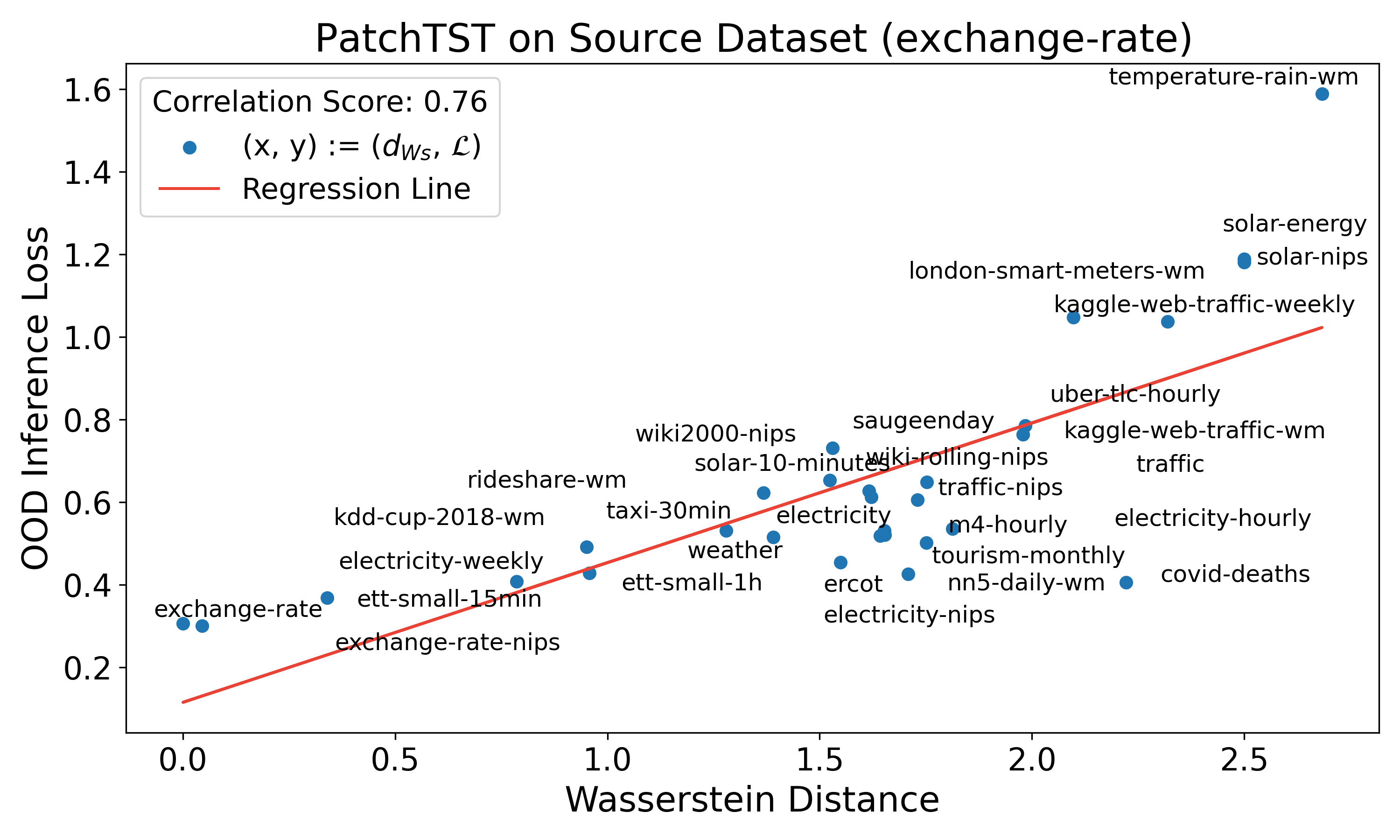}
\includegraphics[width=0.16\linewidth]{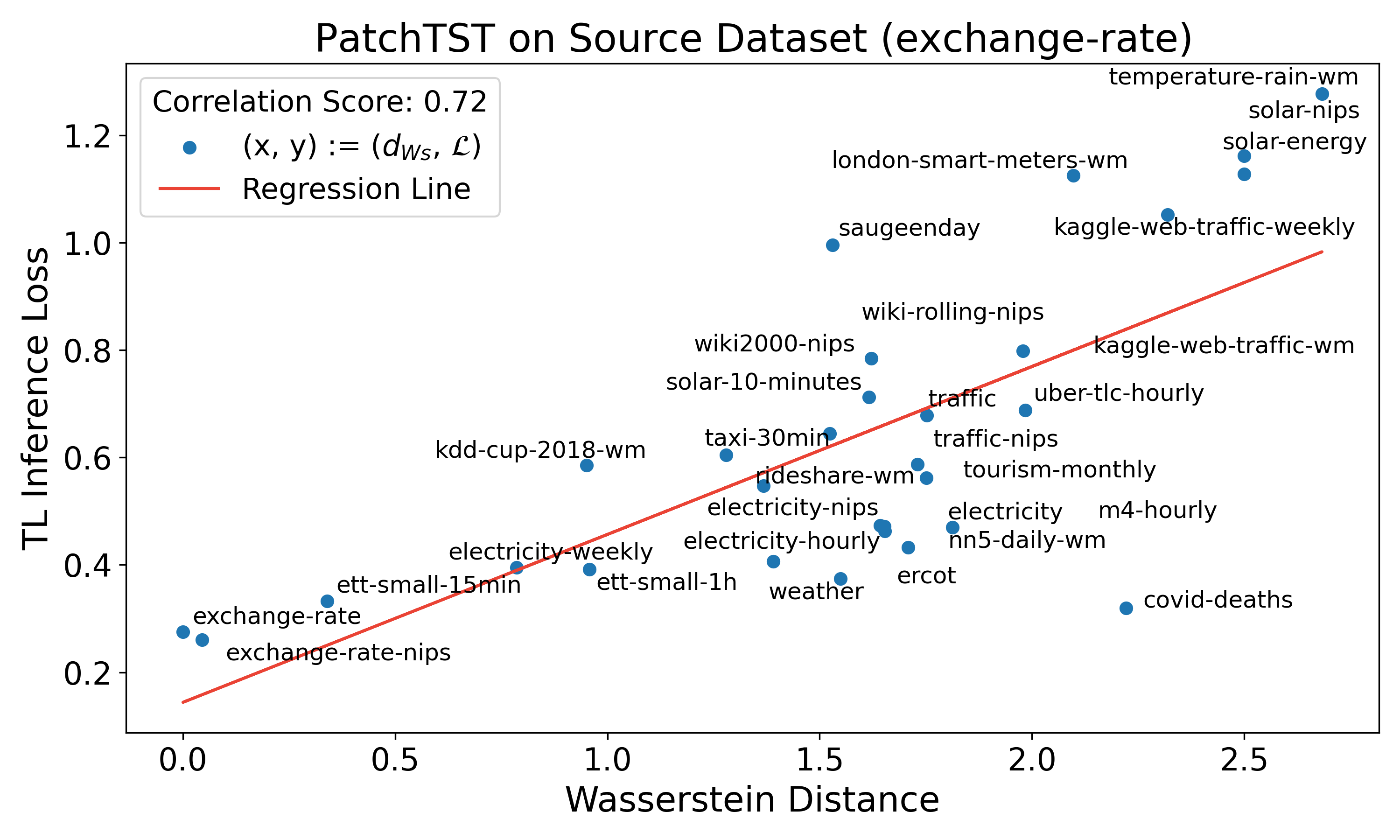}

\includegraphics[width=0.16\linewidth]{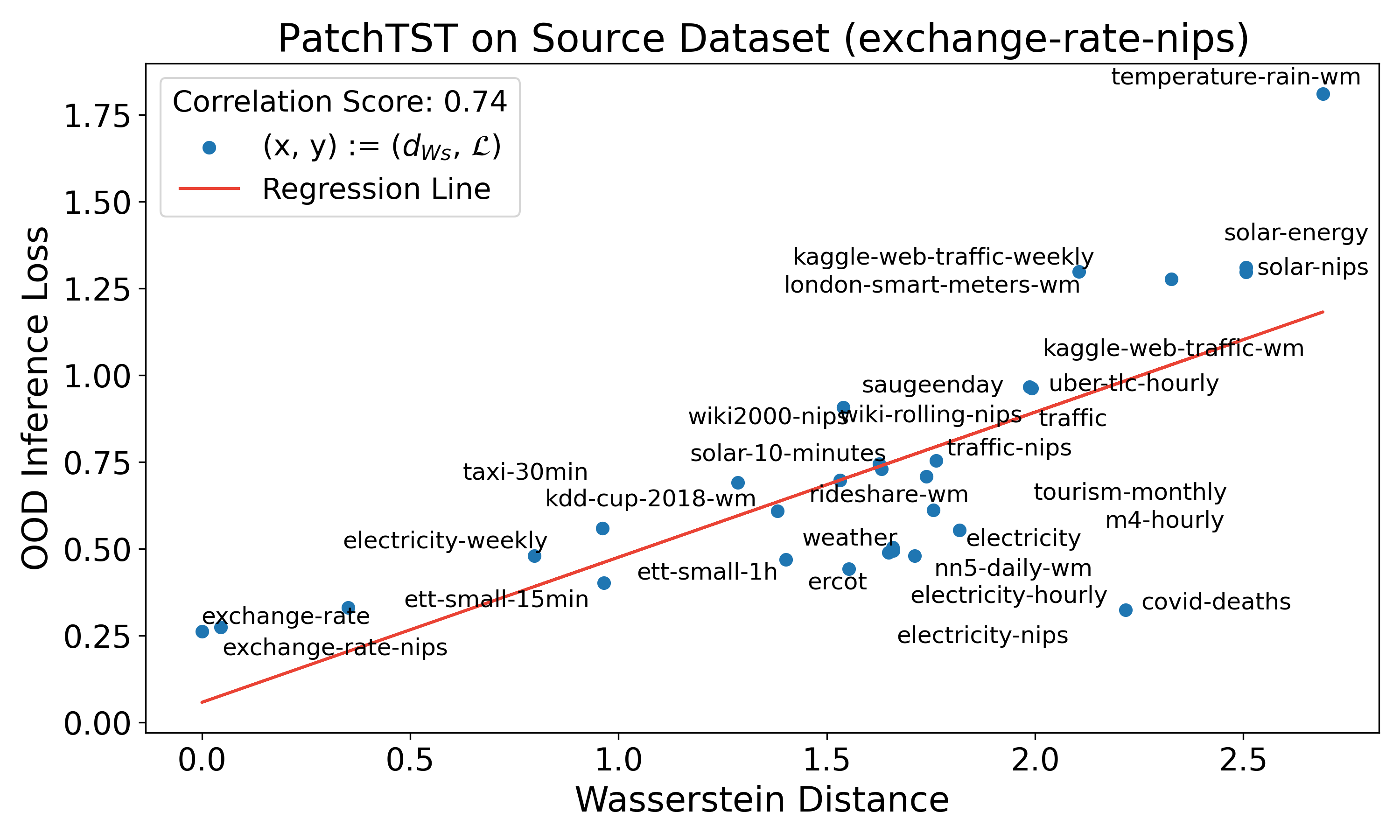}
\includegraphics[width=0.16\linewidth]{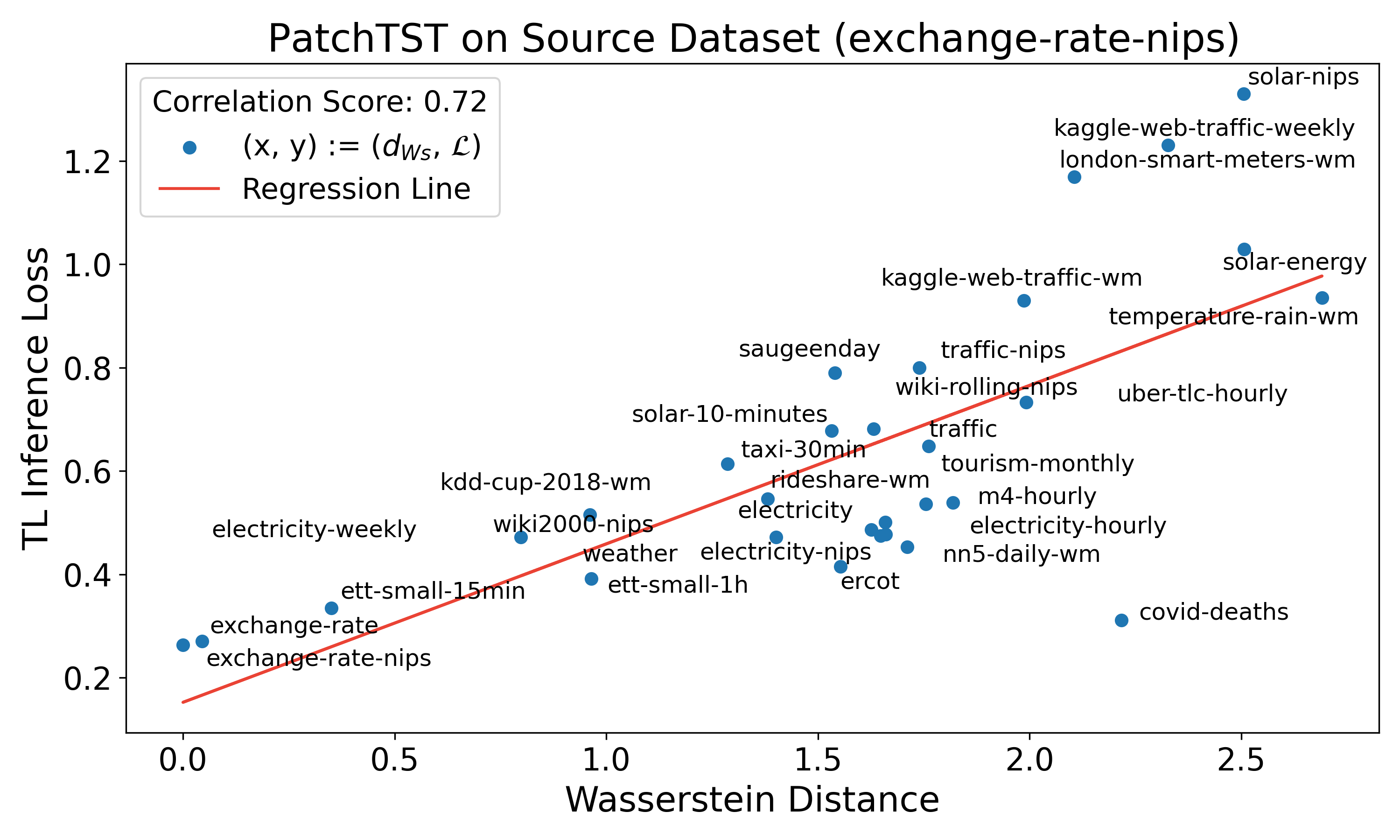}
\includegraphics[width=0.16\linewidth]{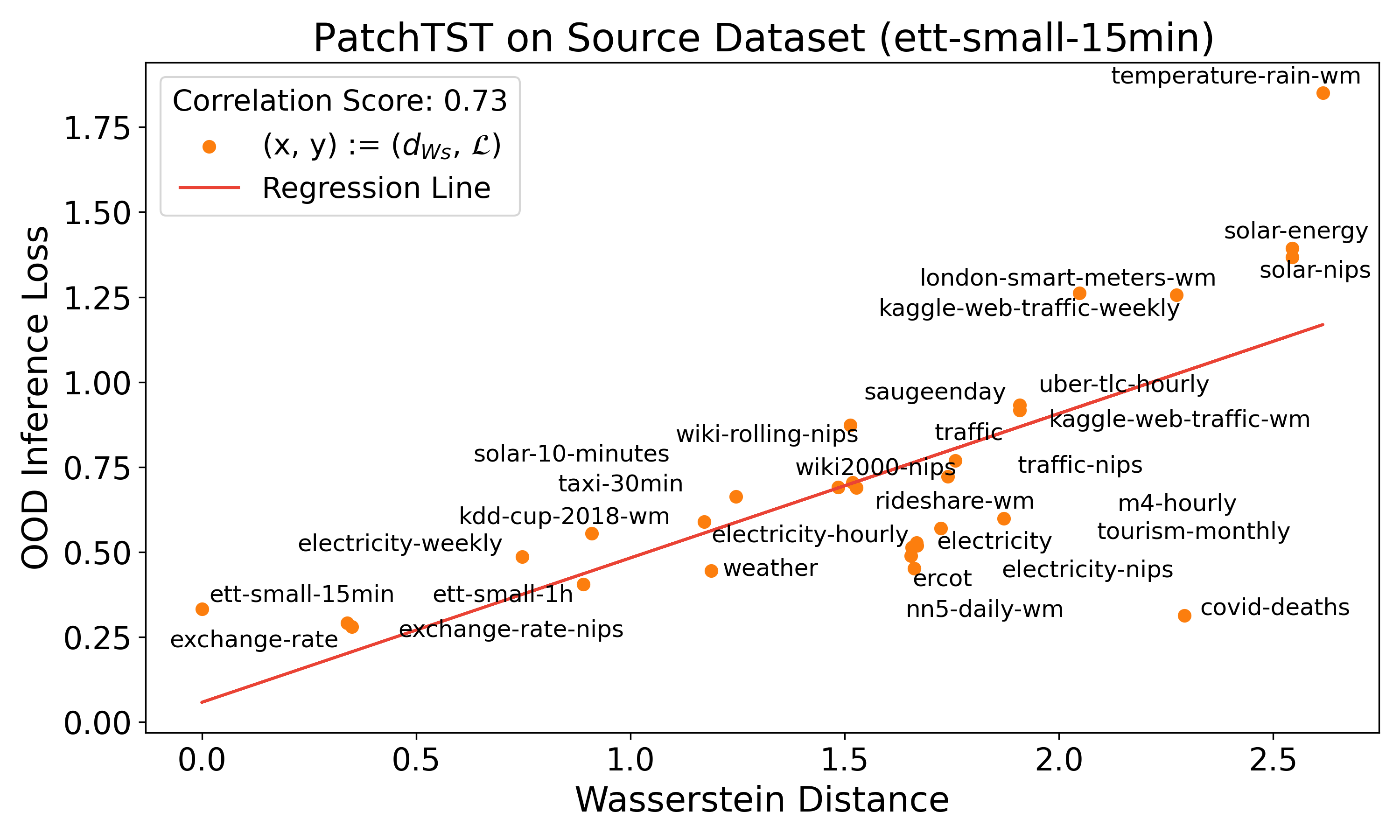}
\includegraphics[width=0.16\linewidth]{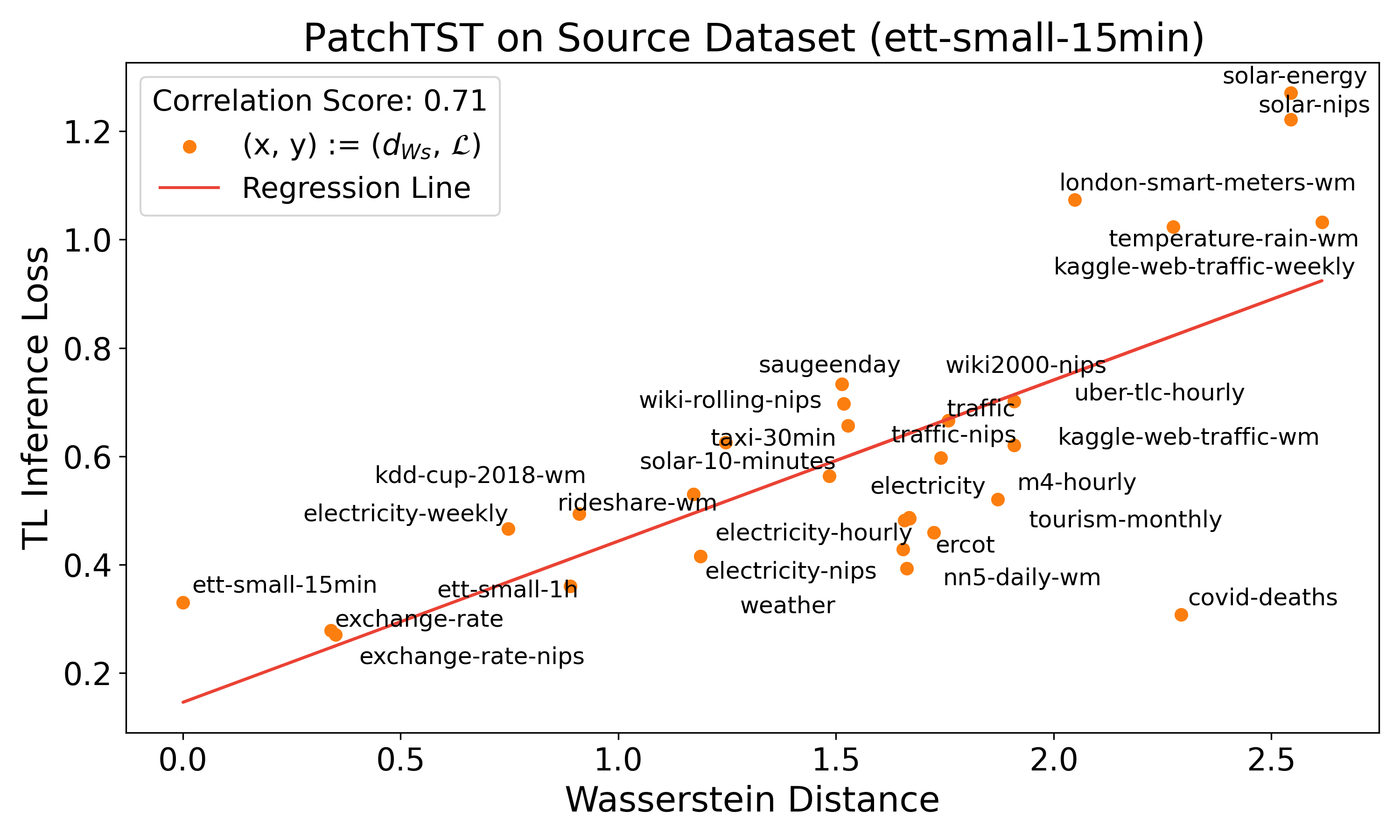}
\includegraphics[width=0.16\linewidth]{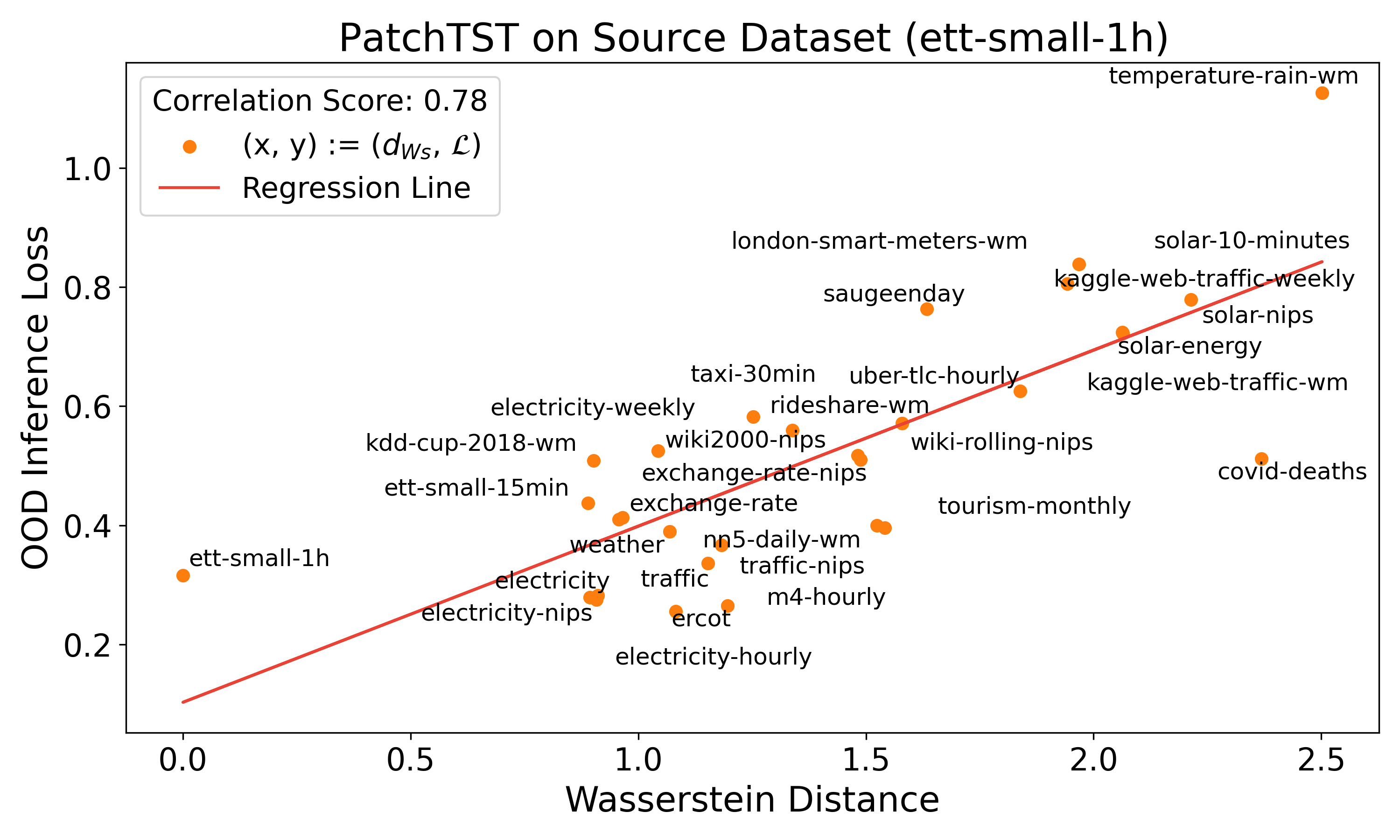}
\includegraphics[width=0.16\linewidth]{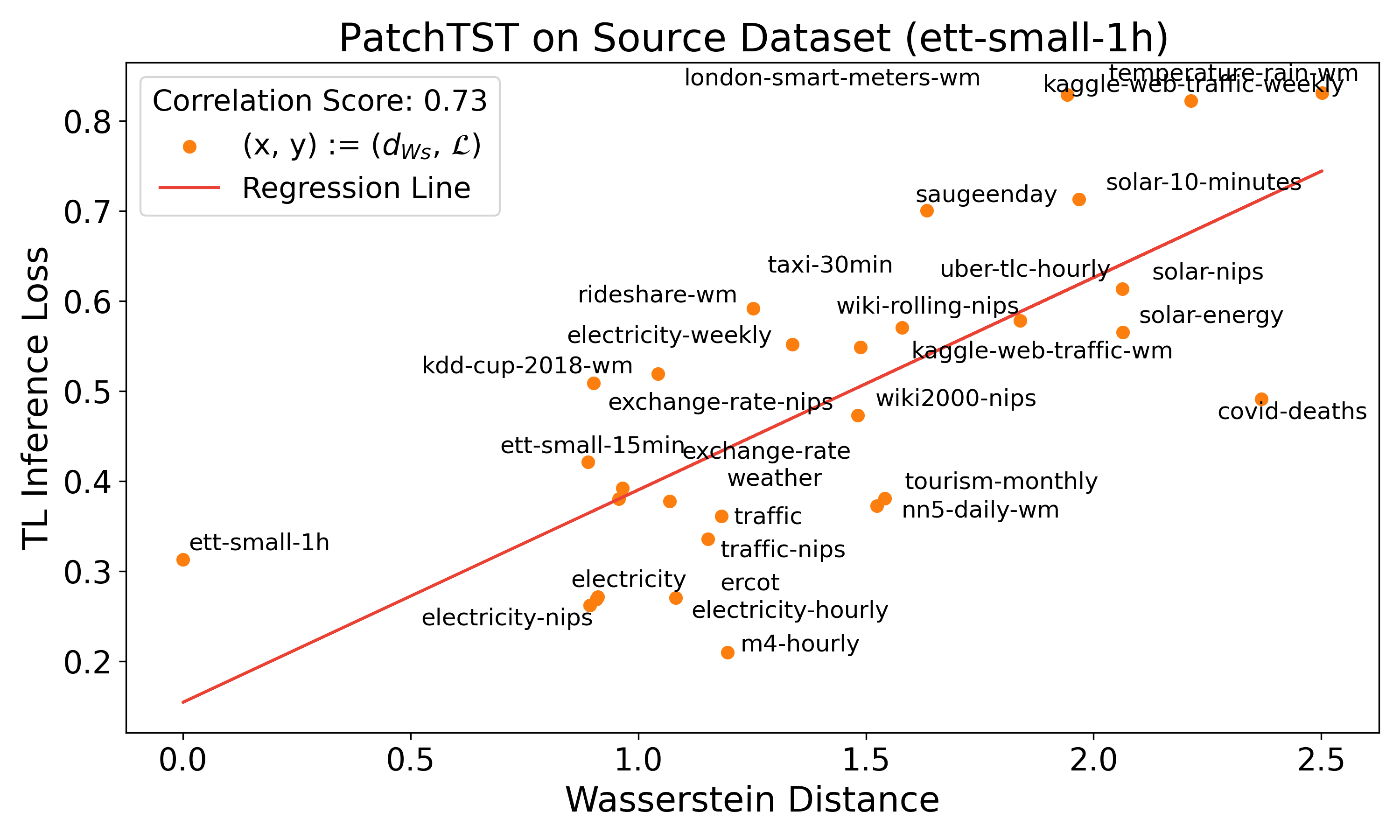}

\includegraphics[width=0.16\linewidth]{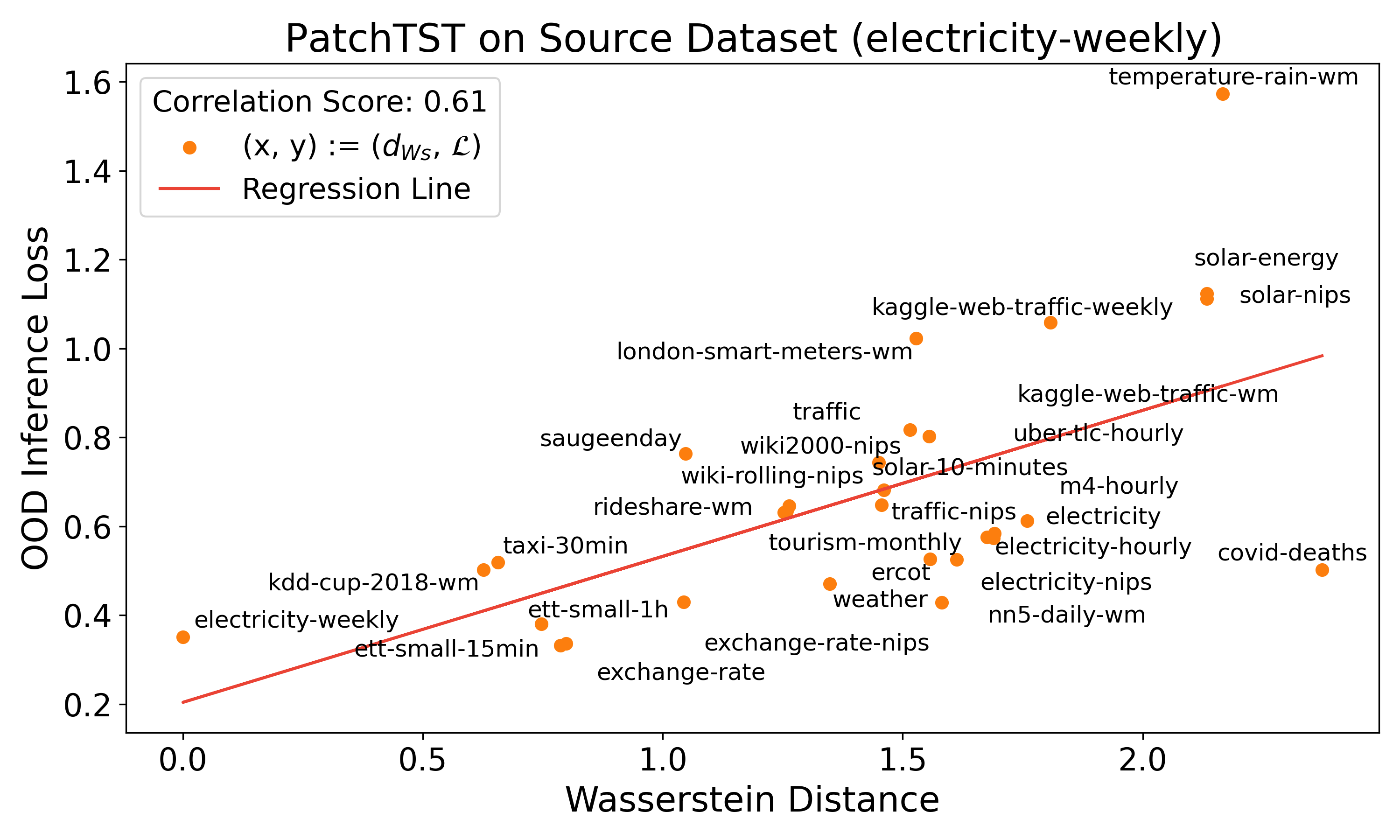}
\includegraphics[width=0.16\linewidth]{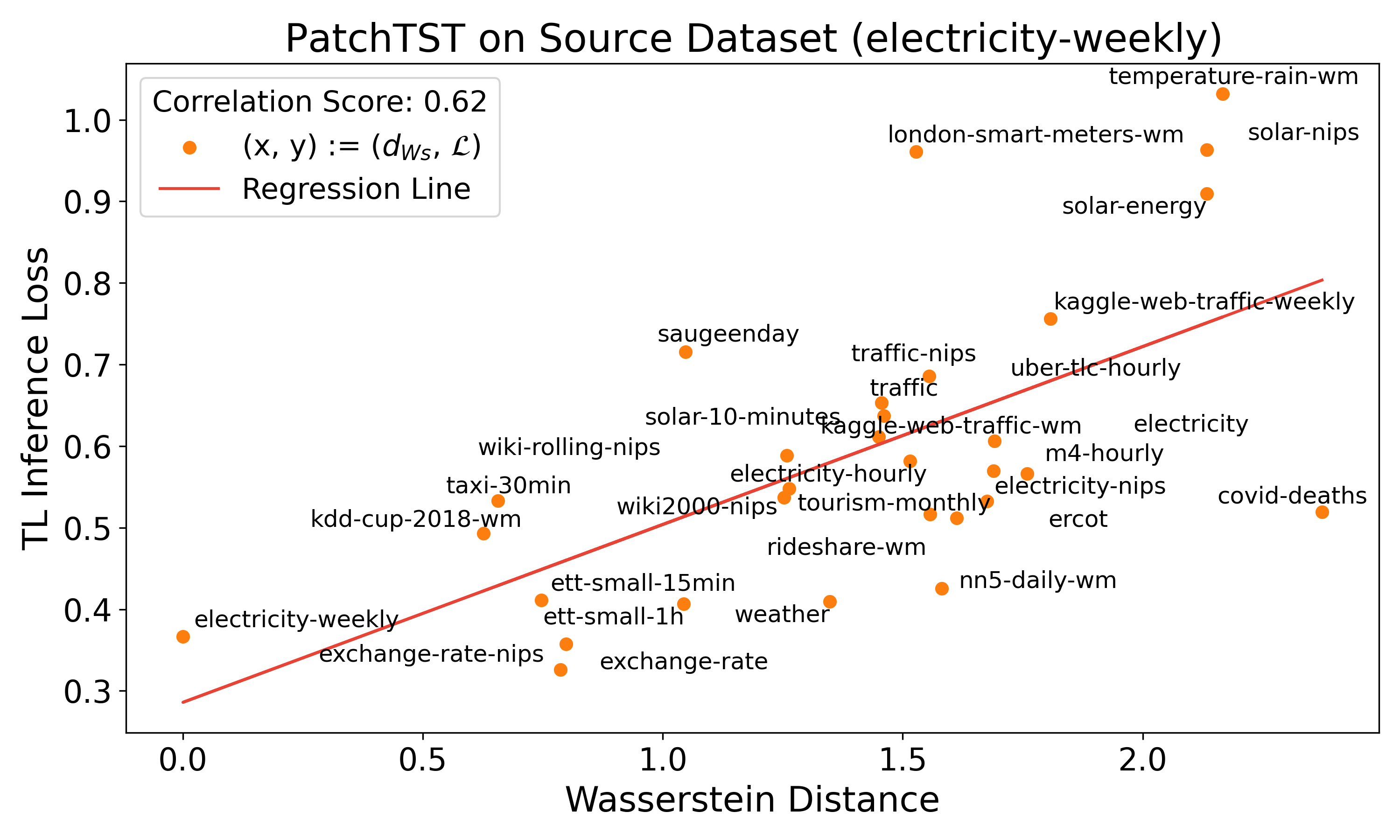}
\includegraphics[width=0.16\linewidth]{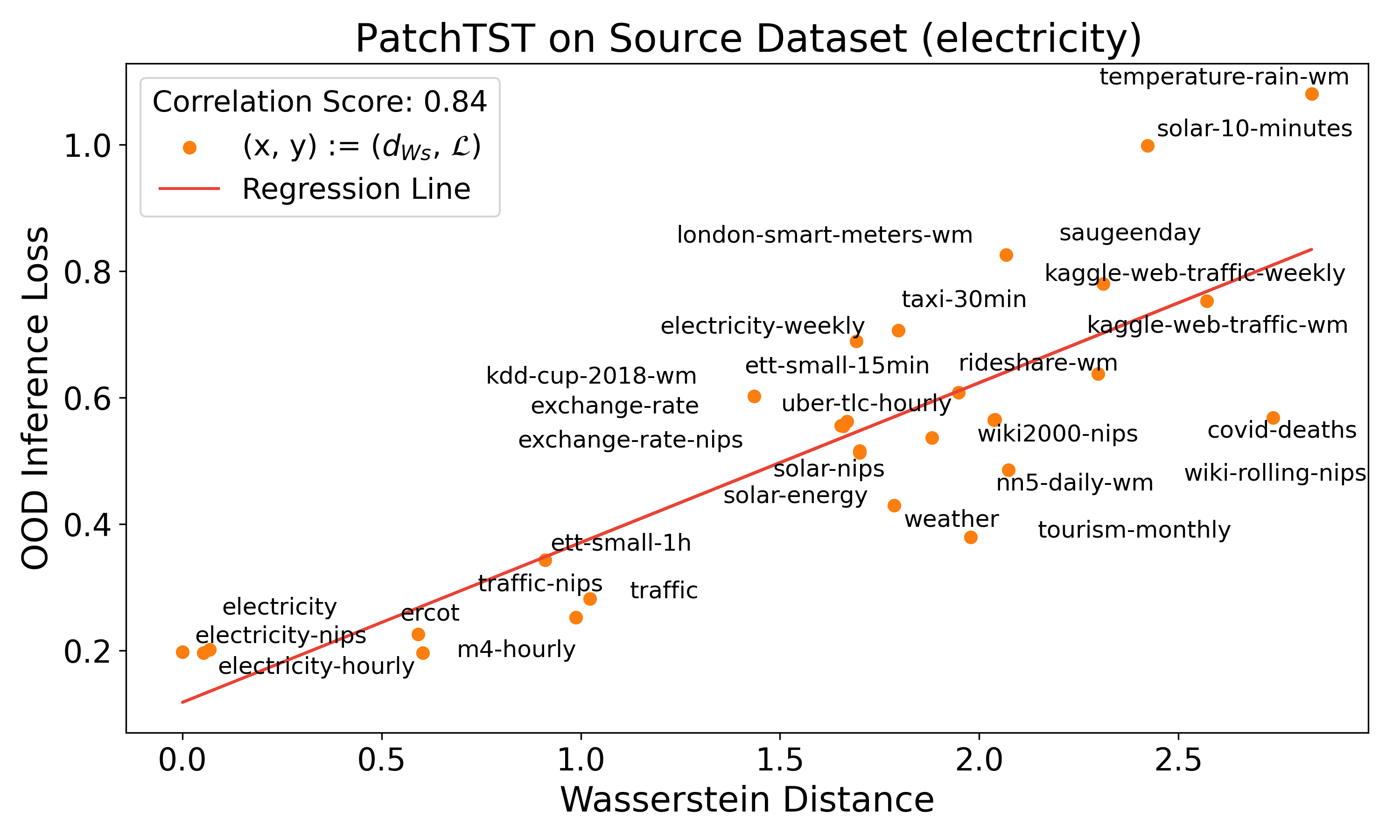}
\includegraphics[width=0.16\linewidth]{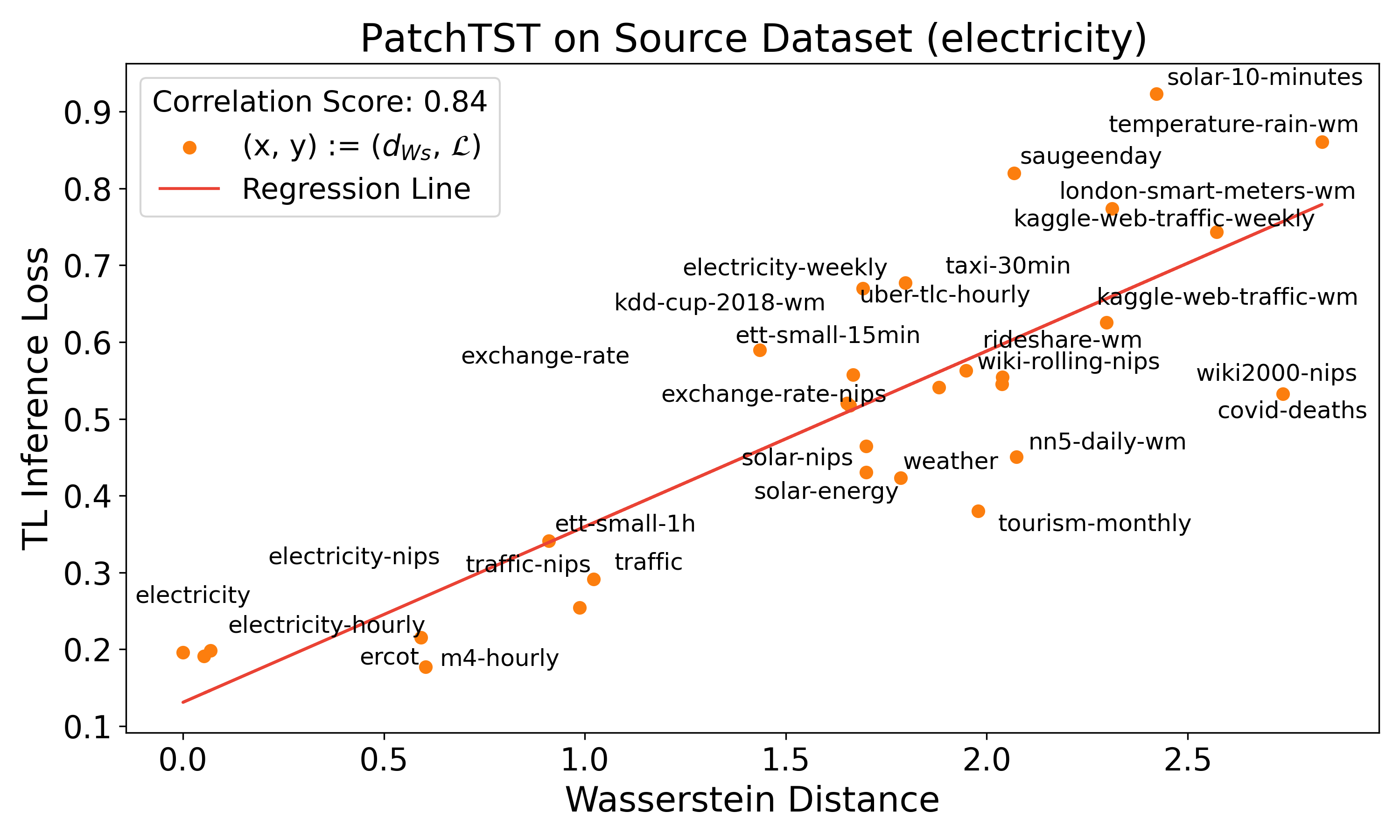}
\includegraphics[width=0.16\linewidth]{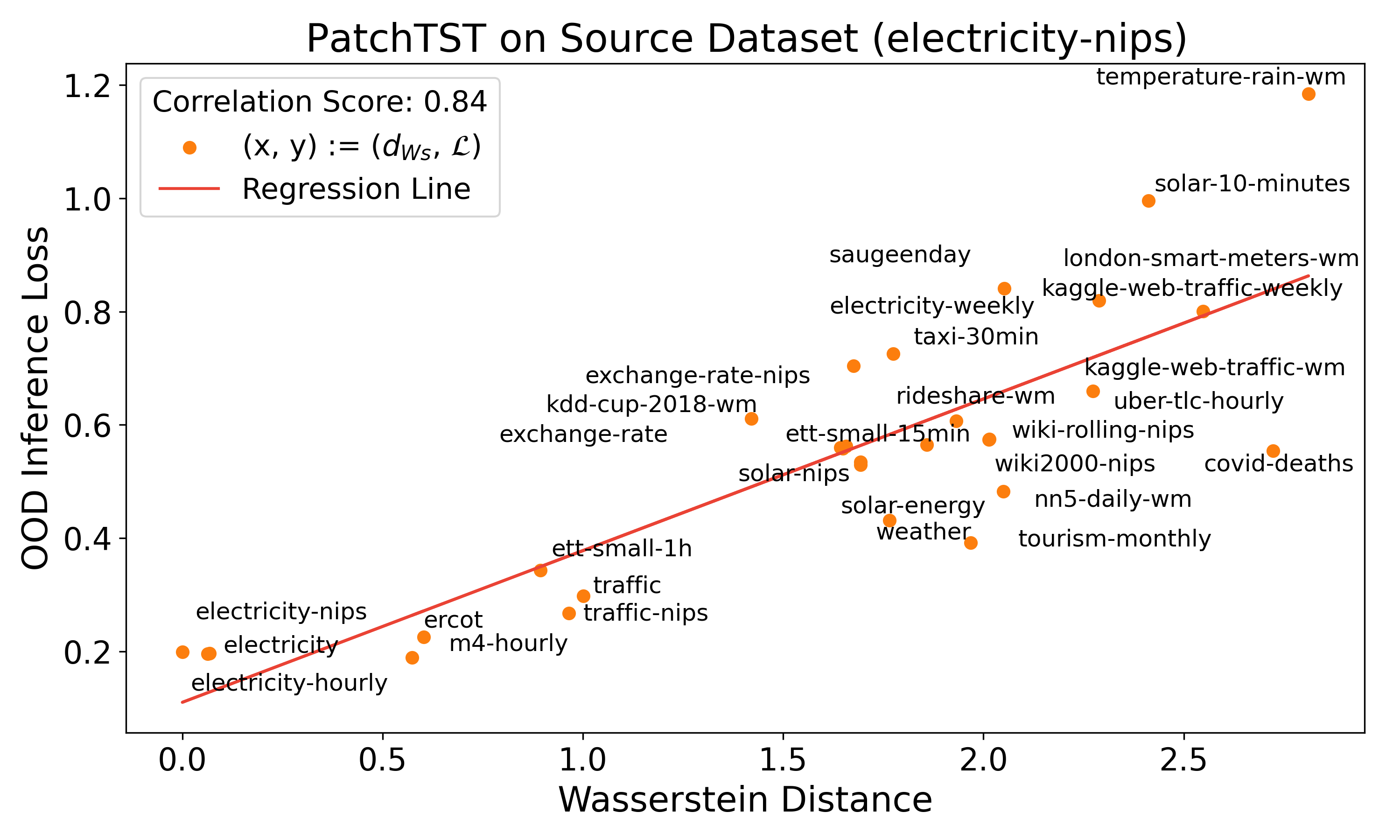}
\includegraphics[width=0.16\linewidth]{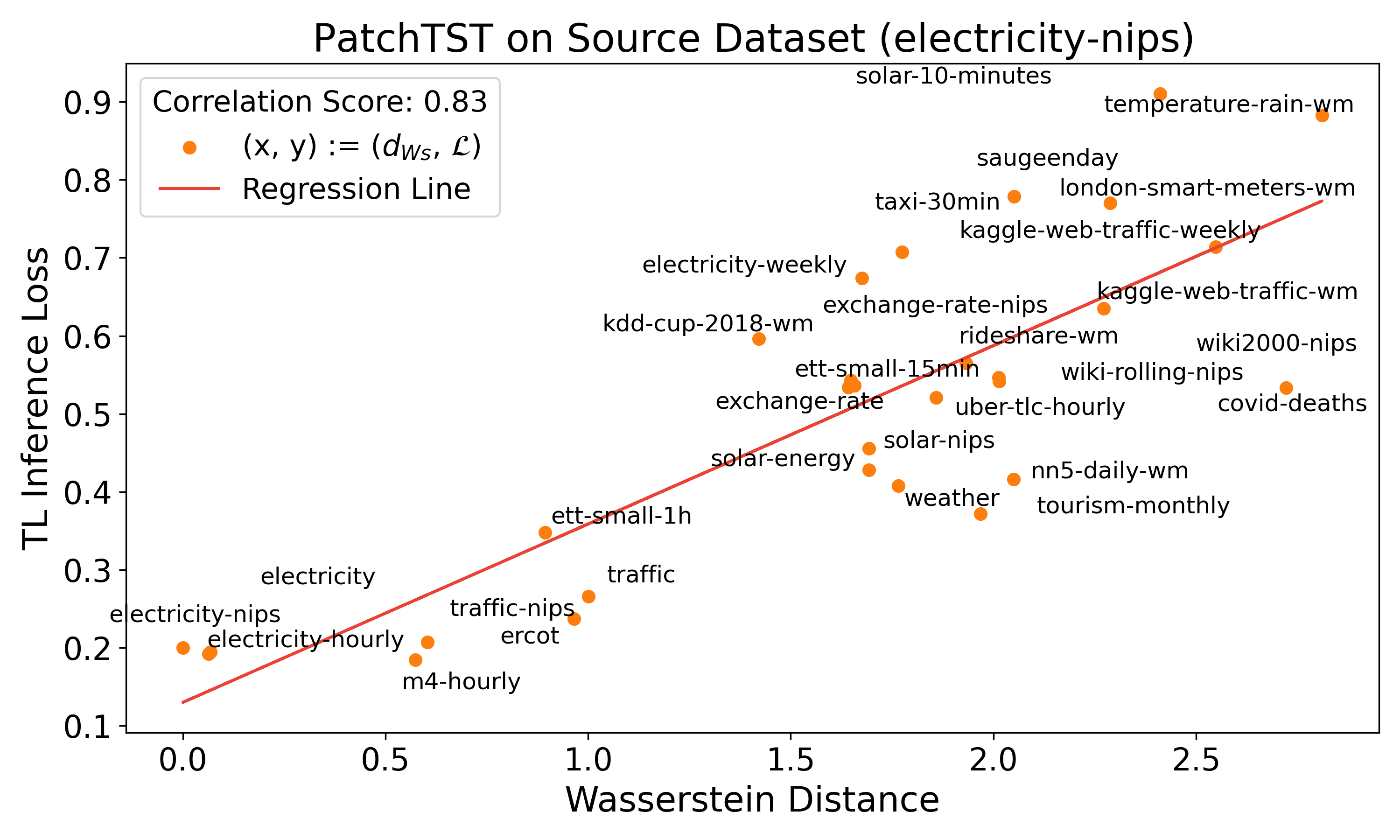}

\caption{
Out-Of-Distribution (OOD) and Transfer Learning (TL) inference losses of \textit{PatchTST} trained on each individual dataset.
}
\label{fig:complete-PatchTST-OOD-TL-pairs}
\end{figure*}

\section{Experimental Details}
\label{sec:append-exp-details}
We use the original training parameters for both \textit{Lag-llama} and \textit{PatchTST}.
For \textit{Lag-llama}, we set learning rate as $0.0005$, number of layers as $8$, number of embeddings per head as $16$, number of heads as $9$, batch size as $64$, and maximum training epochs as $50$.
For \textit{PatchTST}, we set learning rate as $0.001$, patch length as $16$, stride as $8$, number of heads as $4$, feed forward dimension as $128$, dropout as $0.1$, number of encoder layers $2$, batch size as $32$, and maximum training epochs as $50$.
All experiments are run on a machine of 12 CPUs, memory of 125 Gi, NVIDIA A2 of 15356 MiB.

Both models are optimized using the negative log likelihood loss.
To reduce the computational time, we train on all $N=20,000$ samples and test on only $N_{test}=1,000$ samples of each target dataset.
Each \textit{Lag-llama} experiment takes about 6 minutes to complete.
Each \textit{PatchTST} experiment takes about 1 minute to complete.
The overall compute takes about 9 days. 
The used datasets and models are all under Apache-2.0 license.

\subsection{Complete Experimental Results}
\label{sec:append-complete-results}
Fig.~\ref{fig:complete-lagllama-OOD-TL-pairs}-~\ref{fig:complete-PatchTST-OOD-TL-pairs} in the appendix are the complete experimental results for foundation model \textit{Lag-llama} and \textit{PatchTST}, respectively.
The lines of best fit are highlighted in red. For optimal detail inspection, we recommend using the zoom-in tool in the PDF format.

\subsection{Details of Time Complexity Analysis}
\label{sec:append-time-complexity}
In Eq.~\ref{eq:Wd}, calculating the first term $\norm{\hat{\vmu}_{\bmX} - \hat{\vmu}_{\bmY}}^2$ takes $\bigO\pare{NL}$ time complexity, mainly due to the aggregation of the mean vectors $\hat{\vmu}_{\bmX}$ and $\hat{\vmu}_{\bmY}$.
For the second term $\trace\pare{\hat{\mSigma}_{\bmX} + \hat{\mSigma}_{\bmY} - 2\sqrt{\hat{\mSigma}_{\bmX} \hat{\mSigma}_{\bmY}}}$, it takes $\bigO\pare{NL^2}$ to calculate the covariate matrices, $\bigO\pare{L^3}$ to perform the matrix product and compute the matrix square root, and $\bigO\pare{L}$ to calculate the trace of the matrix.
Thus, the overall time complexity is
\begin{align}
    \bigO\pare{NL^2} + \bigO\pare{L^3} + \bigO\pare{L} = \bigO\pare{\pare{N+L} L^2}
\end{align}
The covariance matrices become more accurate as $N$ increases.
The Wasserstein distance allows us to flexibly select the number of samples $N$ to be much greater than the time-series length $L$, such that $N \gg L$.
This further simplifies the time complexity to be $\bigO\pare{NL^2}$.

\clearpage

\end{document}